\documentclass[nohyperref]{article}

\usepackage{microtype}
\usepackage{graphicx}
\usepackage{subfigure}
\usepackage{booktabs} 

\usepackage{tikz}
\usepackage{comment}
\usepackage{cite}
\usepackage{amsmath,amssymb,amsfonts}
\usepackage{algorithmic}
\usepackage{textcomp}
\usepackage{float}
\usepackage{xspace}
\usepackage{makecell}
\usepackage{fixltx2e}
\usepackage{fancyhdr}
\usepackage{lipsum}
\usepackage{soul,xcolor}

\usepackage{filecontents}

\newif\ifdraft
\draftfalse

\newif\ifpreprint
\preprinttrue
\setstcolor{blue}
\newcommand{\figref}[1]{Fig.~\ref{#1}}
\newcommand{\figrefs}[2]{Figs.~\ref{#1}--\ref{#2}}
\newcommand{\tabref}[1]{Tab.~\ref{#1}}
\newcommand{\secref}[1]{\S\ref{#1}}
\newcommand{\secrefs}[2]{\S\ref{#1}--\ref{#2}}
\newcommand{\appref}[1]{App.~\ref{#1}}
\newcommand{\algoref}[1]{Alg.~\ref{#1}}
\newcommand{\eqnref}[1]{Eqn.~\ref{#1}}

\newcommand{\Overrun}{\mathit{Overrun}}
\newcommand{\LossBoundary}{\mathit{L_{bnd}}}
\newcommand{\LossClassification}{\mathit{L_{cls}}}
\newcommand{\ReLU}{\mathit{ReLU}}
\newcommand{\LowerBoundary}[1]{\mathit{bnd^{lower}_{#1}(x)}}
\newcommand{\UpperBoundary}[1]{\mathit{bnd^{upper}_{#1}(x)}}
\newcommand{\TargetOffset}{\mathit{target\_offset}}
\newcommand{\Imagenet}{ImageNet\xspace}
\newcommand{\CondtionalSuccess}{\mathit{ConditionalSuccess}}
\newcommand{\MinimalDistance}{Minimal Difference }
\newcommand{\MDabbrv}{MD\xspace}
\newcommand{\dlr}{DLR\xspace}
\newcommand{\cgd}{CGD\xspace}
\newcommand{\cgduntarg}{CGD\textsubscript{untarg}\xspace}

\newcommand{\pgd}{PGD\xspace}
\newcommand{\cw}{CW\xspace}
\newcommand{\Autopgd}{Auto-PGD\xspace}
\newcommand{\pytorch}{PyTorch\xspace}
\newcommand{\lp}[1]{$\ell_{#1}$}

\newcommand{\DSLH}{DSL+20\xspace}
\newcommand{\WRK}{WRK20\xspace}
\newcommand{\HLM}{HLM19\xspace}
\newcommand{\WZYBMG}{WZY+20\xspace}
\newcommand{\SWMJ}{SWM+20\xspace}
\newcommand{\CRSLD}{CRS+19\xspace}
\newcommand{\WXW}{WXW20\xspace}
\newcommand{\SIEKM}{SIE+20\xspace}
\newcommand{\whitebox}{white-box\xspace}
\newcommand{\robustbench}[0]{\texttt{RobustBench}\xspace}

\usepackage{hyperref}

 \usepackage[accepted]{icml2022}

\usepackage{amsmath}
\usepackage{amssymb}
\usepackage{mathtools}
\usepackage{amsthm}

\usepackage[capitalize,noabbrev]{cleveref}

\theoremstyle{plain}

\theoremstyle{definition}

\theoremstyle{remark}

\usepackage[textsize=tiny]{todonotes}
\usepackage[font=normalsize]{caption}

\icmltitlerunning{Constrained Gradient Descent: A Powerful and Principled
  Evasion Attack Against Neural Networks}

\begin{document}

\twocolumn[
\icmltitle{Constrained Gradient Descent:\\ A Powerful and Principled
  Evasion Attack Against Neural Networks}




\begin{icmlauthorlist}
\icmlauthor{Weiran Lin}{cmu,cylab}
\icmlauthor{Keane Lucas}{cmu,cylab}
\icmlauthor{Lujo Bauer}{cmu,cylab,isr}
\icmlauthor{Michael K.\ Reiter}{duke}
\icmlauthor{Mahmood Sharif}{telaviv}
\end{icmlauthorlist}

\icmlaffiliation{cmu}{Department of Electrical \& Computer Engineering, Carnegie Mellon University, Pittsburgh, PA, US}
\icmlaffiliation{cylab}{Cylab, Carnegie Mellon University, Pittsburgh, PA, US}
\icmlaffiliation{isr}{Institute for Software Research, Carnegie Mellon University, Pittsburgh, PA, US}
\icmlaffiliation{duke}{Departments of Computer Science and Electrical \& Computer Engineering, Duke University, Durham, NC, US}
\icmlaffiliation{telaviv}{School of Computer Science, Tel Aviv University, Tel Aviv, Israel}

\icmlcorrespondingauthor{Weiran Lin}{weiranl@andrew.cmu.edu}
\icmlcorrespondingauthor{Keane Lucas}{kjlucas@andrew.cmu.edu}
\icmlcorrespondingauthor{Lujo Bauer}{lbauer@cmu.edu}
\icmlcorrespondingauthor{Michael K. Reiter}{michael.reiter@duke.edu}
\icmlcorrespondingauthor{Mahmood Sharif}{mahmoods@cs.tau.ac.il}

\icmlkeywords{Machine Learning, ICML, Robustness, Adversarial Attacks, Computer Vision, Deep Learning}

\vskip 0.3in
]



\printAffiliationsAndNotice{}  

\begin{abstract}

We propose new, more efficient targeted white-box attacks against deep neural networks. Our 
attacks better align with the attacker's goal: (1) tricking a model to assign higher 
probability to the target class than to any other class, while (2) staying within an $\epsilon$-distance 
of the attacked input. First, we demonstrate a loss function that explicitly encodes (1) and 
show that Auto-PGD finds more attacks with it. Second, we propose a new attack method, Constrained Gradient Descent (\cgd{}), using
a refinement of our loss function that captures both (1) and (2).
\cgd{} seeks to satisfy
both attacker objectives---misclassification and bounded \lp{p}-norm---in
a principled manner, as part of the optimization, instead of via ad hoc
post-processing techniques (e.g., projection or clipping).
We show that \cgd{} is more successful on CIFAR10
(0.9--4.2\%) and ImageNet (8.6--13.6\%) than state-of-the-art attacks while consuming less time (11.4--18.8\%). Statistical tests
confirm that our attack outperforms others against leading defenses on different datasets and 
values of $\epsilon$.

\end{abstract}

\section{Introduction}
\label{sec:intro}

With the prevalence of machine learning (ML), adversarial ML
techniques that slightly manipulate the inputs of an ML
model to influence its functionality have also been
developed~\cite{icml20:autopgd}.
One type of attack on classification models is the
\textit{evasion} attack, which applies a small perturbation, within a
distance limit, to a classifier's input to 
induce misclassification during inference.
 The \lp{\infty} \textit{distance} is commonly used for specifying
distance limits via $\epsilon$ \textit{boundaries}, which define the maximum
change acceptable in each element of the input.
In previous work it is also common for attackers to have access to all weights of the 
model, known as the \textit{\whitebox} scenario, and the attackers aim to
force a specific misclassification,  
performing \textit{targeted} attacks. Previous work proposes an attack method that iteratively 
perturbs an input
in the 
direction of the gradients of a loss function, which  is not necessarily the same loss function 
used 
by the model, and automatically truncates the attack in each iteration to stay within the distance 
limits. Researchers have shown that such attacks are effective
against state-of-the-art neural networks~\cite{CVPR16:Deepfool,iclr14:Intriguing}.

Croce et al.\ show that well-tuned parameters and carefully designed loss
functions can boost the performance of attacks~\cite{icml20:autopgd}. For example, varying the loss function of Auto Projected
Gradient Descent (\Autopgd), a state-of-the-art attack, between the
cross-entropy, Carlini and Wagner (\cw{}), and Difference of Logits
Ratio (\dlr{}) losses has a substantial impact on the attack's
performance. Guided by this observation, we define a new loss function, the
\textit{\MinimalDistance loss (\MDabbrv loss)}, that better aligns with the
goal of a targeted attack: \MDabbrv loss aims to 
(mis)lead the model to assign higher confidence to the target class than to
\emph{any other class}, even if by just a tiny amount. We empirically show that \Autopgd with the
\MDabbrv loss finds on average 0.5--12.3\%
more adversarial examples, depending on the dataset and
model, than
\Autopgd with other loss functions.

Although \MDabbrv loss substantially improves \Autopgd's performance, we still
notice limitations that hinder the attack's effectiveness. In
particular,
like other attacks in the \pgd{} family, 
\Autopgd uses
projection at the end of each iteration to satisfy the norm constraints,
eliminating changes that fall outside a predefined \lp{p} ball.
However, projecting adversarial examples back into the \lp{p} ball may work against
the attacker's misclassification objective, thus harming the attack's
success. Moreover, because projection is implemented by clipping,
an \emph{ad hoc} operation to eliminate any changes made outside the
\lp{p} ball, balancing 
the attack's two objectives (i.e., misclassification and not leaving the \lp{p} ball)
remains a challenge for \Autopgd{}. To address this shortcoming, we propose the
\textit{Constrained Gradient Descent (\cgd)} attack. \cgd enables the attacker
to balance the two attack objectives---it allows adversarial examples to lie
outside the \lp{p} ball during the attack process and models the violation of the
norm constraints as part of its loss function to gradually learn to stay
within the \lp{p} ball while achieving misclassification. We find
that, with the same \lp{\infty} distance limit, \cgd finds on average 0.3--1.3\%
more adversarial
examples than \Autopgd using the \MDabbrv loss, while consuming less time (11.41--18.76\%).
 Furthermore, we show
that \cgd outperforms \Autopgd with three previously established loss functions
on the CIFAR10 and \Imagenet datasets, with statistical significance.

As an example of using \cgd as a framework to find other types of
attacks, we also define a variant of \cgd for \whitebox untargeted
attacks. Similar to its targeted attack counterpart, this variant also
gradually learns to stay within the \lp{p} ball while achieving
misclassification, due to modeling the violation of the norm
constraints as part of its loss function.  This variant
outperforms the previous best attack by 0.3--2\% on the CIFAR10 dataset, on average.

In a nutshell, our contributions are:
\begin{itemize}
  \item We improve an established targeted attack method by offering a new loss
    function that better captures the goal of targeted attacks
    (\secref{sec:loss}). 
  \item We invent a new attack that learns to stay within the \lp{\infty}
    distance limit rather than using simple clipping (\secref{sec:cgd}), and  
    we empirically demonstrate that it outperforms previously established ones
    (\secrefs{sec:setup}{sec:result}).
  \item We demonstrate how the proposed method can be used as a framework for
    attacks by instantiating it to define a stronger untargeted attack
    (\secref{sec:discussion}).
\end{itemize}

\section{Background}
\label{sec:background}


\subsection{Threat Model}
\label{sec:bg:threatmodels}

\paragraph{Adversary goals}
\label{sec:bg:attackergoals}
We consider a supervised classification setting in which an ML model $F$ is trained to map a sample $x$ to the correct label $y$ by minimizing
a loss function $L(x,y)$ such as the cross-entropy loss ($L_{CE}$). At inference
time, a sample $x$ is assigned the class $i$ with the highest logit $Z_i$ or
highest confidence $P_i$. To find an adversarial example $x'$, the adversary could either launch an untargeted
attack, avoiding correct classification by maximizing $L(x',y)$, or launch a
targeted attack, forcing specific classification to class $t$ by minimizing
$L(x',t)$~\cite{eurosp16:PapernotLimitations}.
We permit the attacker white-box access to $F$, i.e., so that the
attacker knows the internal weights of $F$.

\paragraph{Evaluation metrics}
\label{sec:setup:Evaluation}
Given a specific \lp{p}
distance limit $\epsilon$, the success rate of
an untargeted attack is computed as the percentage of benign inputs
$x$ from which the attack finds $x' \in \{\tilde{x} \mid F(\tilde{x}) \neq y \wedge \ell_p(\tilde{x},x) \le \epsilon\}$,
whereas the success rate of a targeted
attack is defined as the percentage of benign inputs $x$ where
the attack finds
$x' \in \{\tilde{x} \mid F(\tilde{x})=t \wedge \ell_p(\tilde{x},x) \le \epsilon\}$.
In this paper, we primarily study targeted attacks for
$\ell_{\infty}(x, x')=\max_{i,j,k} \left| x_{i,j,k}-x'_{i,j,k} \right|$,
i.e., \lp{\infty} measures the maximum change made across all pixels
and channels, where $i$ and $j$ are pixel coordinates, and $k$ is the
channel index.

Successful adversarial examples should also be in the same
format as benign samples.  
Images 
are normally in 8-bit RGB format: every pixel
consists of three bytes, three integers $\in [0,255]$ normalized
to floats $\in [0,1]$. The value of every channel (of every pixel) should be a
multiple of $1/255$. We picked $\epsilon$ values that are multiples of $1/255$
so that the \lp{\infty} distance limit is in 8-bit RGB format. We noticed that
certain attacks (e.g., \pgd{}) could perturb an adversarial examples into
failed ones after quantization. Hence, in each attack iteration $i$ from starting sample $x$,
we produced an 8-bit RGB format copy of the
current adversarial example $x'_i$
using the formula
\[x_{\mathit{test}}=\mathit{round}(x'_i*255)/255\]
and clipped $x_{\mathit{test}}$ to be within both $[0,1]$ and
$[x-\epsilon,x+\epsilon]$ (we denote this operation by $\mathit{clip(x_{\mathit{test}})}$), projecting it onto the
\lp{\infty}
ball that is also bounded by the range of values of valid images.
We then classified $\mathit{clip}(x_{\mathit{test}})$ using
the model and compared the output with the target class.
If they matched, we stopped
perturbing this example and counted the attack as successful. Otherwise, the attack
continued normally, with $x'_i$ remaining in the continuous domain.

\subsection{Attack Methods}
\label{sec:bg:attacks}

We now introduce prominent established attacks.

\paragraph{PGD}
An improved version of fast gradient-sign method~\cite{iclr15:adversarial}, the
projected gradient descent (PGD)
attack~\cite{iclr18:PGD}, which iteratively calculates a projection on the imperfect
$\epsilon$ ball around the benign source image so that the adversarial
example is a valid image and within the max distance limit:
\begin{equation}
x'_{i+1}=\mathit{clip}(x'_i-\alpha \cdot sign(\frac{\partial L(x'_i,t)}{\partial x'_i})) \label{eq3}
\end{equation}
where $\alpha$ controls how much perturbation
would be applied in each iteration and loosens the linear assumption of
models for \pgd. By default, \pgd{} is configured to run for
40 iterations, and the step size of each iteration, $\alpha$, is
set to .01, where the values of each channel (of each pixel) in the images are
normalized to $[0,1]$. The default loss function used in \pgd{} is
the CE loss.

\paragraph{Auto-PGD}
Croce and Hein (\citeyear{icml20:autopgd}) were able to boost \pgd{}'s performance
by intelligently setting its parameters (e.g., the step size
$\alpha$). Except for the number of iterations, the
algorithm they proposed, \Autopgd{}, does not require parameter tuning.
Croce and Hein also showed that the loss function in \pgd{} can markedly
influence the attack's success rate.
By default, \Autopgd{} runs with the Difference of Logits Ratio
(\dlr) loss:
\begin{equation}
L_{\mathit{DLR}}=\frac{Z_t-Z_y}{Z_{\pi1}-0.5Z_{\pi3}-0.5Z_{\pi4}} \label{eq8}
\end{equation}
where $Z_\pi$s are logits sorted from largest to smallest.
 Another commonly used
loss function is the Carlini-Wagner (\cw) loss~\cite{Oakland17:CarliniWagner}:
\begin{equation}
L_{\mathit{CW}}=-Z_t+max_{i \neq t} Z_i\label{eq9}
\end{equation}
$L_{\mathit{CW}}$ and $L_{\mathit{DLR}}$ both try to make the logit of the target class larger
than other logits, which would cause the model to assign the highest probability
to the target class.

\subsection{Defenses}
\label{sec:back:defenses}

Researchers have proposed a variety of defenses to mitigate
adversarial examples. Regarded as one of the strongest
defenses~\cite{arxiv21:Akhtar2021AdvancesIA},
\emph{adversarial training} augments the training process
with correctly labeled adversarial examples to enhance
models' robustness against them
(e.g.,~\cite{iclr15:adversarial, iclr18:PGD,
NeurIPS19:FreeTraining, iclr20:WZYBMG20,
iclr17:Kurakin2017AdversarialML}). As is common in related
work~\cite{icml18:Uesato2018AdversarialRA,icml20:autopgd,
arxiv18:Xiao2018GeneratingAE, iclr18:PGD, CVPR18:Momentum},
we evaluate our proposed attacks against adversarially
trained neural networks (see
\secref{sec:setup:models}).

\section{A Stronger Loss Function}
\label{sec:loss}

In this section, we describe how we enhanced \Autopgd's performance by
improving its loss function. We 
first present an example scenario where a previous best-performing loss
function falls short (\secref{ssec:previous-loss}). Then we describe a new loss function that mitigates 
the
previous loss function's weaknesses (\secref{sec:loss:newloss}). We
report on the performance of the new loss function in
\secref{sec:result}.

\subsection{Investigating Established Loss Functions}
\label{ssec:previous-loss}
\begin{figure}[t!]
  \centering
  \includegraphics[width=1.0\columnwidth]{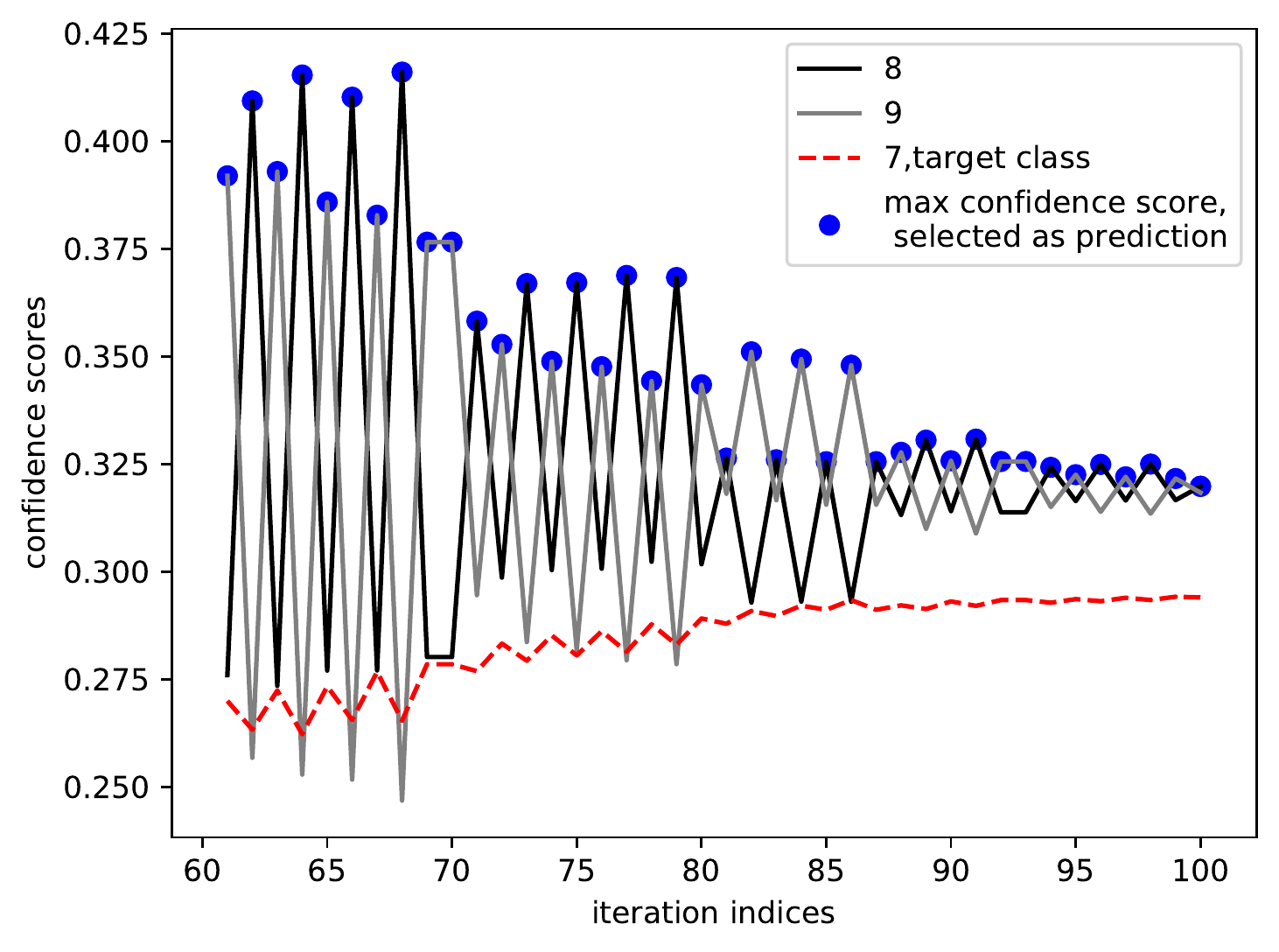}
\caption{Confidence scores of three classes throughout 40 iterations of the
  \Autopgd attack using \cw loss to perturb a CIFAR10 image to class 7. 
  Classes 8 and 9 took turns to have the highest confidence
  score,
  and the attack ultimately failed to produce a successful targeted
  adversarial example.}
\label{cwfails}
\end{figure}

To produce a baseline for comparison, we ran
\Autopgd for 
100 iterations, as it is used in previous work~\cite{icml20:autopgd}, with
three loss functions: CW, \dlr{}, and CE.
We found that CW loss performed the best against six out of seven CIFAR10
models as well as the \Imagenet
model we used.
Still, we identified specific instances in which \Autopgd with CW loss 
failed to find adversarial examples. One of them is demonstrated in \figref{cwfails}. We identified that 
these failures
can be often explained by the definition of CW loss.
CW loss tries to increase the difference
between the logit of the target class and the highest among the logits of the non-target
classes; however, \emph{which} non-target class has the highest logit may change from iteration to iteration.
Namely, decreasing the value of the logit of the highest non-target
class might simply cause the logit of \emph{another}  
non-target class to 
increase and become the highest in the next iteration. Thus, the target class
might never have a chance to
have the highest logit and the attack might never succeed, as
illustrated in \figref{cwfails}.

We noticed that such failures can also happen with more than two non-target classes
taking turns having the highest confidence scores. 
 We used 
 the maximum number of times that the prediction was \textit{changed} to a certain class
 before the perturbation first succeeds or the attack reaches the maximum number of iterations
as a metric to capture
how often the behavior in \figref{cwfails} occurs.
 For 226 of 512 images sampled from CIFAR10, there is some $\epsilon$ at which 
 the prediction was \textit{changed} to the same class at least 10 times when \Autopgd with CW loss attacked the \DSLH~\cite{iclr20:DSLH20} defense. We observed the same phenomenon with 265 of 512 samples on the \WRK~\cite{iclr20:WRK20} defense. More details could be found in \appref{app:fluctuation}.

\subsection{A New Loss Function}
\label{sec:loss:newloss}

The example in \figref{cwfails} shows how a loss function that does
not completely align with the attack's intent can sometimes fail to produce
successful adversarial examples. Intuitively, the attacker's goal is to produce
a perturbation for which the target class' logits and probabilities are higher
than those of \emph{all} other classes, even if by only a miniscule amount, so
that the target class would be selected as the prediction.

We capture this intuition by designing a new loss function containing terms
representing each logit. More precisely, the 
loss function is the sum over all such terms (each representing a logit), where
each term's value is determined as follows. The term is zero if the logit is
smaller than the logit of the target class, as the adversary has no interest 
in directly decreasing small logits that do not influence the model's
prediction. Alternately, the term is positive if the logit does not belong to
the target class and is larger than or equal to the target class's logit, as the
adversary has to decrease all non-target logits to make a successful
perturbation.
By minimizing this loss function, the adversary aims to decrease all the
positive terms simultaneously and consequentially decrease \emph{all} the logits that
are higher than the logit of the target class in the same iteration.

We define this loss function, the \MinimalDistance (\MDabbrv) loss, as:
\begin{equation}
\sum_i ReLU(Z_i+\delta-Z_t)\label{MD}
\end{equation}
where ReLU is the rectified linear unit function, $Z_t$ is the logit of the
target class, and $Z_i$ are the logits
of each
class (as described in \secref{sec:bg:attackergoals}). $\delta$ is a
minimal value introduced to mitigate the cases where non-target classes have equal  
logits with the logit of the target class. We set $\delta$ to
$1\mathrm{e}{-15}$ in our
implementation, because $1.0-(1\mathrm{e}{-16})=1.0$ in Python
due to finite arithmetic. $Z_t$ is not excluded from $Z_i$ as 
$ReLU(Z_t+\delta-Z_t)=\delta$ 
is always a constant and has no effect on the back-propagation
gradients.

In contrast to
\cw loss, \MDabbrv loss aims to decrease \textit{all} logits that do not belong to the target class 
and are higher than the logit of the target class, rather than only
the largest of the non-target-class logits. Logits that meet the
following three requirements are more likely to decrease between iterations if the attacker is using 
\MDabbrv loss instead of \cw loss:
(1) they do not belong to the target class;
(2) are higher than the logit of the target class; and
(3) are not the largest logit.
Hence, with \MDabbrv loss the behavior demonstrated in
\figref{cwfails}
 (i.e., non-target
logits alternating at being the highest logit) is unlikely to
occur and,
intuitively, the attacker is more likely to succeed. 
We found that \MDabbrv loss reduced the maximum number of times that the
prediction was changed to a different class at every value of $\epsilon$ (see
\appref{app:fluctuation}) and confirmed the statistical significance of this
result with a Wilcoxon signed-rank test \cite{stat45:Wilcoxon} (details in
\appref{app:wilcoxon}).
We show in \secref{sec:result} that using \MDabbrv loss instead of \cw
loss consequently
improves \Autopgd's ability to find adversarial examples.

\section{A Stronger Attack Method}
\label{sec:cgd}

In this section, we describe our new attack method,
Constrained Gradient Descent (\cgd{}). We start by explaining the
intuition that drives the design and an enhancement
(\secrefs{sec:limit}{sec:minima}), and then 
detail our algorithm (\secref{sec:cgd-details}).  We
report on the performance of the new attack in
\secref{sec:result}.

\subsection{Learning to Stay Within the Distance Limit}
\label{sec:limit}
\pgd{} attacks, including \Autopgd{}, enforce the \lp{\infty} distance
limit by executing a $\mathit{clip(\cdot)}$ computation in each
iteration, eliminating any perturbations made outside the limit. However, the loss function used in \pgd{} 
attacks does not
take into account that $\mathit{clip(\cdot)}$ will be used. Hence, in \pgd{}
attacks, gradients, which are
derivatives of loss functions against the current perturbation,
may not accurately direct the perturbation. The gradients may
push the attack toward perturbations outside the \lp{\infty}
distance limit, while successful perturbations within the \lp{\infty}
distance limit may lie in other directions.
Previous 
work has shown
that attacks could \emph{learn to minimize} their distance limit from the
original benign
image~\cite{Oakland17:CarliniWagner,iclr14:Intriguing}.
We propose a new loss function that helps attacks \emph{learn to stay within} a 
fixed \lp{\infty} distance limit, and then propose a new attack that utilizes this loss function.
We next describe how we include the \lp{\infty} distance limit as part of the new loss function.

\paragraph{Defining boundaries}
Before starting to create a perturbation $x$, we know that
for an attack to be valid the upper and lower boundaries
of each channel $k$ $\in \{0,1,2\}$ of pixel $(i,j)$ in the final perturbation are
\begin{equation}
\UpperBoundary{i,j,k}= \min(x_{i,j,k}+\epsilon,1)
\end{equation}
and \begin{equation}
\LowerBoundary{i,j,k}=\max(x_{i,j,k}-\epsilon,0)
\end{equation}
These two boundaries are fixed throughout the attack process. As
$x_{i,j,k} \in [0,1]$, in each channel (of each pixel), in any iteration of an attack the updated perturbation could potentially exceed
at most one boundary at a time (for any one channel of any pixel). Thus, we can compute the
distance, for each channel (of each pixel), by which the current perturbation $x'$
goes over the boundary. We call this distance $\Overrun_{i,j,k}(x')$
and define it as:
\begin{equation}
\ReLU(x'_{i,j,k}-\UpperBoundary{i,j,k})+\ReLU(\LowerBoundary{i,j,k}-x'_{i,j,k})
\end{equation}

\noindent $\Overrun_{i,j,k}(x')$ is 0 if $x'$ stays within the $\epsilon$-boundary for channel $k$ of pixel 
$(i,j)$.

\paragraph{Penalizing exceeding boundaries}
We next add to the loss function the following term.
\begin{equation}
\LossBoundary=\sum_{i,j,k} (\Overrun_{i,j,k}(x'){}^2)
\end{equation}
$\Overrun_{i,j,k}(x')$ is squared because when
taking the derivative,
$\frac{\partial \LossBoundary}{\partial x'_{i,j,k}}$
is proportional to $\Overrun_{i,j,k}(x')$ and
could direct the attack to learn to stay within the boundary.
Our new
loss function is now:
\begin{equation}
w*\LossClassification+(1-w)*\LossBoundary \label{ourloss}
\end{equation}
where $\LossClassification$ is MD-loss (as described in
\secref{sec:loss}) and $w\in[0,1]$ is a weighting parameter. By
decreasing $w$ and weighting $\LossBoundary$
more heavily, we guide the attack to
gradually move the perturbation in a direction that leads to the
target class while not crossing the boundary, ultimately creating
a valid adversarial example after rounding and clipping.

\paragraph{Leveraging magnitude of gradients}
We also change the way that the attack iteratively updates the perturbation.
To the best of our knowledge, previous \lp{\infty} attacks that use gradients, $\frac{\partial
  Loss}{\partial x'_{i,j,k}}$, only use the sign of these gradients to generate adversarial examples 
(e.g.,~\cite{iclr17:BasicIterative,iclr18:PGD,iclr15:adversarial}), although some use momentum along with the sign~\cite{icml20:autopgd,arxiv19:nonsign}.
However, the magnitude of the gradients also conveys information about the
amount by which it is helpful to change a channel (of a pixel).
Leveraging the magnitude of the gradients also helps avoid
artificially (and unhelpfully) large step sizes that would make the
perturbations step over the boundary, which we observed in,
e.g., \pgd{} attacks. 
\Autopgd uses an explicit momentum term as well as gradients when computing the changes to be made to
the candidate adversarial example in each iteration. In contrast, we
compute these changes via an Adam optimizer~\cite{iclr15:Adam}, which
internally uses momentum; other optimizers may also be adequate.

\subsection{Driving out of Local Minima}
\label{sec:minima}
While $\LossBoundary$ accounts for any
channel (of any pixel)
potentially crossing the $\epsilon$-boundary, we
observed that attacks could become trapped in local minima of
$\LossBoundary$ that occur when most channels 
are within the boundary but a small number is far beyond the boundary. To
prevent this, we set a threshold distance outside the upper and lower $\epsilon$
boundaries, and we decay this
threshold gradually as the attack progresses.
As $\LossClassification$ decreases, we desire the
attack to also reduce $\Overrun$, ultimately to zero. The
decreasing threshold encourages this by increasing the relative weight of 
$\LossBoundary$. 
Model-specific constants, namely pre-defined fixed ratios and checkpoints, 
control where the threshold starts and how fast it decays.
If the current perturbation is
outside the threshold, we halve the weight $w$ of $\LossClassification$ (in \eqnref{ourloss})
to increasingly urge the attack to stay within the boundary.

We ran grid searches to choose the starting threshold and decay
interval for each defense. We examined starting thresholds 
$\in[.5\epsilon, 9\epsilon]$, starting the weighting parameter of the loss function $w \in [0.01,0.5]$, 
and decay intervals $\in[5, 30]$.
For all defenses, the optimal starting weighting parameter $w$ was 0.1 and the optimal decay interval 
was every 15
iterations. The optimal starting threshold was $8\epsilon$ for \SIEKM,
$5\epsilon$ for \DSLH, and $1.5\epsilon$ for all other defenses (see \secref{sec:setup}).

\subsection{The Constrained Gradient Descent Algorithm}
\label{sec:cgd-details}
\begin{figure}[tb!]
  \centering
  \includegraphics[width=1.0\columnwidth]{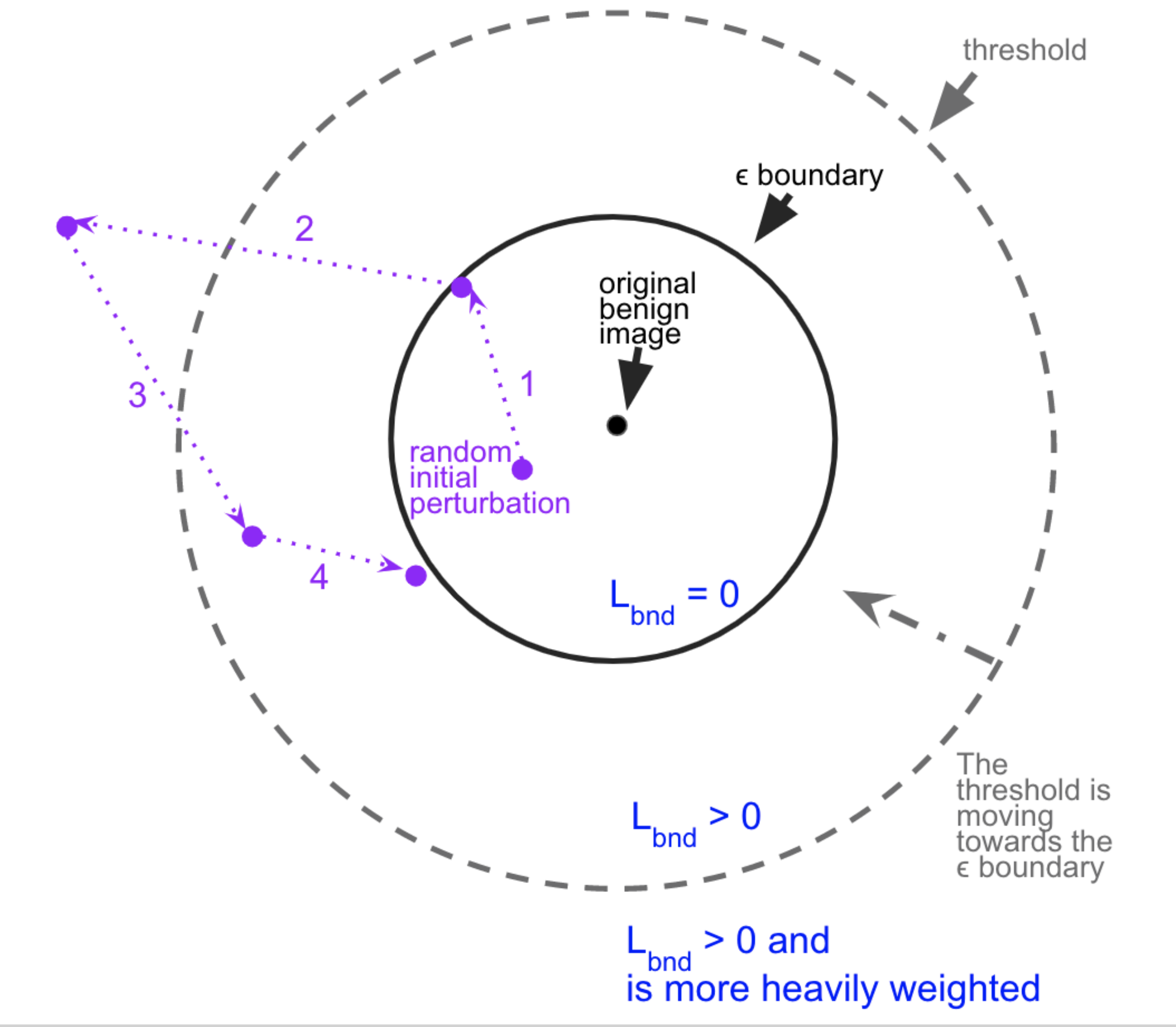}
   \caption{This is an example path of a CGD attack with a \lp{\infty} distance
     limit. We start with a
     random initial perturbation. In stage 1, we push the current
     perturbation to the $\epsilon$ boundary. In stage 2, the current
     perturbation moves beyond the threshold and $\LossBoundary>0$. In
     stage 3, the current perturbation is pushed inside the threshold
     as $\LossBoundary$ is more heavily weighted. In stage 4, the
     current perturbation moves closer to the $\epsilon$ boundary, as
     does the threshold.} 

     \label{fig:heuristic}
\end{figure}

Combining all the above, we define a new attack:
\textit{Constrained Gradient Descent (\cgd)}. \figref{fig:heuristic} illustrates an example path of \cgd 
where the attack seeks to satisfy the \lp{\infty} distance limit.

We start the attack with a random initial perturbation to better explore the space of possible 
adversarial examples, as this was shown to be helpful in prior work~\cite{iclr18:PGD, 
  icml20:Croce2020Minimally, neuripssecml18:Mosbach2018LogitPM}. The attack has four stages.
In stage~1, we move each channel (of each pixel) by $\epsilon$ in the direction of the gradients.
This is a quick way to take a substantial step in the direction of the target
class. 
In stage~2, the candidate perturbation continues to move toward the target
class, and potentially
moves  outside the threshold, as the loss function is dominated by  
$\LossClassification$. If the candidate perturbation moves beyond the
threshold, the algorithm moves to stage~3: since the candidate perturbation is
outside the threshold, $\LossBoundary$ is more heavily weighted, which 
pushes the candidate perturbation back inside 
the threshold. After a fixed number of iterations, the algorithm enters stage~4,
in which the threshold itself moves toward the $\epsilon$ boundary, thus forcing
the candidate perturbation to move closer to the $\epsilon$ boundary. 
The attack could succeed in any stage.
Pseudocode with line-by-line descriptions can be found in \appref{app:cgdalgo}.

\section{Evaluation Setup}
\label{sec:setup}

Factors
other than the algorithm, such as the random initialization
chosen~\cite{NeurIPS20:Initializations} and which classes are targeted, can also
influence attacks' performance. 
Here we summarize how we set up experiments to 
enable meaningful and fair comparisons; more details can be found
in \appref{app:setup}.

\paragraph{Benchmarks}
\label{sec:setup:models}

Because adversarial training is regarded as a strong defense, 
we evaluated attacks against adversarially trained models, in line with prior
work (see \secref{sec:back:defenses}).
Specifically, we used seven pre-established adversarially trained models for
CIFAR10: \CRSLD~\cite{NeurIPS19:CRSLD19}, \DSLH~\cite{iclr20:DSLH20},
\HLM~\cite{icml19:HLM19}, \SWMJ~\cite{NeurIPS20:SWMJ20},
\WRK~\cite{iclr20:WRK20}, \WXW~\cite{NeurIPS20:WXW20} and
\WZYBMG~\cite{iclr20:WZYBMG20}; and two versions of \SIEKM~\cite{NeurIPS20:SIEKM20}, 
pre-established and publicly available adversarially trained models on  
\Imagenet.

\paragraph{Experiment setup}
\label{sec:setup:setup}

Croce et al.\ found that \pgd{} attacks find more adversarial examples the more
iterations they run~\cite{icml20:autopgd}. \Autopgd declares the number of
iterations as its only parameter. In this work, we ran all attacks for 100
iterations---the default configuration of \Autopgd~\cite{icml20:autopgd}---to fairly
compare the attack methods.  
For CIFAR10, we measured the success rate against seven defenses, using the same target, 20 random
initial perturbations, and two $\epsilon$ values per image, for
a total of 280 sets of 10,000 attack attempts.
For 
\Imagenet, we used five random initializations, five targets,
and two
$\epsilon$ values per image, thus resulting in 50 sets of 50,000 attack attempts.

\section{Evaluation Results}
\label{sec:result}

In this section, following the setup described in  \secref{sec:setup},
we compare \Autopgd using our \MDabbrv loss with \Autopgd using
previously established loss functions and also
our \cgd attack with \Autopgd. We first report on raw results
(\secref{sec:single}) and then on the statistical tests we performed (\secref{sec:stats})
to demonstrate that \cgd
outperformed the previously best \Autopgd with statistical
significance. We also compare the time cost of attacks (\secref{sec:result:time}) and discuss the uniqueness of the adversarial examples generated (\secref{sec:result:uniqueness}).

\subsection{Raw Results}
\label{sec:single}
\begin{figure}[ht!]
\centerline{\includegraphics[width=0.95\columnwidth]{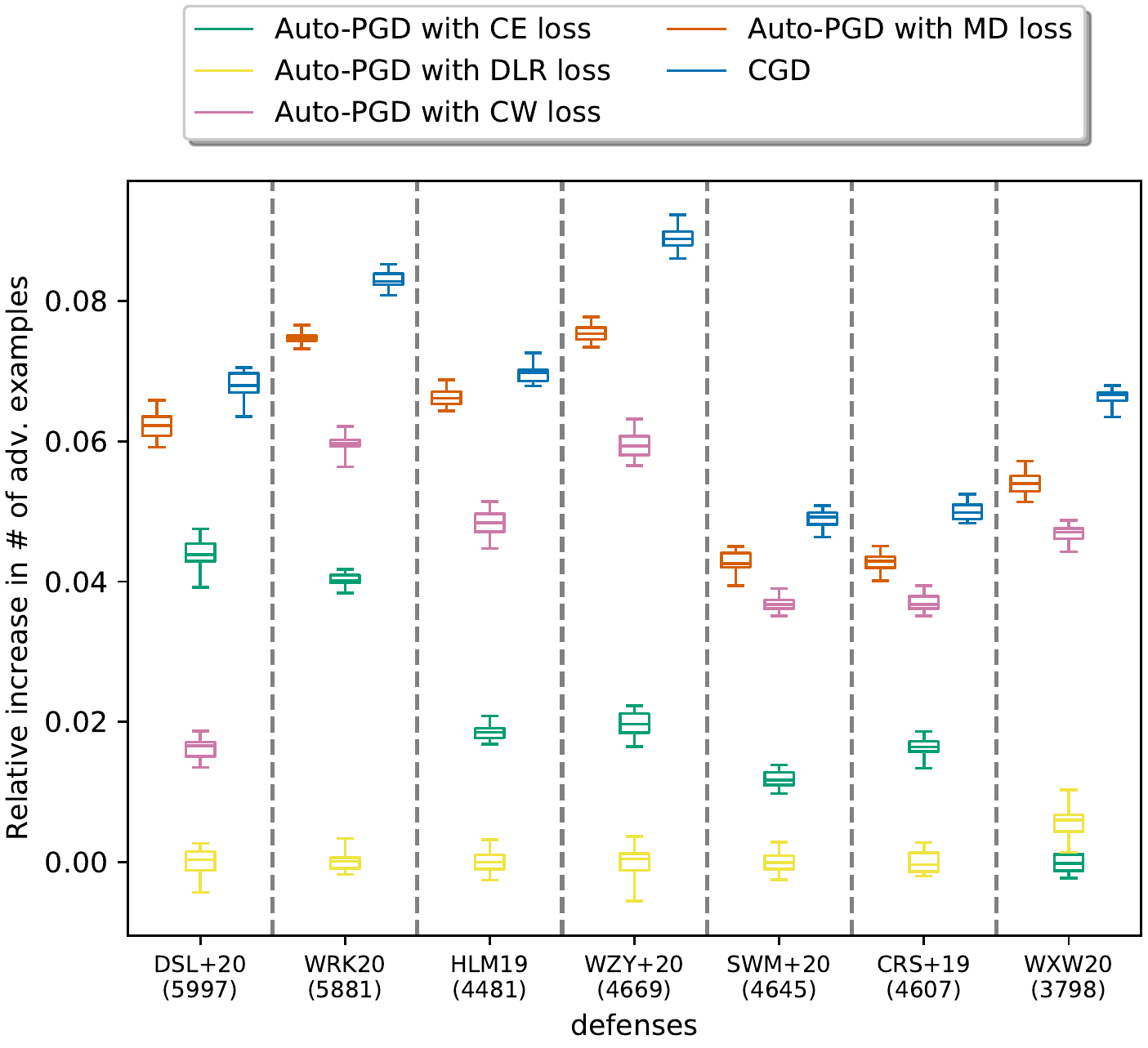}}
\caption{The relative improvement in the number of adversarial examples found by
  attacks on different defenses compared to the worst-performing
  attack. Experiments were performed using $10{,}000$ images from the test set
  of CIFAR10, $\epsilon=16/255$, 20 different random initial perturbations, and a fixed random
  target offset. The worst-performing attack for each defense (with a median of
  0) was selected as the baseline. The y-axis denotes the improvement
  compared to the average performance of the baseline. For example, 0.08 on the y-axis 
  indicates 8\% more adversarial examples found compared to the baseline. The number in parentheses under each defense is the average number of times which the
baseline succeeded out of 10,000 attempts.}

\label{fig:boxperformanceCIFAR16}
\end{figure}

As described in \secref{sec:setup:setup}, we made 280 sets of 10,000 attack attempts on CIFAR10 and 
50 sets of 50,000 attack attempts on \Imagenet .
We compared our two improvements, \Autopgd using our \MDabbrv loss (\secref{sec:loss})
and \cgd (\secref{sec:cgd}), to \Autopgd using three
pre-established loss functions: CE loss, \dlr loss, \cw loss. Implementations of \Autopgd with CE loss 
and \dlr loss are the ones published by the authors~\cite{icml20:autopgd}. We
performed this comparison on two datasets and multiple defenses and
values of $\epsilon$ (see \secref{sec:setup}).

On average, \Autopgd 
with \MDabbrv loss found more adversarial examples than \Autopgd with
any of the three other loss functions
, thus demonstrating the benefits of the \MDabbrv{} loss compared to previous conventional loss functions.
Additionally, \cgd performed better than \Autopgd with \MDabbrv loss, further demonstrating the advantages of the \cgd{} attack strategy to satisfy the \lp{p}-bound constraints compared to ad hoc clipping. 
 The ranking of attacks other than \Autopgd with
\MDabbrv loss and \cgd varied slightly depending on the defense.

\figref{fig:boxperformanceCIFAR16} shows the relative
number of adversarial examples these attack methods found on CIFAR10 with 
$\epsilon=16/255$. 
Among the previously proposed attacks (\Autopgd with one of CE, \dlr,
or \cw loss), which performed least well and which performed best varied by defense.
The best attacks were on average 1.6--6.0\% better than the baseline
attacks.
\Autopgd with \MDabbrv loss was on average 0.6--4.5\% better than
the best performing \Autopgd that did not use \MDabbrv loss;
and \cgd was on average 1.2--5.1\% better than
the best performing \Autopgd that did not use \MDabbrv loss.
\cgd outperformed  \Autopgd with \MDabbrv loss in 139 out of 140 sets of attempts and outperformed 
\Autopgd (with any other loss function) in all of the sets of attempts. Depending on the defense, the
  ranking of the other attack methods changed, but \cgd and \Autopgd
  with \MDabbrv loss always performed better
  than other attack methods.
We observed similar results with different
values of $\epsilon$ and when using the \Imagenet dataset, as shown in
\appref{sec:appendix:performance}. When evaluated against \SIEKM on
the \Imagenet dataset with $\epsilon=4/255$, \cgd was 11.0\% better
than \Autopgd with \cw loss (the previous best \Autopgd).
When using $\epsilon=8/255$, \cgd was 13.6\% better than \Autopgd with \cw loss
against \SIEKM on \Imagenet.

\subsection{Statistical Analysis}
\label{sec:stats}

We also performed statistical analysis to compare the performances of different attack methods. We 
defined a variable $\CondtionalSuccess_i$ for an image $i$ as the total number of successful 
perturbations 
made by an attack with all random initial perturbations, and all the random 
$\TargetOffset$s
we 
tried given the specific dataset, defense, and $\epsilon$.  $\CondtionalSuccess_i$ $\in [0,20]$  on CIFAR10 
while $\CondtionalSuccess_i$ $\in [0,25]$ on \Imagenet. Each $\CondtionalSuccess_i$ is independent. 
We used the Wilcoxon signed rank test to compare
$\CondtionalSuccess_i$ of CGD and \Autopgd with 
MD loss.

We performed the one-sided Wilcoxon signed rank test \cite{stat45:Wilcoxon} with 
the null hypotheses that \Autopgd with MD loss had
equal or better performance than \cgd for each combination of
$\epsilon$, dataset, and defense that we tried.
Overall, we conducted 16 statistical tests, for the 16 different combinations
we had. To account for
the multiple tests, we used Bonferroni correction to adjust the confidence
level $\alpha$ to $.05/16=0.003125$.  We found the $p$-values are below $\alpha$ in 11 out
of 16 tests. Namely, \cgd performed statistically significantly better
than \Autopgd with MD loss across 11 combinations of $\epsilon$,
dataset, and defense that we tried.

We performed a similar one-sided Wilcoxon signed rank test \cite{stat45:Wilcoxon} with the null
hypotheses that the best performing attack among \Autopgd using the \dlr
loss, CW loss, and CE loss performed equal to or better than 
\cgd in each combination of value of $\epsilon$, dataset, and defense that we tried. 
Again, we used the adjusted normal approximation of
the test statistic and Bonferroni corrections. 
All the $p$-values were
far smaller than  $\alpha=0.003125$,
and so we reject the null
hypotheses in all 16 cases, hence demonstrating that \cgd
significantly outperformed \Autopgd with the losses proposed in prior work across each combination of 
$\epsilon$, dataset, and defense that we tried. More details of these tests can be found in 
\appref{app:stat}.

\subsection{Time Complexity}
\label{sec:result:time}

We ran all attacks for 100 iterations, as described in \secref{sec:setup:setup}, and conducted
30 and 100 time measurements per attack-defense pair for the CIFAR10 and \Imagenet datasets, respectively. The results are shown in \tabref{time}. 

\begin{table}[ht!]
\small
\caption{The average time in seconds used to perturb batches of 512 images from CIFAR10 or 10 images from 
\Imagenet, using NVIDIA GeForce RTX 3090 GPUs.
  There are two versions of \SIEKM, trained with 
  $\epsilon=4/255$ (\SIEKM-4) and $\epsilon=8/255$ (\SIEKM-8).
}
\begin{center}
\setlength\tabcolsep{1pt}
\begin{tabular}{|r|r|r@{\hspace{2mm}}r@{\hspace{2mm}}r@{\hspace{2mm}}r@{\hspace{2mm}}r|}
\hline
\multicolumn{2}{|c|}{\textit{attack methods}}& Auto &Auto&Auto&Auto &\cgd\\
\multicolumn{2}{|c|}{}& -PGD &-PGD&-PGD&-PGD &\\
\multicolumn{2}{|c|}{\textit{loss}}& CE &\cw &\dlr& \MDabbrv&\\

\textit{dataset} &\textit{defense} &\multicolumn{5}{c|}{\textit{time (seconds)}}\\
\hline
&\DSLH&26.4&26.9&26.9&26.8&22.0\\
&\WRK&17.0&17.7&17.6&17.5&14.5\\
&\HLM&115.8&117.0&115.9&116.7&100.9\\
CIFAR10&\WZYBMG&116.5&116.6&117.1&116.2&103.3\\
&\SWMJ&115.0&115.2&115.6&114.8&100.1\\
&\CRSLD&116.2&116.7&116.7&116.3&101.5\\
&\WXW&116.4&116.7&116.4&116.3&102.6\\
\hline
\Imagenet&\SIEKM-4&7.7&7.7&7.8&7.8&6.4\\
&\SIEKM-8&7.7&7.7&7.8&7.8&6.4\\
\hline
\end{tabular}
\end{center}
\label{time}
\end{table}

We found that \cgd was on average faster than all \Autopgd attacks against all defenses by 11.41--18.76\%.
We confirmed  this relationship with 
one-sided Wilcoxon tests (\appref{app:wilcoxon}). As we have nine defenses and four baseline attacks, we used 
Bonferroni correction to adjust the confidence
level $\alpha$ to $.05/36=0.0014$.

\subsection{Uniqueness of Attacks}
\label{sec:result:uniqueness}
\begin{figure}[ht!]
\centerline{\includegraphics[width=0.95\columnwidth]{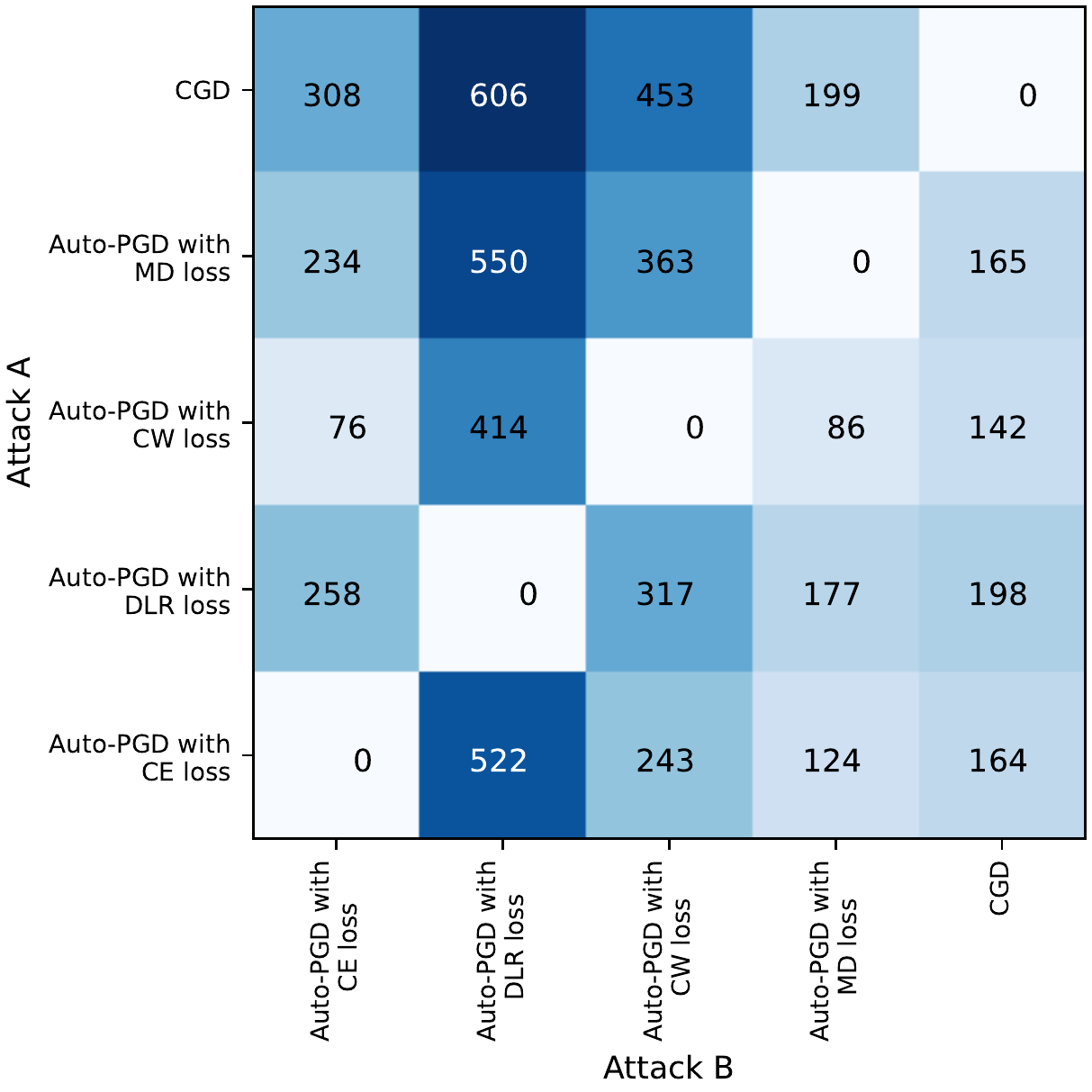}}
\caption{The average number of  successful adversarial examples found by attack A 
but not by attack B in each set of 10,000 attempts on the testing set of CIFAR10 using 
$\epsilon=16/255$ 
against the \DSLH~\cite{iclr20:DSLH20} defense, rounded to whole numbers.}
\label{1v1DSLH16}
\end{figure}

\begin{figure}[ht!]
\centerline{\includegraphics[width=0.95\columnwidth]{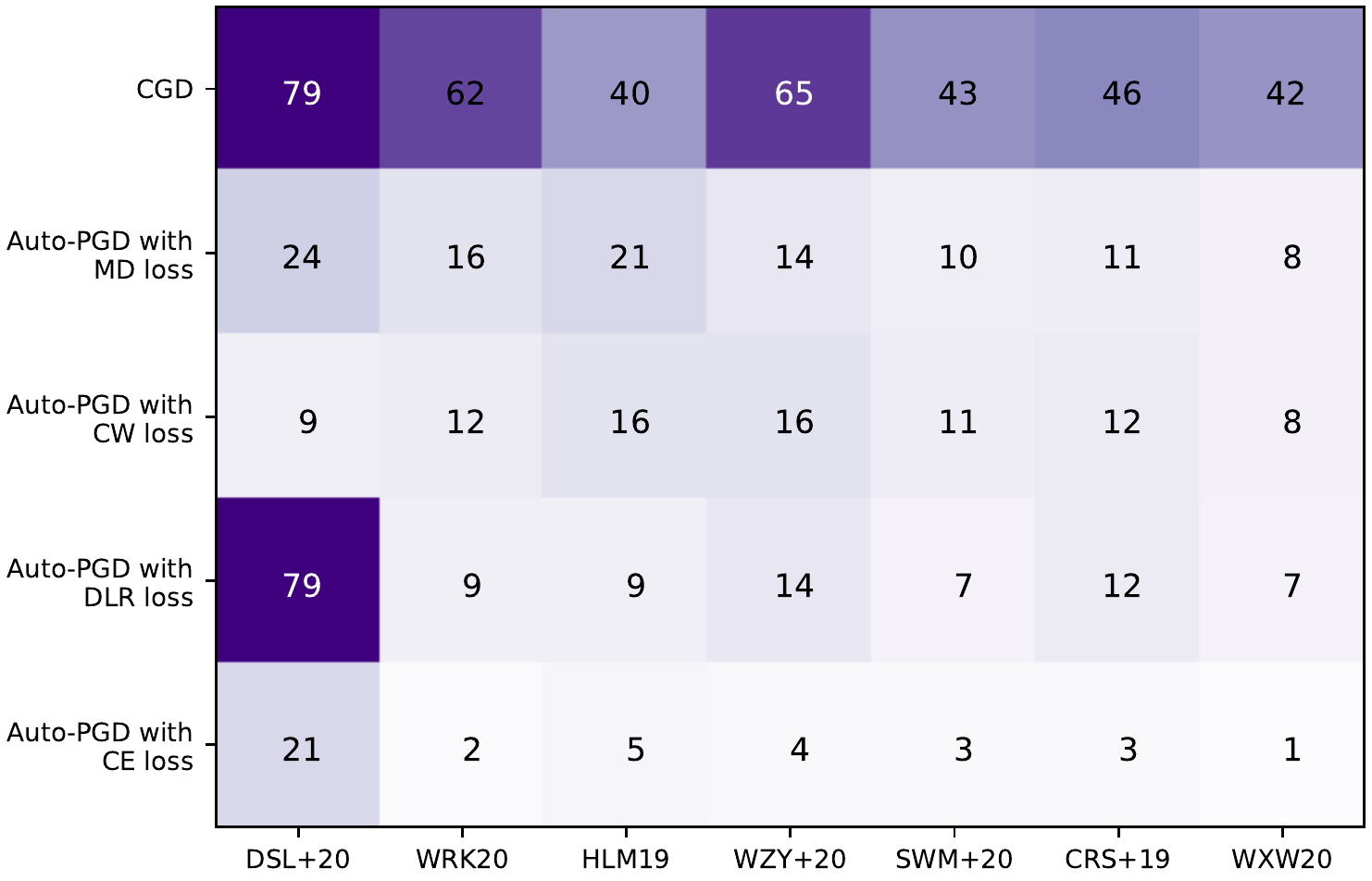}}
\caption{The average number of successful adversarial examples found by an attack, but not by 
any of other attack methods, against each defense  in each set of 10,000 attempts on the testing set 
of CIFAR10 using $\epsilon=16/255$, rounded to whole numbers.}
\label{CIFAR10-16-1VSothers}
\end{figure}

As we observed in \secrefs{sec:single}{sec:stats} that \cgd found more adversarial examples than other attack 
methods did in most of the sets of attempts, we wondered if the adversarial examples other 
attacks found could be a subset of those \cgd found.
However, we discovered that among the attack methods we
tried, each of them found a slightly different set of successful
adversarial examples, as shown in 
\figref{1v1DSLH16}.
In addition, each attack succeeded in finding an adversarial example for some specific input for which no other attack succeeded, 
as shown in 
\figref{CIFAR10-16-1VSothers}. We observed the same phenomenon 
across attack methods, values of $\epsilon$, defenses, and datasets. More details can be found in \appref{app:result:uniqueness}.

\begin{figure}[ht!]
\centerline{\includegraphics[width=0.95\columnwidth]{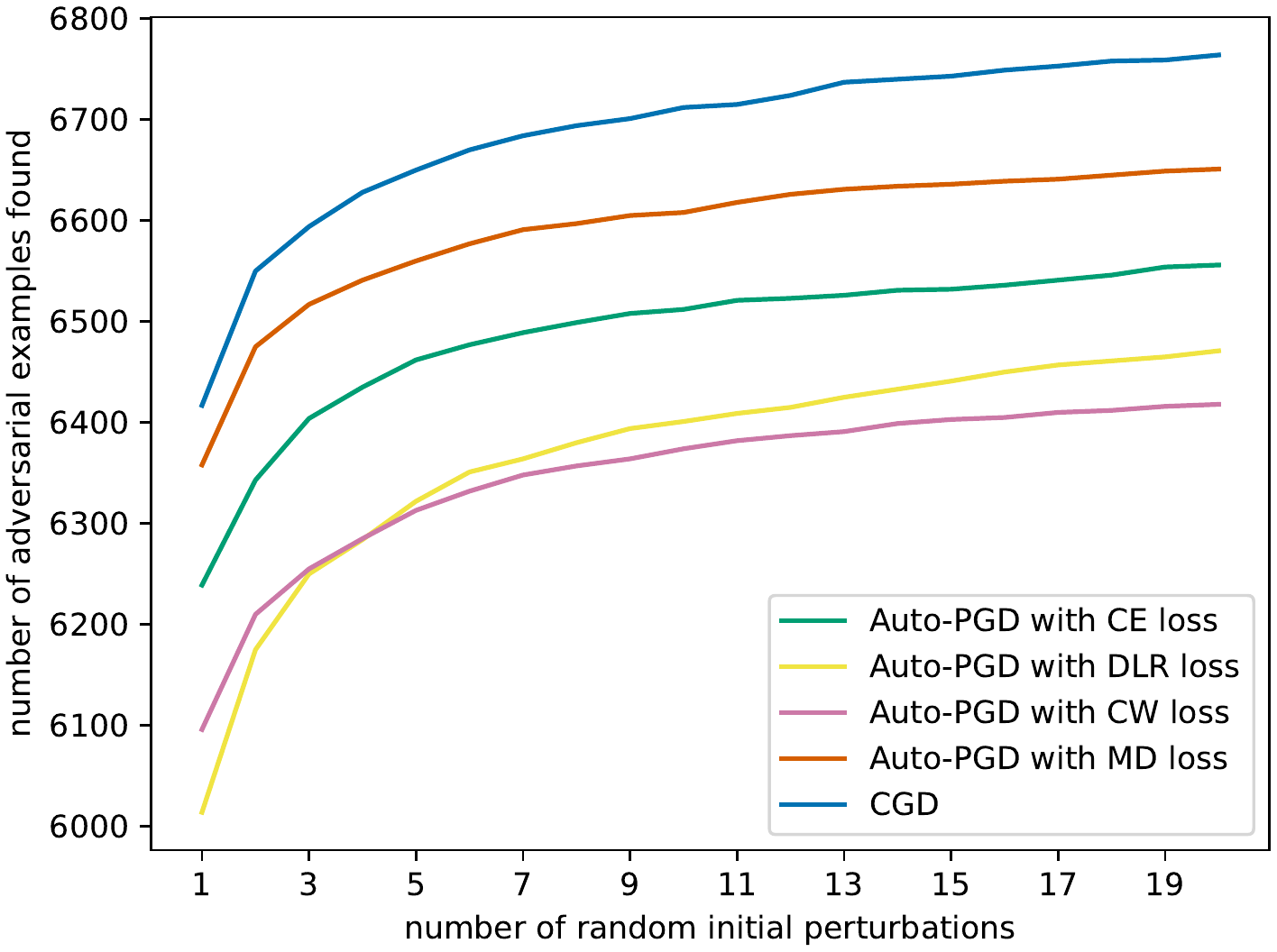}}
\caption{The image shows the number of adversarial examples each of the attacks found when using the specified number of random initial perturbations,
 on all 10,000 images from the testing set of CIFAR10, with $\epsilon=16/255$, against the 
\DSLH~\cite{iclr20:DSLH20} defense. }
\label{CIFAR10-DSLH-seeds16}
\end{figure}

At the same time, when the same attack method is given
different random initial perturbations, the attack always found a slightly different set of
adversarial examples, within the 20 random initial perturbations
we tried (an example is shown in \figref{CIFAR10-DSLH-seeds16}).  We observed the same
phenomenon across attack methods, values of $\epsilon$, and defenses on CIFAR10. More details can be found in \appref{app:result:uniqueness}.

\section{Discussion}
\label{sec:discussion}

Here we discuss the potential to use \cgd as a \emph{framework}
for attacks (\secref{sec:framework}), e.g., with different distance
metrics and loss functions. We demonstrate one such use, where
we instantiate \cgd for untargeted attacks (\secref{sec:discussion:untargeted}).

\subsection{CGD as a Framework}
\label{sec:framework}

In \secref{sec:result} we showed that CGD outperformed the previous best
attack in targeted tasks, with statistical
significance and substantial effect size.
Similarly to how Auto-PGD is a member of the PGD family attacks, the specific attack we explore in this paper could be viewed as member of a broader CGD
family. Auto-PGD improves the performance of PGD attacks by using
alternative loss functions and wisely tuning parameters; similar
tweaking could also apply to CGD attacks: using
alternative loss functions and wisely tuning parameters could yield a
stronger attack within the CGD family. 
Meanwhile, as the loss function of \cgd can be separated into two components, 
$\LossBoundary$ 
and $\LossClassification$, each of those could be tuned. 
We demonstrated
that CGD implemented with specific loss functions and parameters
outperformed the previous best attack; other variants of CGD could perform
better yet.
CGD might also be extended to other distance metrics besides
\lp{\infty} and also to untargeted attacks. In general, our
work opens the door for future work to find stronger attacks
using CGD as a framework. 

\subsection {Applying CGD to Untargeted Attacks}
\label{sec:discussion:untargeted}
We demonstrated that CGD outperformed the previous best \whitebox
targeted attacks in \secref{sec:result}. Here we explore an
untargeted variant of CGD as an
example of extending CGD to other types of attacks, and, specifically, how other loss functions can be incorporated in the general CGD approach. 

Previous works find untargeted adversarial examples by using one of several loss functions.
One example is
cross entropy loss, which we described in
\secref{sec:bg:attackergoals}: $L_{\mathit{CE}}=-\log P_y=-{Z_y}
+\log(\sum_{j=1}^K
e^{Z_j})$. Other work
proposed the untargeted attack version of the Difference of
Logits Ratio loss (DLR loss):
\begin{equation}
L_{\mathit{\dlr}}=\frac{Z_y-max_{i\neq y} Z_i}{Z_{\pi1}-Z_{\pi3}} 
\end{equation} and showed that \Autopgd with $L_{\mathit{\dlr}}$ finds more adversarial examples than 
\Autopgd with $L_{\mathit{CE}}$~\cite{icml20:autopgd}.
Yet other work uses the untargeted version of \cw loss~\cite{Oakland17:CarliniWagner}:
\begin{equation}
L_{\mathit{\cw}}=-Z_y+max_{i\neq y} Z_i
\end{equation}
Previous works show that by iteratively maximizing these loss functions the
adversary can find untargeted adversarial
examples~\cite{Oakland17:CarliniWagner,icml20:autopgd}.

Because the concept of staying within the \lp{\infty} distance limit is the same in targeted and 
untargeted attacks, the loss component for capturing the task of staying within the \lp{\infty} distance limit, 
$\LossClassification$, 
is the same as described in \secref{sec:cgd}. $\LossClassification$ decreases to 0 if and only if 
the adversarial 
example is within the \lp{\infty} distance limit. Hence, we design the loss component that captures 
the task of forcing misclassification, such as the \MDabbrv loss, to also decrease to 0 when the
  adversary succeeds. The purpose of this design is to have the
  overall loss function, which is a weighted sum of the two 
components, decrease to 0 if and only both tasks have been successfully completed.
To create the loss component that captures 
the task of avoiding correct classification, we define a variant
of the \cw loss as follows:
\begin{equation}
L_{\mathit{\cw^{*}}}=ReLU(Z_y+\delta-max_{i\neq y} Z_i)
\end{equation}
where $\delta$ is a minimal value, set to $1\mathrm{e}{-15}$ (see
\secref{sec:loss:newloss}). We replace 
\MDabbrv loss in \algoref{algorithmCGD} with $L_{\mathit{\cw^{*}}}$ as
$\LossClassification$ to obtain the untargeted variant of the
\cgd attack, \cgduntarg.

Minimizing $L_{\mathit{\cw^{*}}}$ achieves the same result as
maximizing $L_{\mathit{\cw}}$: when $Z_y \ge max_{i\neq y} Z_i$,
both loss functions minimize $Z_y-max_{i\neq y} Z_i$;
and when $Z_y<max_{i\neq y} Z_i$, the untargeted attack has already
succeeded. Hence, an \Autopgd that minimizes $L_{\mathit{\cw^{*}}}$ and an \Autopgd that maximizes  
$L_{\mathit{\cw}}$ behave exactly the same on each image; we report results only for the former. 

We ran \Autopgd with $L_{\mathit{\dlr}}$, \Autopgd with $L_{\mathit{\cw^{*}}}$, 
and
\cgduntarg to compare their performance. Details of the setup can be found in \appref{app:UntargetedSetup}.
On average, \cgduntarg outperformed \Autopgd with $L_{\mathit{\dlr}}$ and \Autopgd
with $L_{\mathit{\cw^{*}}}$.
\cgduntarg outperformed the next best method, \Autopgd with
$L_{\mathit{\cw^{*}}}$ in 28 out of the 35 sets of attempts at $\epsilon=4/255$,
 in 31 out of the 35 sets of attempts at $\epsilon=8/255$, 
 and in all 35 sets of attempts at $\epsilon=16/255$. 
 More details can be found in \appref{sec:appendix:untargeted}.

\section{Conclusion}
\label{sec:conc}
In this work we improved a previously established \whitebox, targeted evasion attack by using a 
new loss function. We also proposed a yet stronger attack that learns to approach and explore the $\epsilon$-boundary.
We demonstrated the efficacy of both the new loss function and the new
attack on two datasets (CIFAR10 and ImageNet), for multiple values of $\epsilon$, and against
multiple defenses; in all cases, our methods outperformed the best of
the attacks we compared against, finding targeted adversarial examples more successfully while taking significantly less time to run.
Finally, we showed how to use our new attack method as a general
framework for attacks and demonstrated its utility by instantiating it into a stronger \emph{untargeted} attack.

\section*{Acknowledgments}

This paper was supported in part 
by the Department of Defense under contract FA8702-15-D-0002;
%
by NSF grants 1801391, 2112562, and 2113345;
by the National Security Agency under award H9823018D0008;
by the Maof prize for excellent young faculty;
and by Len Blavatnik and the Blavatnik Family foundation.

\bibliographystyle{icml2022}
\bibliography{references.bib}
\newpage
\appendix
\label{sec:appendix}

\section{The CGD algorithm}
\label{app:cgdalgo}
\begin{algorithm}[ht!]
   \caption{CGD}
\label{algorithmCGD}
\begin{algorithmic}
 \STATE {\bfseries Input:} $x$, $model(\cdot)$, $n_{iterations}$, $\epsilon$, $threshold$, $checkpoints$, 
$target$
 \end{algorithmic}
\begin{algorithmic}[1]
  
    \STATE $x’\leftarrow clip(x+2*rand(x.shape)-1)$\;
   \STATE $w \leftarrow 0.1$\;
   \STATE $\UpperBoundary{i,j,k} \leftarrow min(1,x_{i,j,k}+\epsilon)$\;
   \STATE $\LowerBoundary{i,j,k} \leftarrow max(0,x_{i,j,k}-\epsilon)$\;
    \FOR{$iteration \leftarrow 1$ {\bfseries to} $n_{iterations}$~}
    \IF{$iteration$ $\in$ $checkpoints$}
        \STATE       $threshold \leftarrow threshold/2$\;
    \ENDIF
    \STATE $Z \leftarrow model(x')$\;
    \STATE $\LossClassification \leftarrow \sum_i ReLU(Z_i+\delta-Z_t)$\;
    \STATE$\Overrun_{i,j,k}(x')\leftarrow 
ReLU(x'_{i,j,k}-\UpperBoundary{i,j,k})+ReLU(\LowerBoundary{i,j,k}-x'_{i,j,k})$\;
    \STATE$\LossBoundary \leftarrow \sum_{i,j,k} (\Overrun_{i,j,k}(x'){}^2)$\;
    \IF{$\exists i,j,k \in \Overrun_{i,j,k}(x') > threshold$}
        \STATE    $w \leftarrow w/2$\;
    \ENDIF    
    \STATE $loss \leftarrow w*\LossClassification+(1-w)*\LossBoundary$\;
    \STATE $gradients \leftarrow \frac{\partial loss }{\partial x'}$\;
        
    \IF{$iteration==1$}
        \STATE$x' \leftarrow clip(x'-\epsilon*sign(gradients))$
     \ELSE
        \STATE $changes \leftarrow Adamoptimizer(gradients)$\;
        \STATE $x' \leftarrow x'-changes$\;
    \ENDIF
        
    \hspace*{-\fboxsep}\colorbox{pink}{\parbox{180 pt}{
    \STATE    $x_{test} \leftarrow clip(round(x'*255)/255)$ \;
    \IF{$argmax(model(x_{test}))==target$}
            \STATE {\bfseries Return}{$x_{test}$}\;
    \ENDIF    
    }}
    \ENDFOR
   
    {\bfseries Return} $\mathit{failed~attack}$\;
\end{algorithmic}
\end{algorithm}

\algoref{algorithmCGD} shows the pseudocode of the \textit{Constrained Gradient
Descent} (\cgd) algorithm.
The inputs to the algorithm are: the benign 
example $x$; the $model(\cdot)$ function that
emits the logits for given inputs; the number of iterations $n_{iterations}$; the $\epsilon$ distance; a set 
of constants tuned per model, $threshold$ and $checkpoints$; and the target class $target$.  We allow 
the tuning of $threshold$ and $checkpoints$ per model because we are running
\whitebox attacks and hence the adversary has the freedom to choose the attack's
parameters per model. All other parameters do not require tuning per model.
In line 1, we add a random initial perturbation to the benign sample, moving it to the $\epsilon$ 
boundary.
In lines 3 and 4 of the algorithm, before we start the iterations, we compute $\UpperBoundary{}$ and 
$\LowerBoundary{}$. 
From line 6 to 8, we adjust the threshold based on pre-set constants. We 
compute the logits $Z$ of the model regarding the current perturbation $x'$ in line 9 and 
compute the MD loss in line 10. Then we compute the $\Overrun$ in line 11 and the 
$\LossBoundary$ 
in line 12, which motivates the adversarial example to move to be within the \lp{\infty} distance limit. 
From line 13 to line 15, we adjust the weight $w$, and then use this weight to 
compute the total loss as a weighted sum in line 16. We compute the gradients from the loss function in 
line 17 and apply changes from line 18 to line 23. In the first iteration, the candidate perturbation is 
pushed 
by a distance of $\epsilon$ to more quickly reach the boundary. In later iterations, the 
Adam optimizer is used to reduce fluctuation. 
From line 
24 to 27 (highlighted in
pink), we perform a  normal rounding check.  This rounding check enforces the successful adversarial 
example to be both inside the \lp{\infty} distance limit and in 8-bit RGB format.
When the candidate perturbation is very close to the
$\epsilon$ boundary, clipping has minimal
effects on the logits, $\LossClassification$, and probabilities
assigned to classes, meaning that an image just outside the $\epsilon$
boundry is rounded to become both within the $\epsilon$ boundary and
potentially a successful attack.

\section{Wilcoxon Signed-Rank Test}
\label{app:wilcoxon}

The Wilcoxon signed-rank test~\cite{stat45:Wilcoxon}, is a non-parametric test to examine the 
relationship between two related paired samples. 
In our experiements, we ran one-sided Wilcoxon signed-rank tests to verify if one set of samples is greater 
than another with statistical significance.
While the Wilcoxon signed-rank test assumes that the two sets differ
on each pair of samples, in our case, the samples are equal in some pairs.
Previous work suggests a mitigation to enable using the Wilcoxon signed-rank test
in such cases: using an adjusted
normal approximation of the test statistic of the Wilcoxon signed-rank
test instead of using the standard test statistic to compute the
$p$-value~\cite{ASA67:Normal}, and using the pairs that are equal to produce the rankings 
but excluding them afterwards~\cite{stat59:Pratt}.
This approach requires
more than 25 independent trials; our data satisfiy this requirement.
In addition, to account for
the multiple tests, we always used Bonferroni correction to adjust the confidence
level $\alpha$.

\section{Measurement of Changes in Predictions}
\label{app:fluctuation}

\begin{figure}[ht!]
  \centering
  \includegraphics[width=1.0\columnwidth]{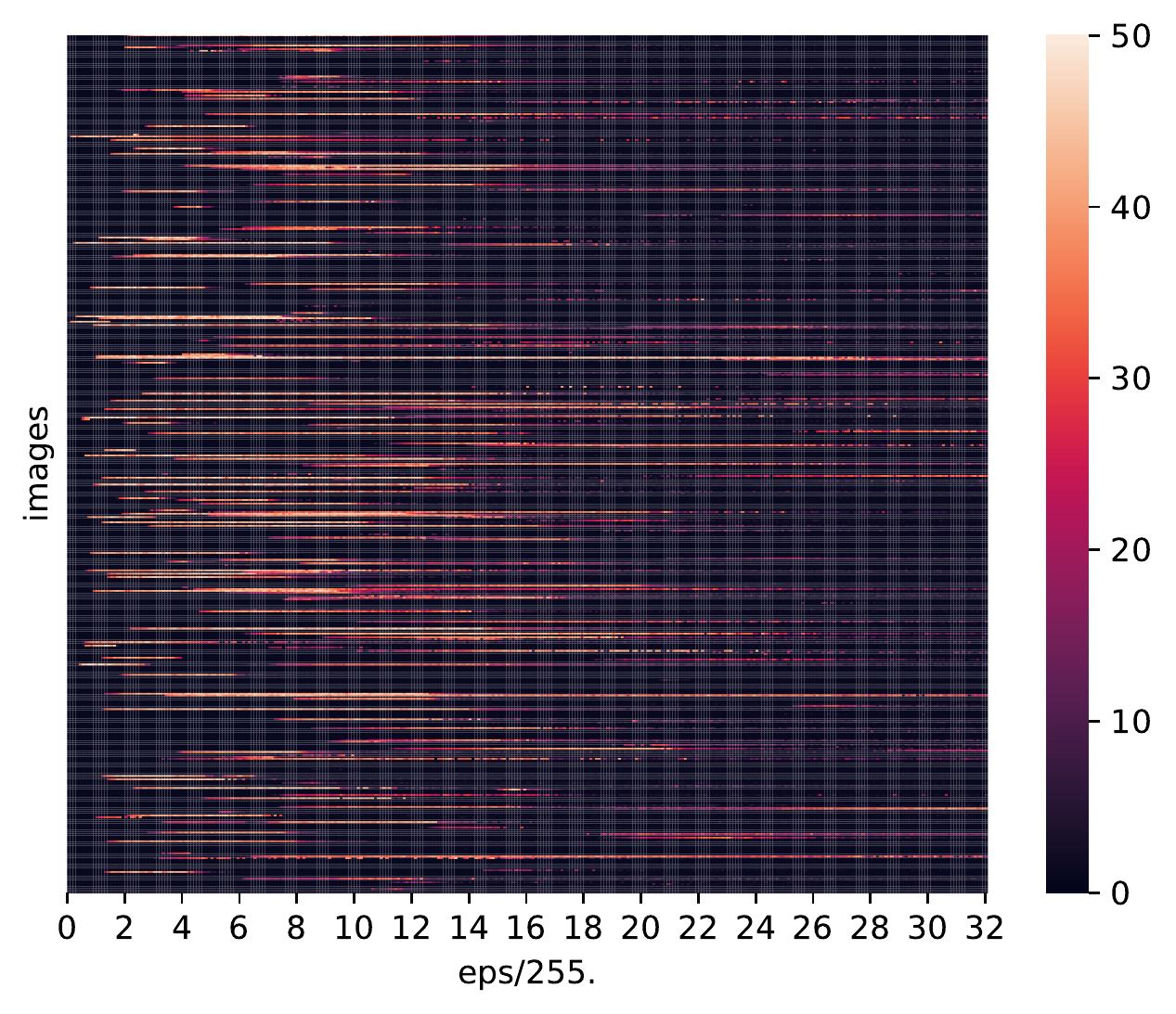}
\caption{The maximum number of times that the prediction was changed to a class, 
throughout the process of perturbing each of the 512 image before success using \Autopgd with CW 
loss against \DSLH. }
\label{cwDSL+20FLUC}
\end{figure}

\begin{figure}[ht!]
  \centering
  \includegraphics[width=1.0\columnwidth]{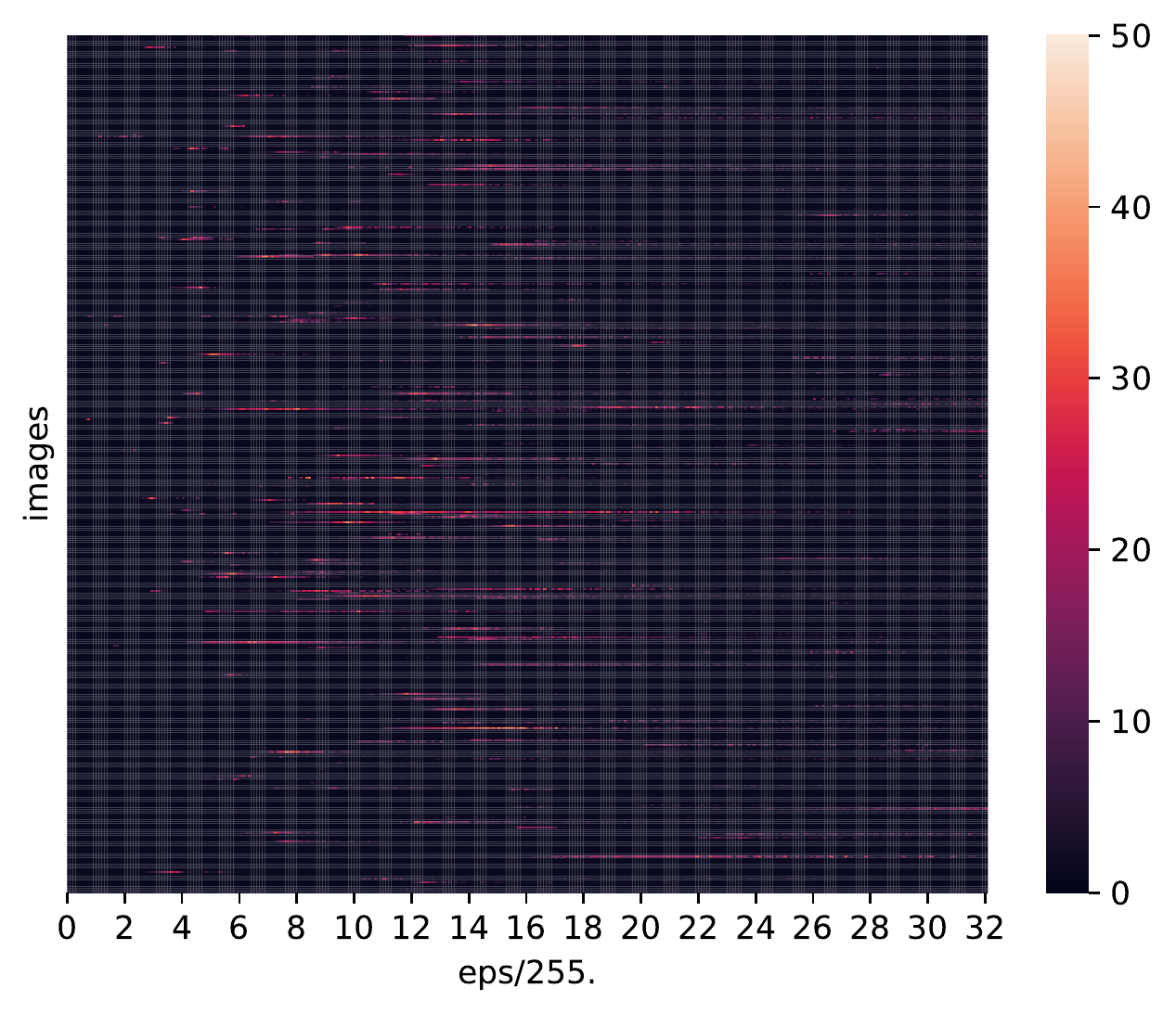}
\caption{The maximum number of times that the prediction was changed to a class, 
throughout the process of perturbing each of the 512 image before success using \Autopgd with 
\MDabbrv loss against \DSLH. }
\label{wrDSL+20FLUC}
\end{figure}

As we described in \secref{sec:loss}, we used the number of times that the prediction was changed to a certain class before the perturbation first succeeds as a metric to capture how often the undesirable behavior (mentioned in \secref{ssec:previous-loss}) of the \cw loss occurs. 
We used 512 images pertaining to a diversity of classes from CIFAR10, and
perturbed them while targeting randomly picked classes. We ran \Autopgd with $\epsilon$ 
values from $0.1/255$ to $32/255$ by every $0.1/255$. The results for \Autopgd with the \cw loss are 
shown in \figrefs{cwDSL+20FLUC}{cwWRK20FLUC}, while the
results for \Autopgd with the \MDabbrv loss are shown in \figrefs{wrDSL+20FLUC}{wrWRK20FLUC}. Against both \DSLH~\cite{iclr20:DSLH20} and 
\WRK~\cite{iclr20:WRK20}, the \MDabbrv loss reduced the maximum number of times that the prediction 
was changed to a class.
We also ran a Wilcoxon signed-rank test (see \appref{app:wilcoxon}) to determine whether the differences between the losses are statistically significant. Specifically,
we conducted 320 statistical tests on each defense, for the 320 different values of $\epsilon$
we used. To account for
the multiple tests, we used Bonferroni correction to adjust the confidence
level $\alpha$ to $.05/320=0.00015625$. We found that for 203 out of 320 values of $\epsilon$ against
\DSLH and for 309 out of 320 values of $\epsilon$ against \WRK, the maximum
number of times that the prediction was changed to a class is statistical
significantly smaller when using the \MDabbrv loss compared to the \cw loss.
Notably, this maximum number of changes is smaller when 
using the \MDabbrv loss for every value of $\epsilon$ in $[0.8/255,16.3/255]$ against \DSLH, and for
every value of $\epsilon$ in $[0.7/255,27.3/255]$ against \WRK. Both ranges include values 
of $\epsilon$ commonly used in \lp{\infty} attacks.

\begin{figure}[ht!]
  \centering
  \includegraphics[width=1.0\columnwidth]{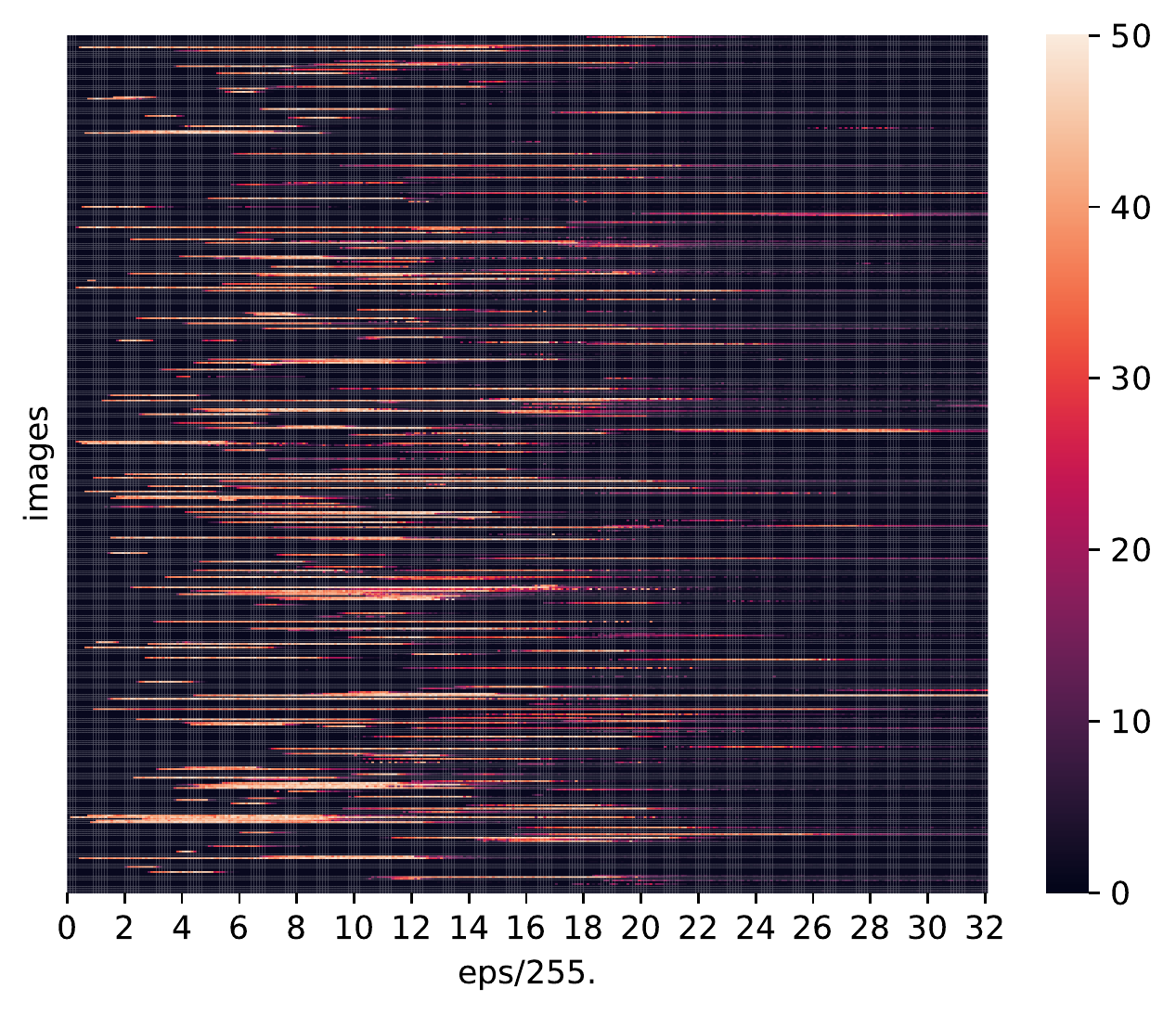}
\caption{The maximum number of times that the prediction was changed to a class, 
throughout the process of perturbing each of the 512 image before success using \Autopgd with CW 
loss against \WRK. }
\label{cwWRK20FLUC}
\end{figure}

\begin{figure}[ht!]
  \centering
  \includegraphics[width=1.0\columnwidth]{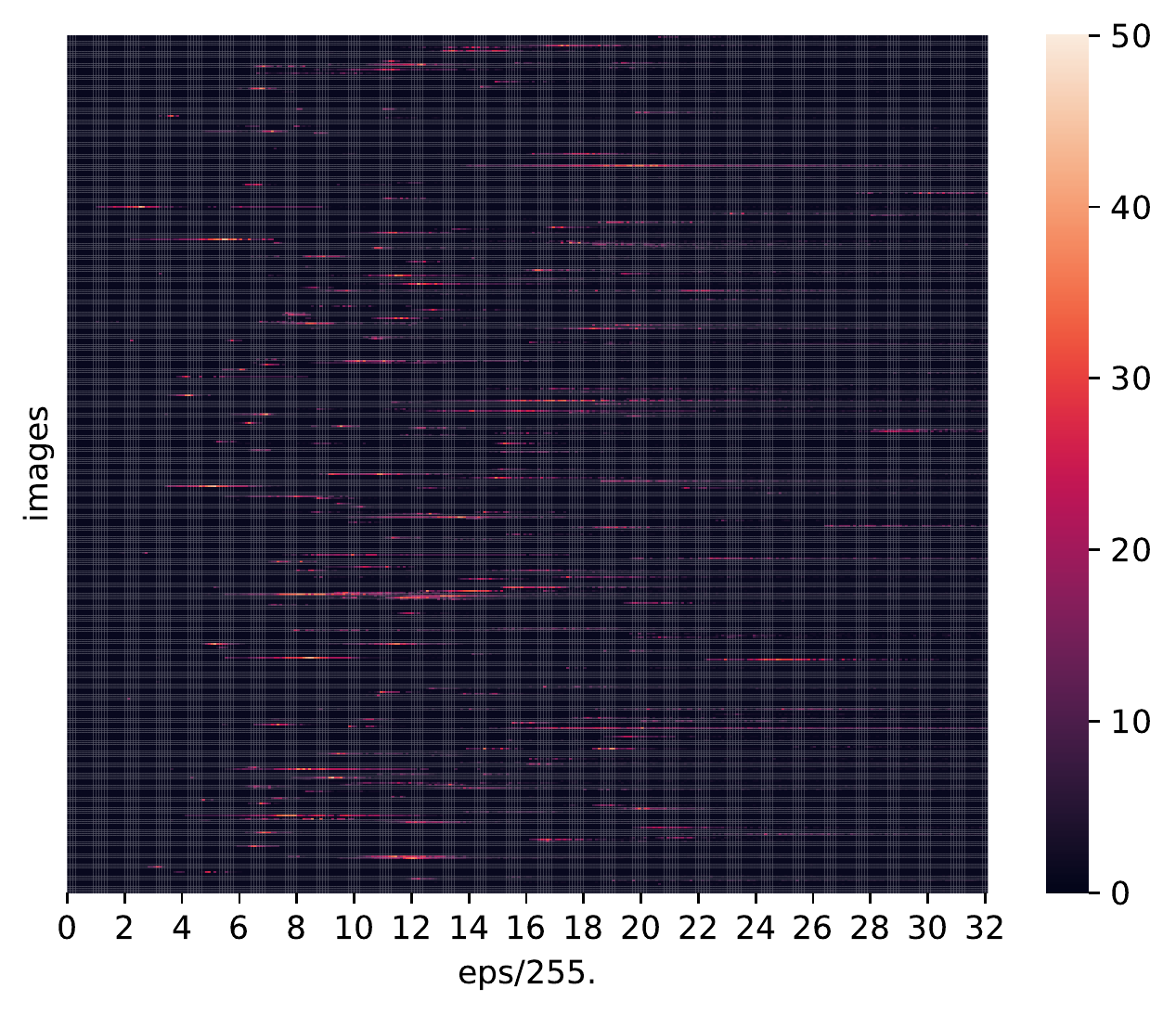}
\caption{The maximum number of times that the prediction was changed to a class, 
throughout the process of perturbing each of the 512 image before success using \Autopgd with 
\MDabbrv loss against \WRK. }
\label{wrWRK20FLUC}
\end{figure}

\section{More Evaluation Setup for Targeted Attacks}
In this section, we provide more detail about how we executed
reproducible attacks to enable meaningful and fair comparisons.
\label{app:setup}
\subsection{Datasets}
\label{app:setup:dataset}
In this work, we used the CIFAR10 and \Imagenet datasets---two standard datasets
that are commonly used for classification tasks. Both
datasets contain colored images of objects in 8-bit RGB format as described in 
\secref{sec:setup:Evaluation}. Each image in CIFAR10 has $32
\times 32$ pixels, whereas we resized each image in \Imagenet to have $224
\times 224$ pixels. We evaluated attacks on the test set of the datasets. CIFAR10
has 10,000 images in its test set and \Imagenet has 50,000.

\subsection{Benchmarks}
\label{app:setup:models}

Many defense strategies have been explored. For example, 
different input transformations have been proposed to remove adversarial
perturbations from inputs prior to classification (e.g.,~\cite{iclr18:JPEG,
  iclr18:DGAN, ndss18:Squeezing, cvpr18:Liao2018DefenseAA}). Unfortunately, 
the majority of these defenses can be evaded by adaptive attacks that craft
adversarial perturbations that survive the
transformations~\cite{icml18:Obfuscated, woot17:ensembles, arxiv18:Athalye2018OnTR}.
Certain defenses attempt to detect adversarial examples
(e.g.,~\cite{arxiv17:Detect, iclr18:LID, iclr17:metzen2017detecting}). The
majority of these defenses are unable to classify adversarial examples
correctly, when detected. Moreover, researchers have also found that detection
methods can often be evaded by adaptive attacks~\cite{icml18:Obfuscated,
  aisec17:detection}. 
Finally, certified defenses provide provable
accuracy guarantees on adversarial examples (e.g.,~\cite{icml19:Provable,
  icml19:CSmooth, oakland19:DPDefense, iclr20:Zhang2020TowardsSA, iclr18:Raghunathan2018CertifiedDA}). However, these defenses are
often effective for perturbations of smaller norms than adversarial
training~\cite{oakland19:DPDefense}, or are ineffective against the threat
models we study (e.g., high-dimensional perturbations with bounded
\lp{\infty}-norms)~\cite{icml20:CSmooth,arxiv20:CompareDefenses}.

As we described in \secref{sec:back:defenses} and \secref{sec:setup:models},
adversarial training is regarded as ne of the strongest
defenses~\cite{arxiv21:Akhtar2021AdvancesIA}, and hence
we used seven pre-established adversarially trained models for
CIFAR10: \CRSLD~\cite{NeurIPS19:CRSLD19}, \DSLH~\cite{iclr20:DSLH20},
\HLM~\cite{icml19:HLM19}, \SWMJ~\cite{NeurIPS20:SWMJ20},
\WRK~\cite{iclr20:WRK20}, \WXW~\cite{NeurIPS20:WXW20} and
\WZYBMG~\cite{iclr20:WZYBMG20}. All of these models were trained with
$\epsilon=8/255$ and are publicly available via
\robustbench~\cite{arxiv21:robustbench}, a standard library for
robustness evaluation of neural networks. We selected the models using the
following criteria: they rank highly on robustness in the public \robustbench
evaluation; they have a \pytorch implementation; and they fit into the 11GB
memory of our NVIDIA GeForce RTX 2080 GPUs. These models have have also been found to
be robust in evaluations other than \robustbench's~\cite{arxiv21:fixing, arxiv20:uncover}.
The seven models can be categorized into four non-exclusive groups: some propose surrogate loss 
functions~\cite{iclr20:DSLH20,NeurIPS20:WXW20,iclr20:WZYBMG20}, some use pre-training 
or semi-supervised learning techniques~\cite{icml19:HLM19,iclr20:WZYBMG20,NeurIPS19:CRSLD19}, 
some try to wisely tune the training process~\cite{iclr20:WRK20}, and some apply 
pruning to their models~\cite{NeurIPS20:SWMJ20}. Similarly to prior work
(e.g.,~\cite{iclr18:PixelDefend,iclr20:WRK20}), we evaluated attacks against
these models with $\epsilon=8/255$ and $16/255$.
We also found two versions of \SIEKM~\cite{NeurIPS20:SIEKM20}, 
pre-established and publicly available adversarially trained models on  
\Imagenet---one adversarially trained with
$\epsilon=4/255$ and another with $\epsilon=8/255$. We 
evaluated attacks against each version of this model with the same $\epsilon$
with which it was trained.

\subsection{Experiment Setup}
\label{app:setup:setup}
We observed that some \pytorch functions could yield unreproducible results in
different runs. To keep our measurement results reproducible,
we set \texttt{torch.backends.cudnn.benchmark} to \texttt{False} and
\texttt{torch.backends.cudnn.deterministic} to \texttt{True}. In addition, we found
that PGD attacks, including Auto-PGD, always start with a random
perturbation and this could slightly influence the result. Hence, we
ran these attacks multiple times, each time with a different random
seed. We also fixed the batch size so that with the same random seed we
got the same random initial perturbation. We used a batch size of 512
images for CIFAR10 and a batch size of 10 images for \Imagenet. In
addition to the random initialization, we also picked a random target
class for each image in the testing sets of datasets which we used to
evaluate attacks. Due to limited computing resources, we were not able
to run attacks targeting every incorrect class, especially for
\Imagenet which has 1,000 classes. The target class was intentionally
selected to be different from the label, the correct class of the
image. We chose the difference $\TargetOffset$ between the target
class and the correct class, using the following formula:
\begin{equation}
\TargetOffset_i=\mathit{floor}(\mathit{rand}()*(N_{classes}-1))+1
\end{equation}
where $i$ is the index of images in the testing
set, $i \in [0,10000)$ for CIFAR10 and $i \in [0,50000)$ for \Imagenet, and
$N_{classes}$ is the number of classes, 10 for CIFAR10 and 1,000 for
\Imagenet. The $\mathit{rand}()$ function generates a float $\in [0,1)$.
The target class was
\begin{equation}
t_i=(y_i+\TargetOffset_i) \bmod N_{classes}
\end{equation}
for $i \in [0,N_{images})$. We only used one random seed (specifically, 0) for
the $\TargetOffset$ in CIFAR10, as on average there are
$10000/(9*10)$ images in each source-target class pair, whereas we
used five random seeds (specifically, 0--4)
for the $\TargetOffset$ of the \Imagenet
dataset, as 50,000 images cannot cover all the 999$\times$1000 source-target
class pairs.  We also observed that some class
pairs in \Imagenet (e.g., ``African crocodile'' and ``American
alligator'') are closer than other class pairs (e.g., ``African crocodile'' and
``thunder snake''), and are significantly more 
easy to perturb into each other.
Random numbers were generated as a vector of length 10,000 for CIFAR10 and 50,000
for \Imagenet. Hence, 
the $\TargetOffset$ was the same for the same image with the same random seed, but $\TargetOffset 
s$  
was not the same across all images when we used the same random seed.

\section{Performance of Targeted Attacks}

\begin{figure}[ht!]
\centerline{\includegraphics[width=0.95\columnwidth]{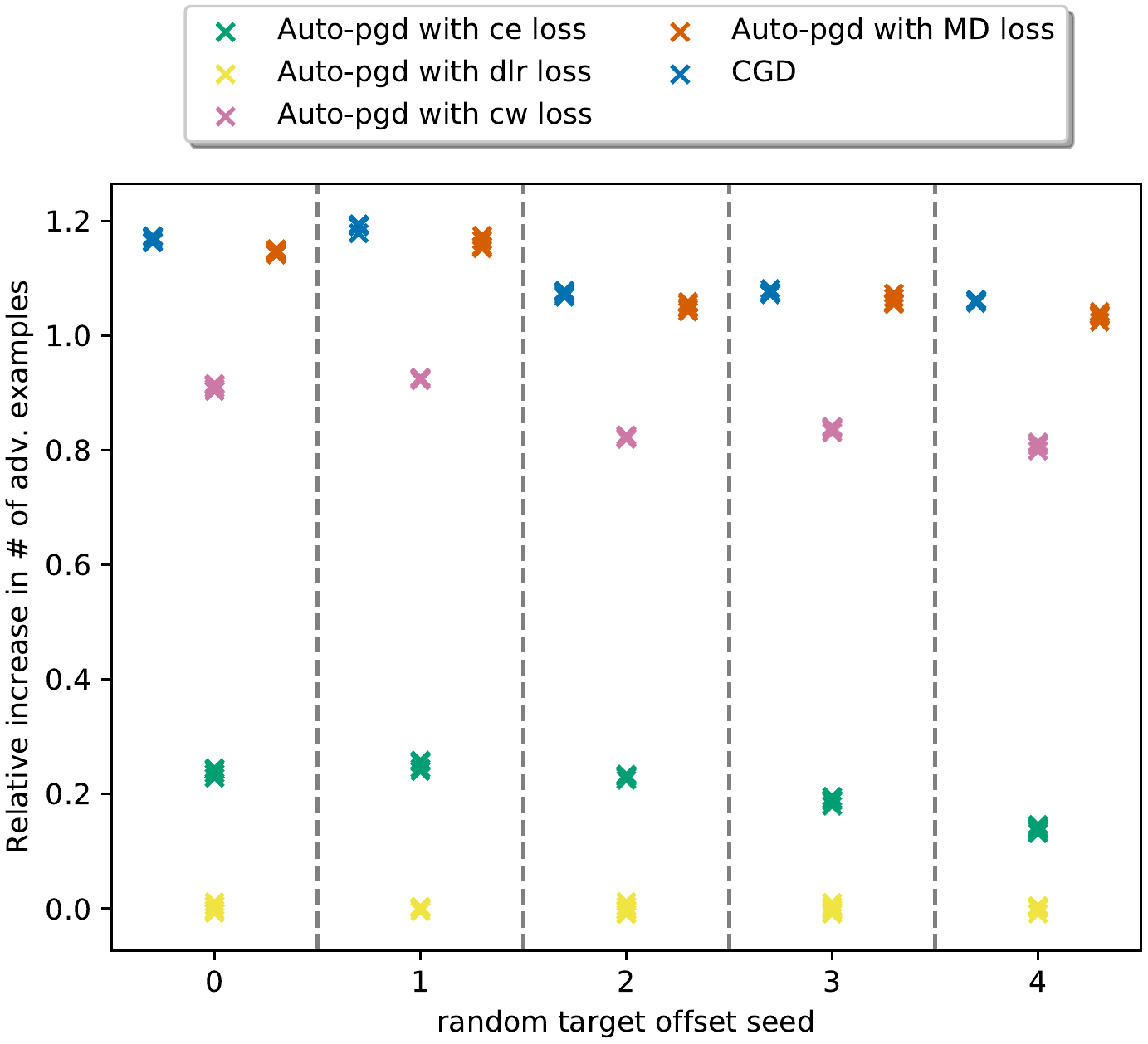}}
\caption{This figure shows the relative improvement in the number of adversarial examples found by
  attacks on different defenses compared to the worst-performing
  attack. Experiments were performed using 50,000 images from the testing set of \Imagenet. We ran 
attacks using $\epsilon=8/255$ against \SIEKM~\cite{NeurIPS20:SIEKM20}, five different random initial 
perturbations with seeds 0--4, and five different random target offsets with seeds 0--4. The result is 
normalized by the mean of the worst performing method for each target offset.}
\label{fig:boxperformanceImagenet8}
\end{figure}

As we demonstrated in \secref{sec:single}, on average, \Autopgd 
with \MDabbrv loss  finds more adversarial examples than \Autopgd with all three other loss functions, 
and \cgd attacks performs better than \Autopgd
with \MDabbrv loss. We observe similar results across
datasets, values of $\epsilon$, and defenses, as we show
in \figrefs{fig:boxperformanceImagenet8}{fig:boxperformanceCIFAR8}.

\label{sec:appendix:performance}

\begin{figure}[ht!]
\centerline{\includegraphics[width=0.95\columnwidth]{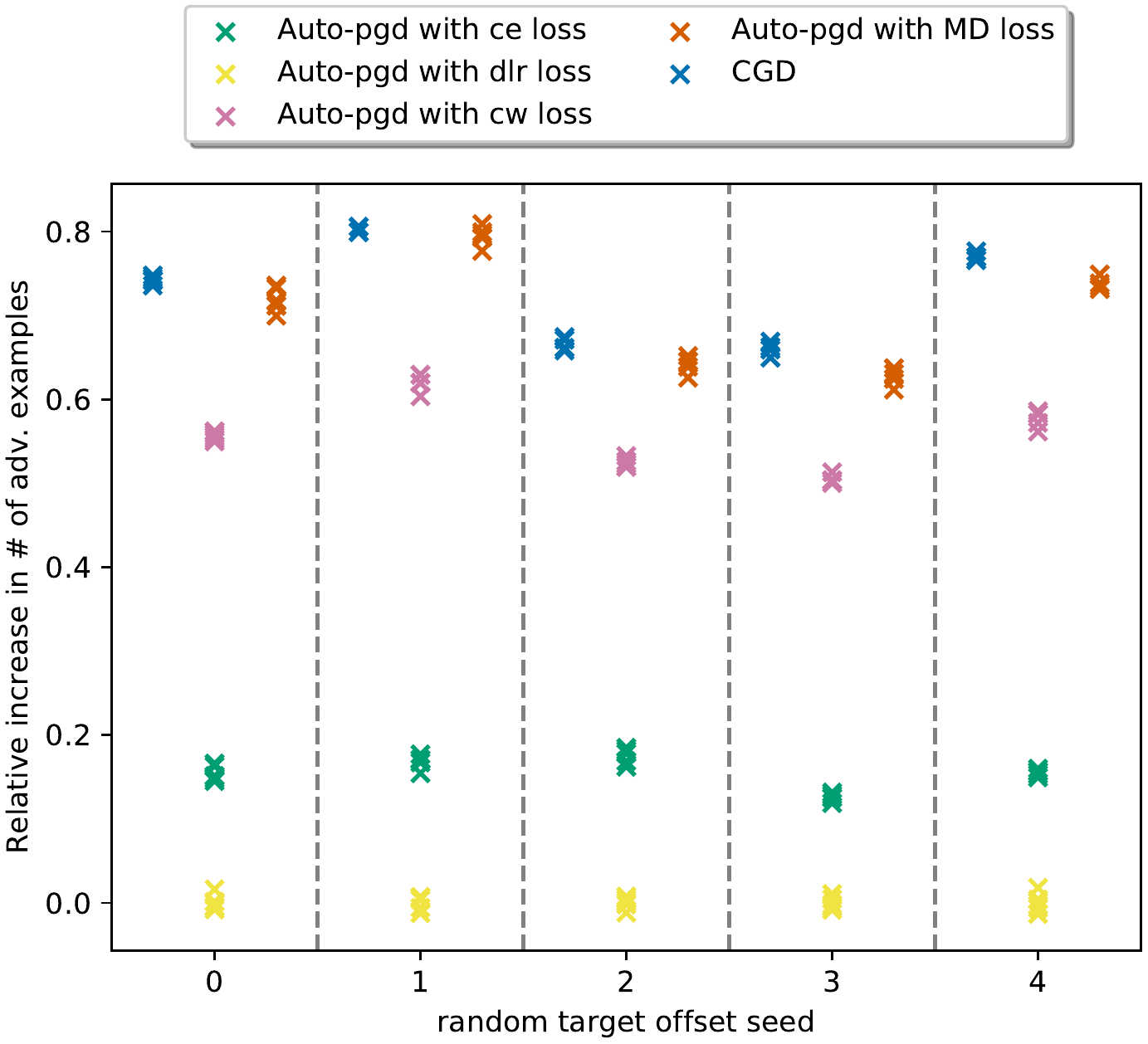}}
\caption{ This figure shows the relative improvement in the number of adversarial examples found by
  attacks on different defenses compared to the worst-performing
  attack. Experiments were performed using 50,000 images from the testing set of \Imagenet. We ran 
attacks using $\epsilon=4/255$ against \SIEKM~\cite{NeurIPS20:SIEKM20}, five different random initial 
perturbations with seeds 0--4, and five different random target offsets with seeds 0--4. The result is 
normalized by the mean of the worst performing method for each target offset.}
\label{fig:boxperformanceImagenet4}
\end{figure}

\begin{figure}[ht!]
\centerline{\includegraphics[width=0.95\columnwidth]{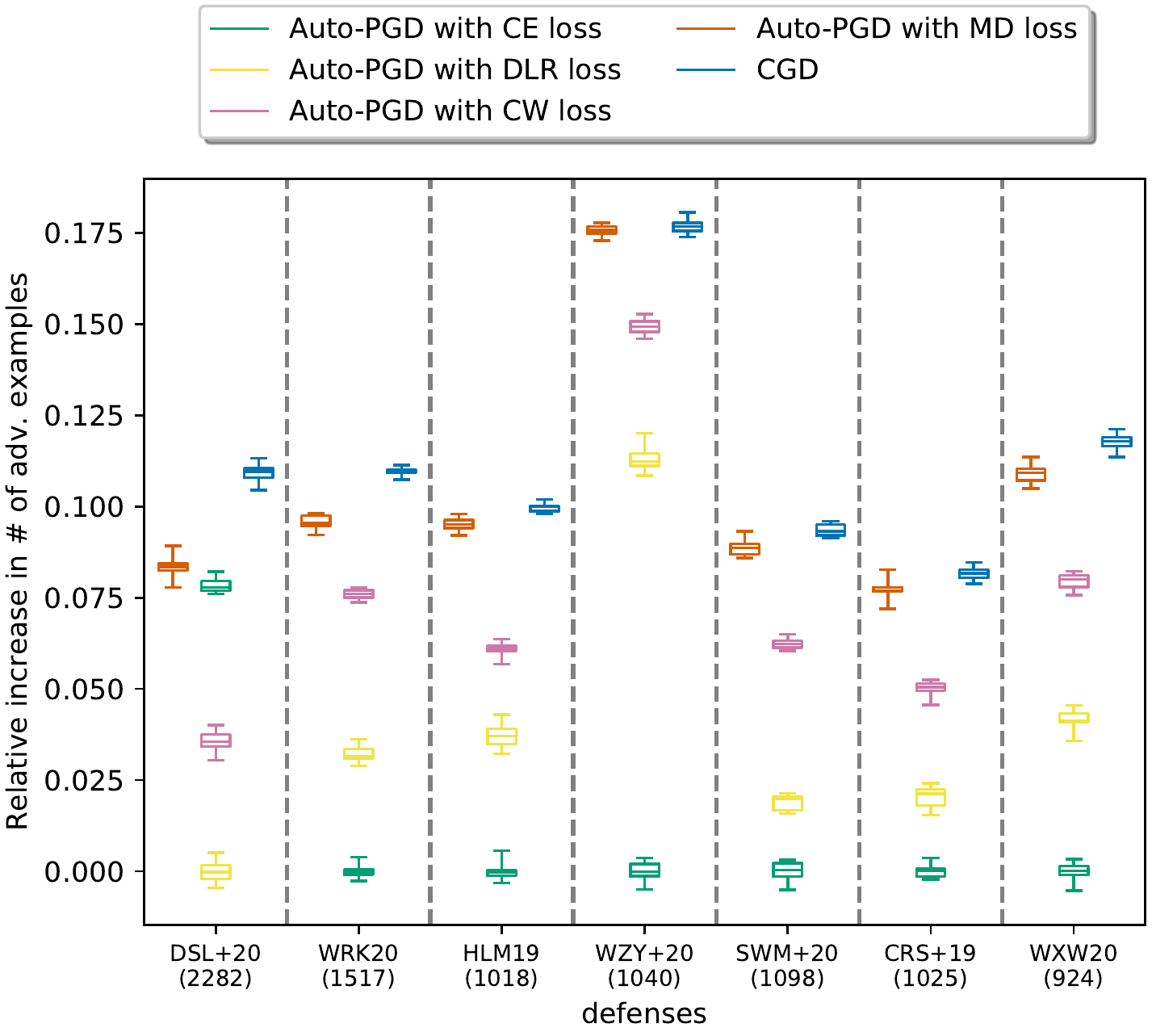}}
\caption{ This figure shows the relative improvement in the number of adversarial examples found by
  attacks on different defenses compared to the worst-performing
  attack. Experiments were performed using 10,000 images from the testing set of CIFAR10. We ran 
attacks using $\epsilon=8/255$, 20 different random initial perturbations with seeds 0--19, and a fixed 
random target offset with seed 
0. The result is normalized by the mean of the worst performing method against each model.}
\label{fig:boxperformanceCIFAR8}
\end{figure}

\section{Statistical Tests On the Performance of Targeted Attacks}
\label{app:stat}
As we described in \secref{sec:stats} and \appref{app:wilcoxon}, we used one-sided Wilcoxon 
signed-rank 
tests\cite{stat45:Wilcoxon} to compare the performance between \Autopgd with MD loss and \cgd, whose results are shown in \tabref{APGDmdVScgdSTAT}. We also compared the 
performance between \cgd and the best performing attack among \Autopgd using the \dlr
loss, CW loss, and CE loss, whose results are shown in \tabref{BestVScgdSTAT}. We conducted 16 
statistical tests in each group, for the 16 different combinations
we had. Thus, to account for
the multiple tests, we used Bonferroni correction to adjust the confidence
level $\alpha$ to $.05/16=0.003125$. 

\begin{table}[htbp]
\small
\caption{This table shows the result of the Wilcoxon signed rank test
  with an adjusted normal approximation for the null hypotheses that
  \Autopgd using the \MDabbrv loss performed equal
  or better than \cgd in each combination of value of $\epsilon$,
  dataset, and defense that we tried. 
    The defenses we used include
  \DSLH~\cite{iclr20:DSLH20}, \WRK~\cite{iclr20:WRK20}, \HLM~\cite{icml19:HLM19},
  \WZYBMG~\cite{iclr20:WZYBMG20}, 
  \SWMJ~\cite{NeurIPS20:SWMJ20}, \CRSLD~\cite{NeurIPS19:CRSLD19}, 
\WXW~\cite{NeurIPS20:WXW20}, 
  and \SIEKM~\cite{NeurIPS20:SIEKM20}.
  We follow the common practice to
  include the Wilcoxon statistics along with the $p$-values. Test
  results where $p$-values are smaller than $\alpha$ are shown in
  bold. We reject the null hypothesis in 11 of the 16 tests.}
\begin{center}
\setlength\tabcolsep{1pt}
\begin{tabular}{|r|r|r@{\hspace{2mm}} r|r@{\hspace{2mm}} r|}
\hline
\textit{dataset} &\textit{defense} &\multicolumn{4}{c|}{$\mathit{l_{\infty}}$ \textit{distance limit}}\\
\Xhline{5\arrayrulewidth}
&& \multicolumn{2}{c|}{$\mathit{\epsilon=8/255}$}&\multicolumn{2}{c|}{$\mathit{\epsilon=16/255}$}\\
\cline{3-6} 
&& \textit{W statistic}&\textit{p value}& \textit{W statistic}&\textit{p value}\\
\cline{2-6} 
&\DSLH&\textbf{2456346.0}&\textbf{2e-10}&5284845.5&0.09\\
&\WRK&\textbf{518476.5}&\textbf{1e-6}&\textbf{2539938.5}&\textbf{3e-8}\\
CIFAR10&\HLM&259521.0&0.02&2377428.0&0.02\\
&\WZYBMG&279327.0&0.07&\textbf{3297549.0}&\textbf{2e-12} \\
&\SWMJ&259521.0&4e-3&\textbf{2247924.0}&\textbf{2e-5}\\
&\CRSLD&\textbf{299366.5}&\textbf{3e-3}&\textbf{2526231.0}&\textbf{1e-5} \\
&\WXW&\textbf{249649.0}&\textbf{3e-3}&\textbf{1788273.0}&\textbf{2e-9}\\
\Xhline{5\arrayrulewidth}
&& \multicolumn{2}{c|}{$\mathit{\epsilon=4/255}$}&\multicolumn{2}{c|}{$\mathit{\epsilon=8/255}$}\\
\cline{3-6} 
\Imagenet&& \textit{W statistic}&\textit{p value}& \textit{W statistic}&\textit{p value}\\
\cline{2-6} 
&\SIEKM&\textbf{9621163.5}&\textbf{7e-12}&\textbf{22184577.0}&\textbf{2e-13}\\
\hline
\end{tabular}
\label{APGDmdVScgdSTAT}
\end{center}
\end{table}

\begin{table}[htbp]
\small
\caption{This table shows the result of the Wilcoxon signed rank tests
  with an adjusted normal approximation for the null hypotheses that
  the best performance among \Autopgd using the \dlr loss, CW loss and
  CE loss performed equal or better than the performance of \cgd on each
  image across all values of $\epsilon$, datasets, and defenses. 
  The defenses we used include
  \DSLH~\cite{iclr20:DSLH20}, \WRK~\cite{iclr20:WRK20}, \HLM~\cite{icml19:HLM19},
  \WZYBMG~\cite{iclr20:WZYBMG20}, 
  \SWMJ~\cite{NeurIPS20:SWMJ20}, \CRSLD~\cite{NeurIPS19:CRSLD19}, 
\WXW~\cite{NeurIPS20:WXW20}, 
  and \SIEKM~\cite{NeurIPS20:SIEKM20}.
  We follow the common practice to include the Wilcoxon statistics along with the $p$-values. Test 
results where $p$-values are smaller than $\alpha$ are shown in bold. We
  reject the null hypothesis in all 16 tests.}

\begin{center}
\setlength\tabcolsep{1pt}
\begin{tabular}{|r|r|r@{\hspace{2mm}} r|r@{\hspace{2mm}} r|}
\hline
\textit{dataset} &\textit{defense} &\multicolumn{4}{c|}{$\mathit{l_{\infty}}$ \textit{distance limit}}\\
\Xhline{5\arrayrulewidth}
&& \multicolumn{2}{c|}{$\mathit{\epsilon=8/255}$}&\multicolumn{2}{c|}{$\mathit{\epsilon=16/255}$}\\
\cline{3-6} 
&& \textit{W statistic}&\textit{p value}& \textit{W statistic}&\textit{p value}\\
\cline{2-6} 
&\DSLH&\textbf{2311330.5}&\textbf{1e-6}&\textbf{6709608.5}&\textbf{7e-10}\\
&\WRK&\textbf{657798.5}&\textbf{3e-8}&\textbf{3576641.5}&\textbf{7e-26}\\
CIFAR10&\HLM&\textbf{538492.0}&\textbf{2e-8}&\textbf{3271871.0}&\textbf{4e-14}\\
&\WZYBMG&\textbf{508499.5}&\textbf{2e-5}&\textbf{4021878.0}&\textbf{9e-25}\\
&\SWMJ&\textbf{508657.5}&\textbf{1e-8}&\textbf{2540675.0}&\textbf{2e-9}\\
&\CRSLD&\textbf{508587.0}&\textbf{1e-6}&\textbf{2731369.5}&\textbf{9e-7}\\
&\WXW&\textbf{498731.5}&\textbf{1e-8}&\textbf{2004237.0}&\textbf{1e-13}\\
\Xhline{5\arrayrulewidth}
&& \multicolumn{2}{c|}{$\mathit{\epsilon=4/255}$}&\multicolumn{2}{c|}{$\mathit{\epsilon=8/255}$}\\
\cline{3-6} 
\Imagenet&& \textit{W statistic}&\textit{p value}& \textit{W statistic}&\textit{p value}\\
\cline{2-6} 
&\SIEKM&\textbf{19071030.0}&\textbf{3e-55}&\textbf{54798304.5}&\textbf{1e-168}\\
\hline
\end{tabular}
\label{BestVScgdSTAT}
\end{center}
\end{table}

\section{More Evaluation Setup for Untargeted Attacks}
\label{app:UntargetedSetup}
We evaluated untargeted attacks on the CIFAR10 dataset. We use the \lp{\infty} distance metric for 
these attacks as
we did for targeted attacks. We use three values of $\epsilon$: $4/255$,
$8/255$, and $16/255$. We use the same seven defenses ~\cite{NeurIPS19:CRSLD19, 
iclr20:DSLH20,icml19:HLM19,NeurIPS20:SWMJ20,
  iclr20:WRK20,NeurIPS20:WXW20,iclr20:WZYBMG20} as we did in \appref{app:setup}, again trained 
with $\epsilon=8/255$ from
robust-bench~\cite{arxiv21:robustbench}, as benchmarks to
evaluate untargeted attacks. We evaluated untargeted attacks by the
number of adversarial examples they found within the \lp{\infty}
distance limit. We ran all attacks, again with 100 iterations as in its
default configuration. We ran each attack with five different random
initial perturbations, using seeds 0--4, and a batch size of 512.

\section{Performance of Untargeted Attacks}

\begin{figure}[ht!]
\centerline{\includegraphics[width=0.95\columnwidth]{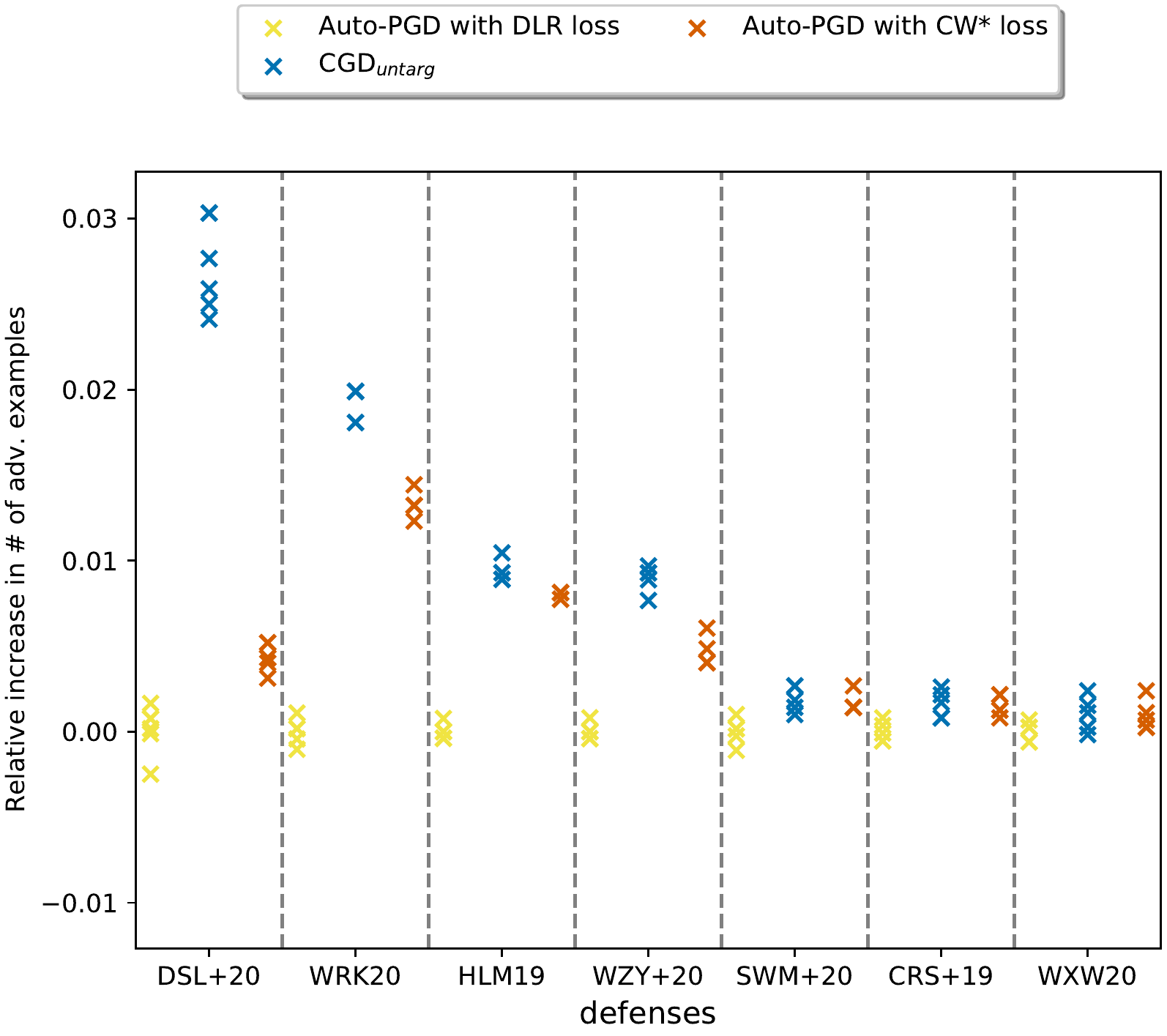}}
\caption{This 
figure shows the relative improvement in the number of adversarial examples found by
 untargeted attacks on different defenses compared to the worst-performing
  attack. Experiments were performed using 10,000 images from the testing set of CIFAR10. We ran 
attacks using $\epsilon=4/255$,
  and five different random initial perturbations with seeds 0--4. The result is
  normalized by the mean of the worst performing method against each
  model.}
\label{untargeted4}
\end{figure}

As we introduced in \secref{sec:discussion:untargeted}, on average, \cgduntarg performed better than 
\Autopgd with
  $L_{\mathit{\dlr}}$ and \Autopgd with $L_{\mathit{\cw^{*}}}$.
   \figrefs{untargeted4}{untargeted16} show similar results when we use different values of 
$\epsilon$.

\label{sec:appendix:untargeted}

\begin{figure}[ht!]
\centerline{\includegraphics[width=0.95\columnwidth]{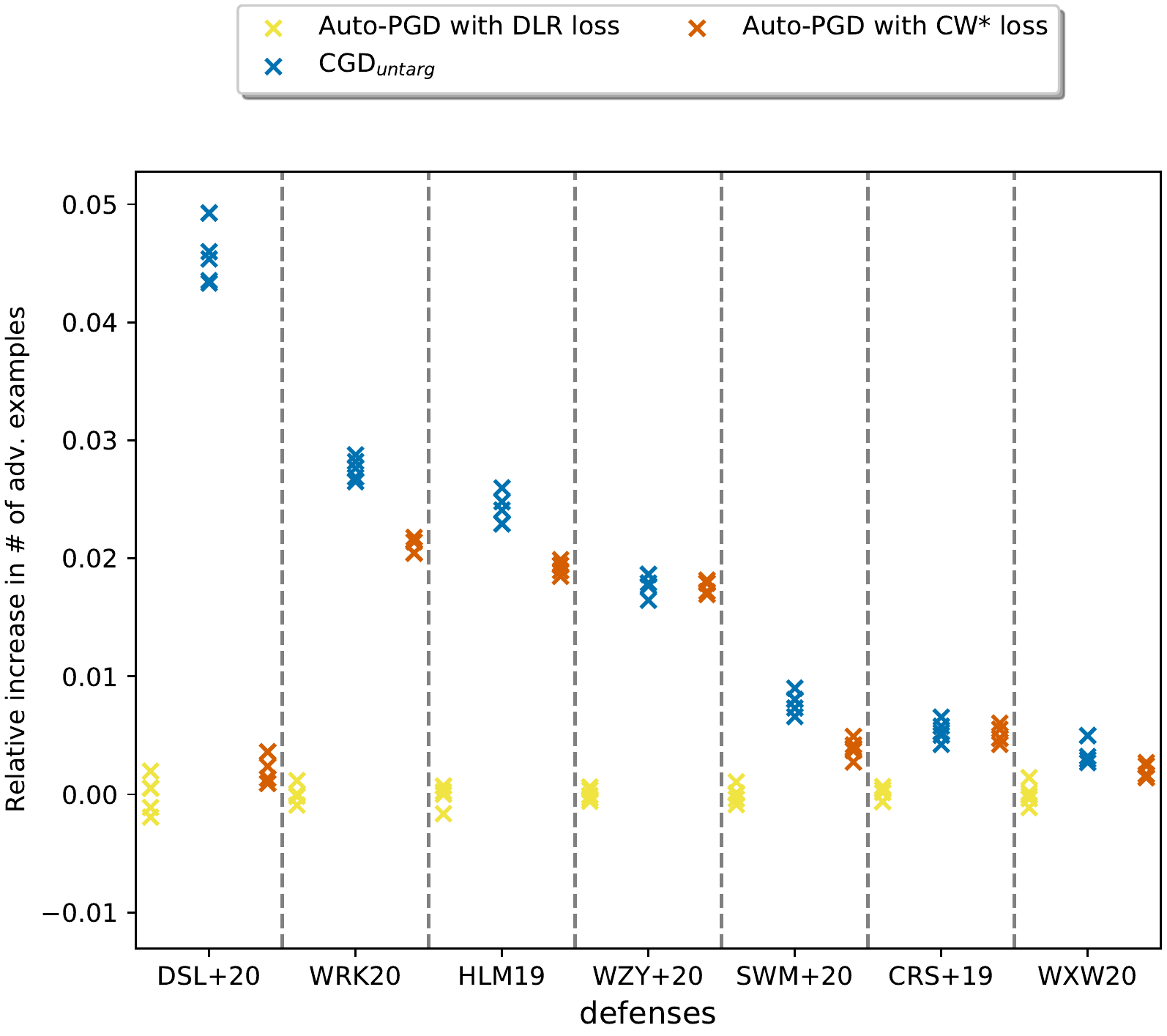}}
\caption{This figure shows the relative improvement in the number of adversarial examples found by
untargeted attacks on different defenses compared to the worst-performing
  attack. Experiments were performed using 10,000 images from the testing set of CIFAR10. We ran 
attacks using $\epsilon=8/255$,
  and five different random initial perturbations with seeds 0--4. The result is
  normalized by the mean of the worst performing method against each
  model.}
\label{untargeted8}
\end{figure}

\begin{figure}[ht!]
\centerline{\includegraphics[width=0.95\columnwidth]{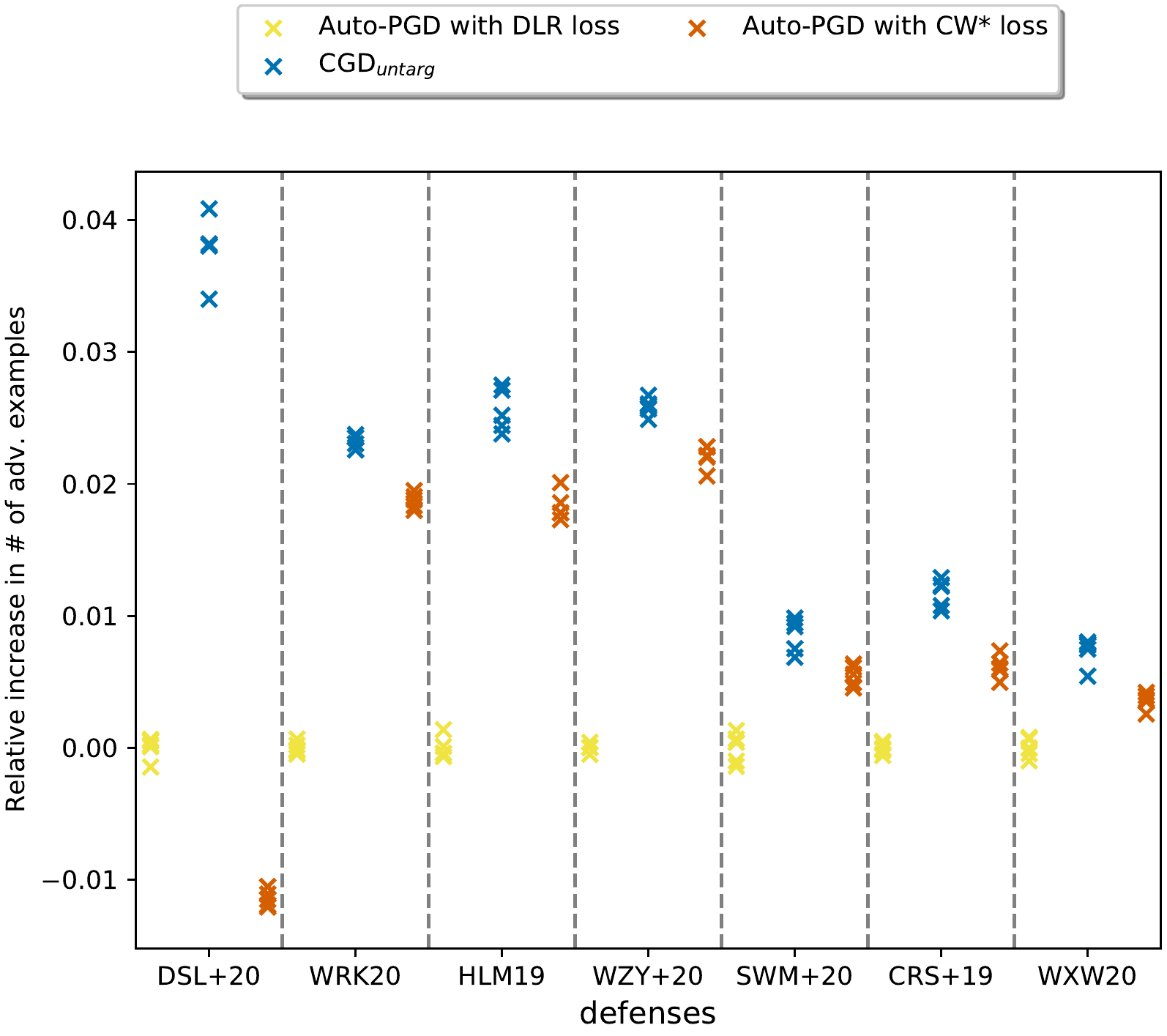}}
\caption{This figure shows the relative improvement in the number of adversarial examples found by
untargeted attacks on different defenses compared to the worst-performing
  attack. Experiments were performed using 10,000 images from the testing set of CIFAR10. We ran 
attacks using $\epsilon=16/255$,
  and five different random initial perturbations with seeds 0--4. The result is
  normalized by the mean of the worst performing method against each
  model.}
\label{untargeted16}
\end{figure}

\section{Uniqueness of Attacks}
\label{app:result:uniqueness}
\begin{figure}[ht!]
\centerline{\includegraphics[width=0.95\columnwidth]{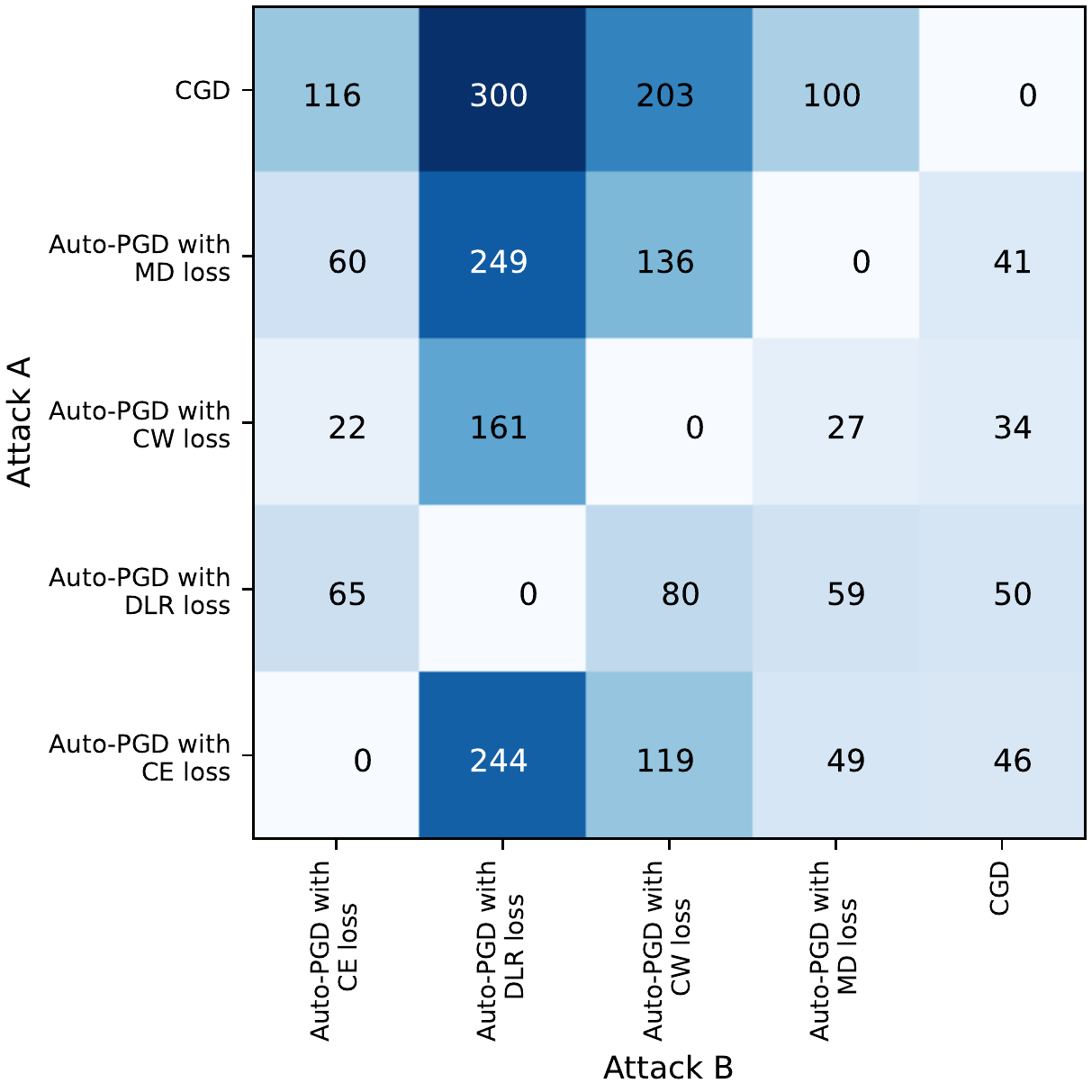}}
\caption{ This image shows the average number of  successful adversarial examples found by attack A 
but not by attack B in each set of 10,000 attempts on the testing set of CIFAR10 using 
$\epsilon=8/255$ 
against the \DSLH~\cite{iclr20:DSLH20} defense, rounded to whole numbers.}
\label{1v1DSLH8}
\end{figure}

As we described  \secref{sec:result:uniqueness},  among the attack methods we
tried, each of them found a slightly different set of successful
adversarial examples, and each 
attack found some
adversarial examples which any of the other methods did not.  We observe the same phenomena 
across attack methods, values of $\epsilon$, defenses, and datasets, as shown in
\figrefs{1v1DSLH8}{ImageNet1VSothers}.

\begin{figure}[ht!]
\centerline{\includegraphics[width=0.95\columnwidth]{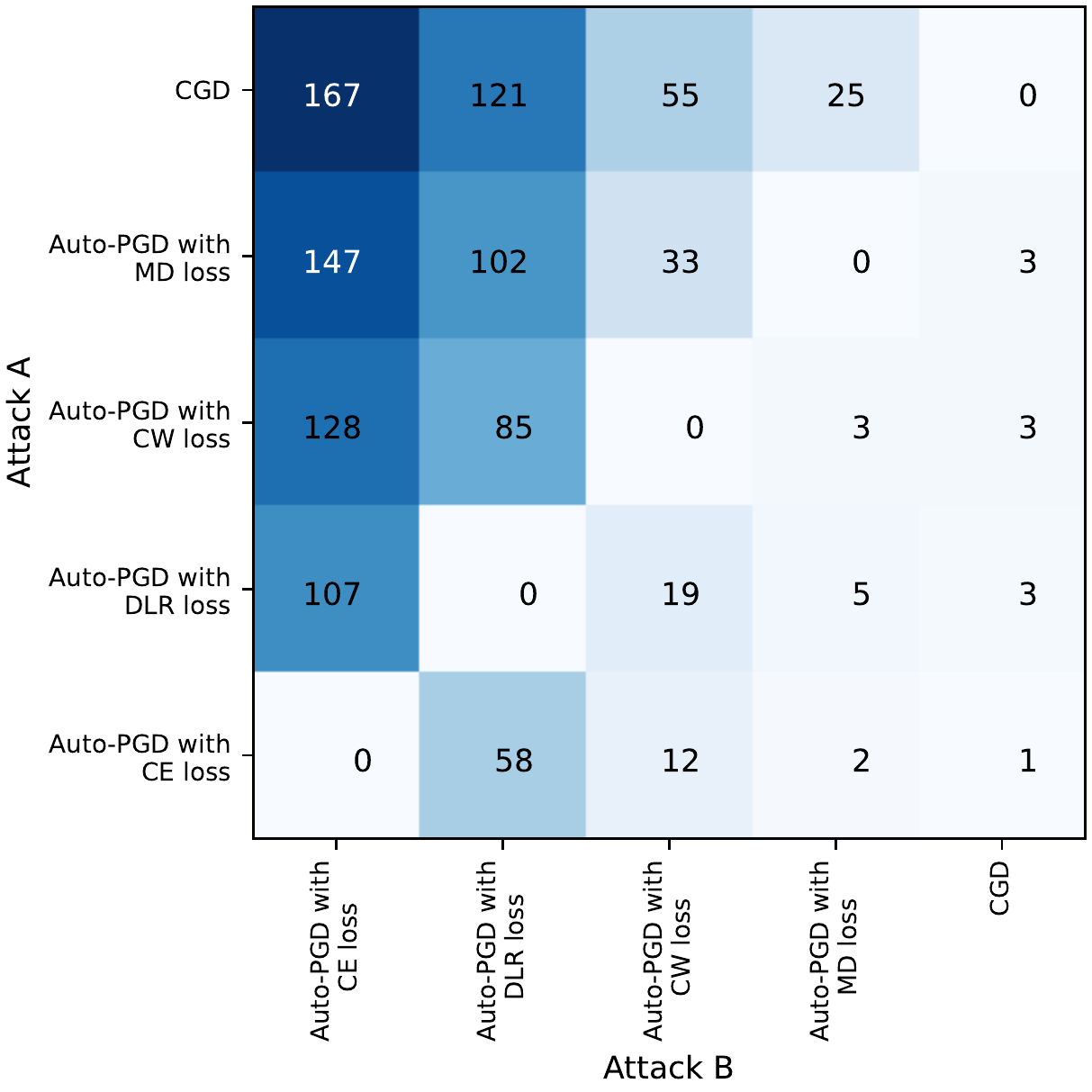}}
\caption{ This image shows the average number of  successful adversarial examples found by attack A 
but not by attack B in each set of 10,000 attempts on the testing set of CIFAR10 using 
$\epsilon=8/255$ 
against the \WRK~\cite{iclr20:WRK20} defense, rounded to whole numbers.}
\end{figure}

\begin{figure}[ht!]
\centerline{\includegraphics[width=0.95\columnwidth]{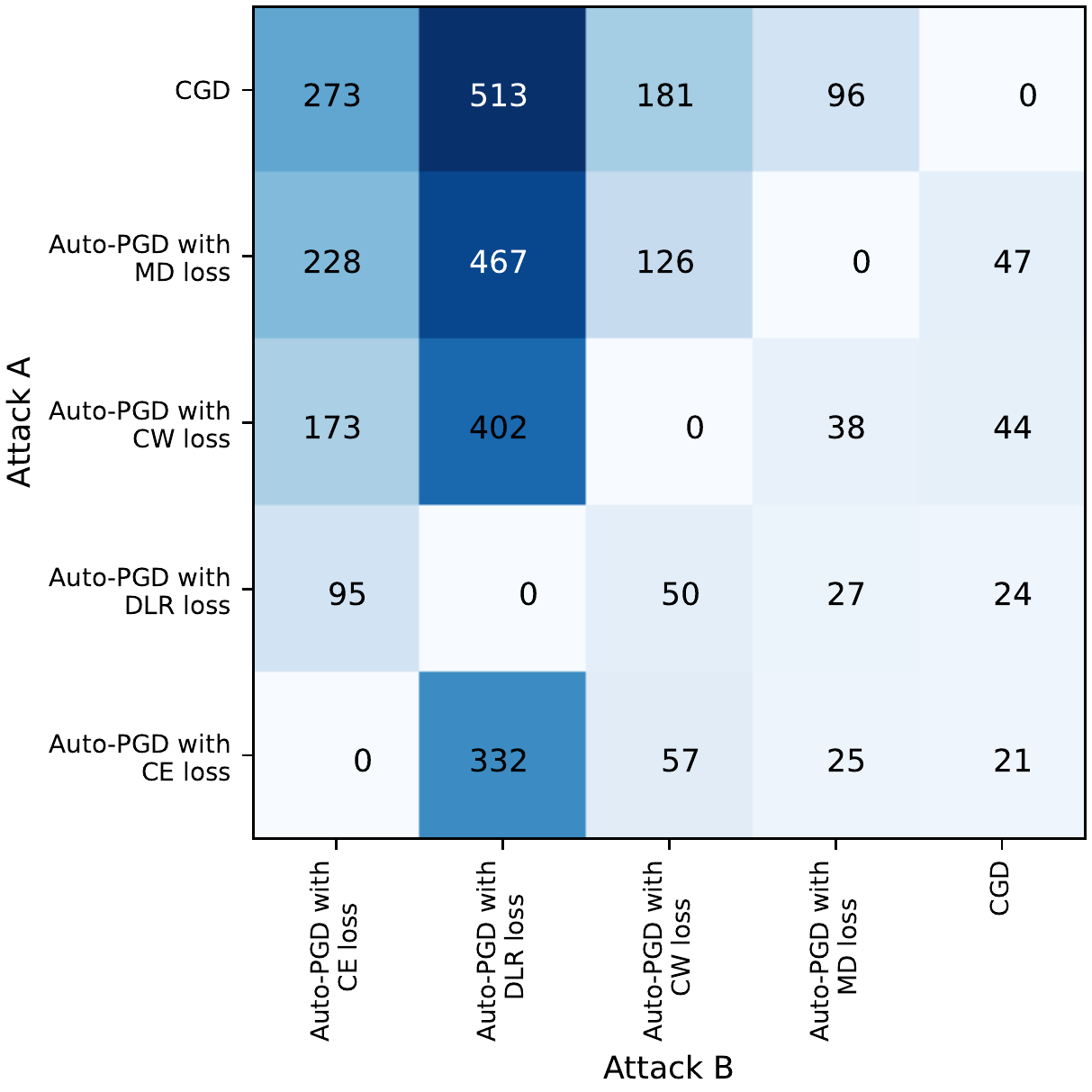}}
\caption{ This image shows the average number of  successful adversarial examples found by attack A 
but not by attack B in each set of 10,000 attempts on the testing set of CIFAR10 using 
$\epsilon=16/255$ 
against the \WRK~\cite{iclr20:WRK20} defense, rounded to whole numbers}
\end{figure}

\begin{figure}[ht!]
\centerline{\includegraphics[width=0.95\columnwidth]{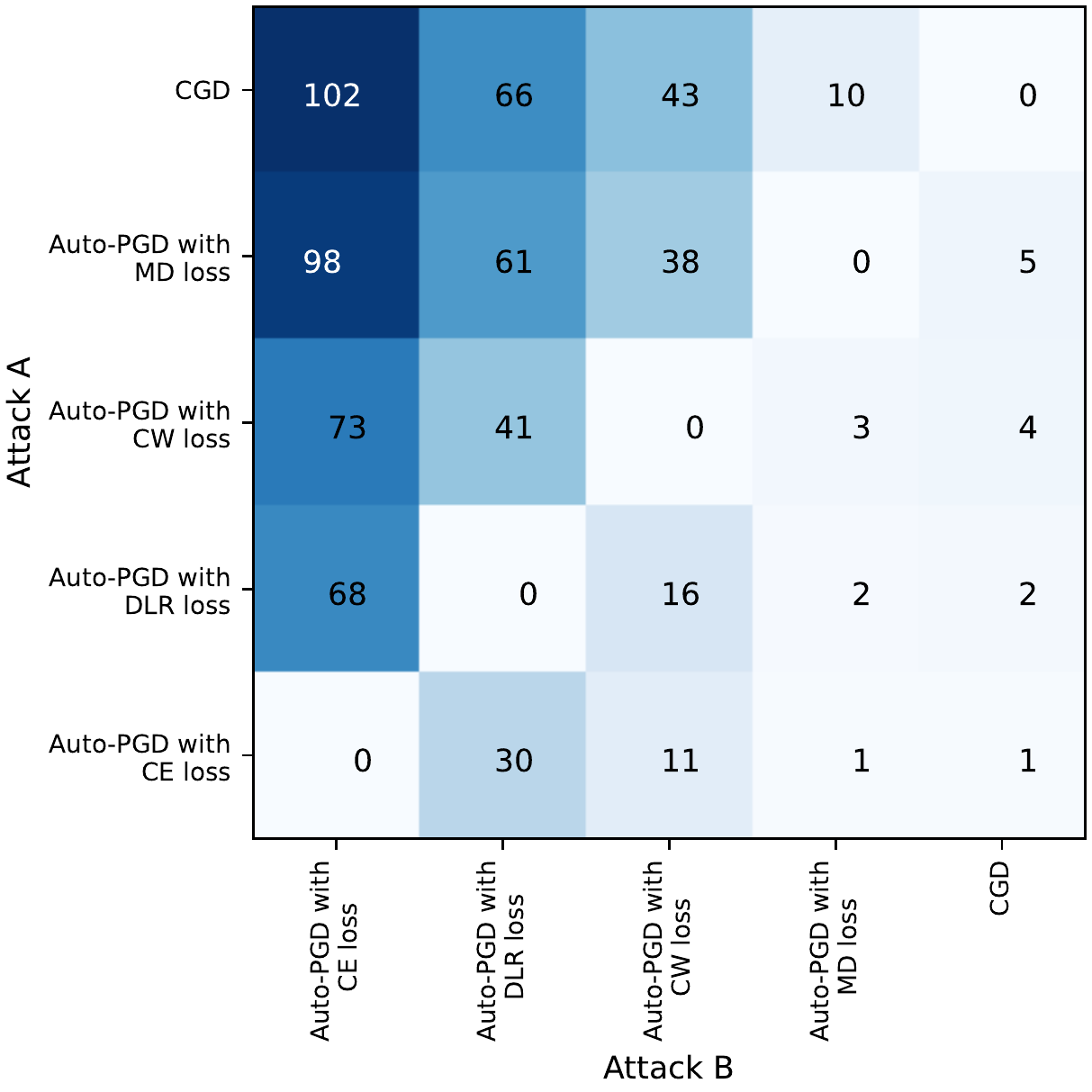}}
\caption{ This image shows the average number of  successful adversarial examples found by attack A 
but not by attack B in each set of 10,000 attempts on the testing set of CIFAR10 using 
$\epsilon=8/255$ 
against the \HLM~\cite{icml19:HLM19} defense, rounded to whole numbers.}
\end{figure}

\begin{figure}[ht!]
\centerline{\includegraphics[width=0.95\columnwidth]{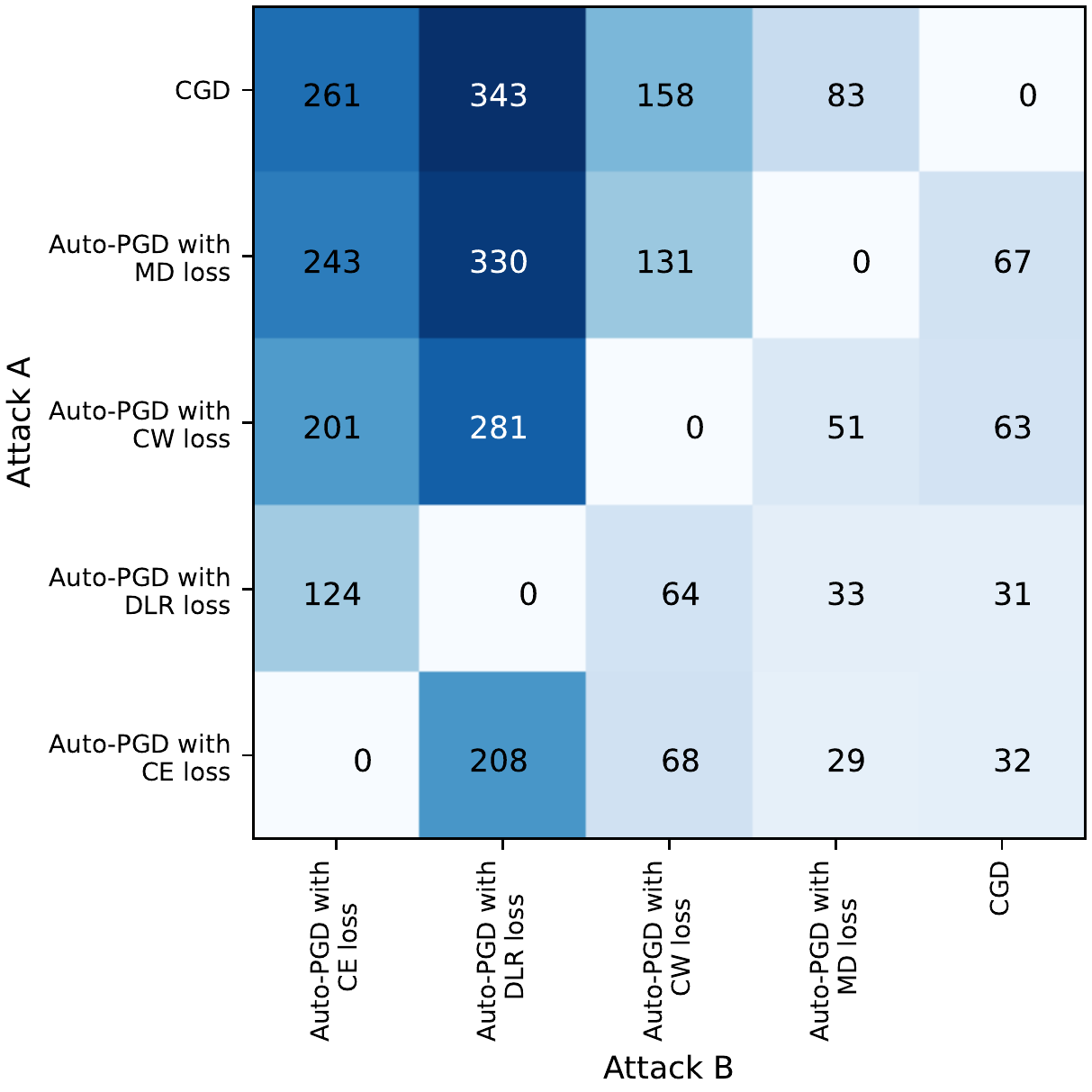}}
\caption{ This image shows the average number of  successful adversarial examples found by attack A 
but not by attack B in each set of 10,000 attempts on the testing set of CIFAR10 using 
$\epsilon=16/255$ 
against the \HLM~\cite{icml19:HLM19} defense, rounded to whole numbers.}
\end{figure}

\begin{figure}[ht!]
\centerline{\includegraphics[width=0.95\columnwidth]{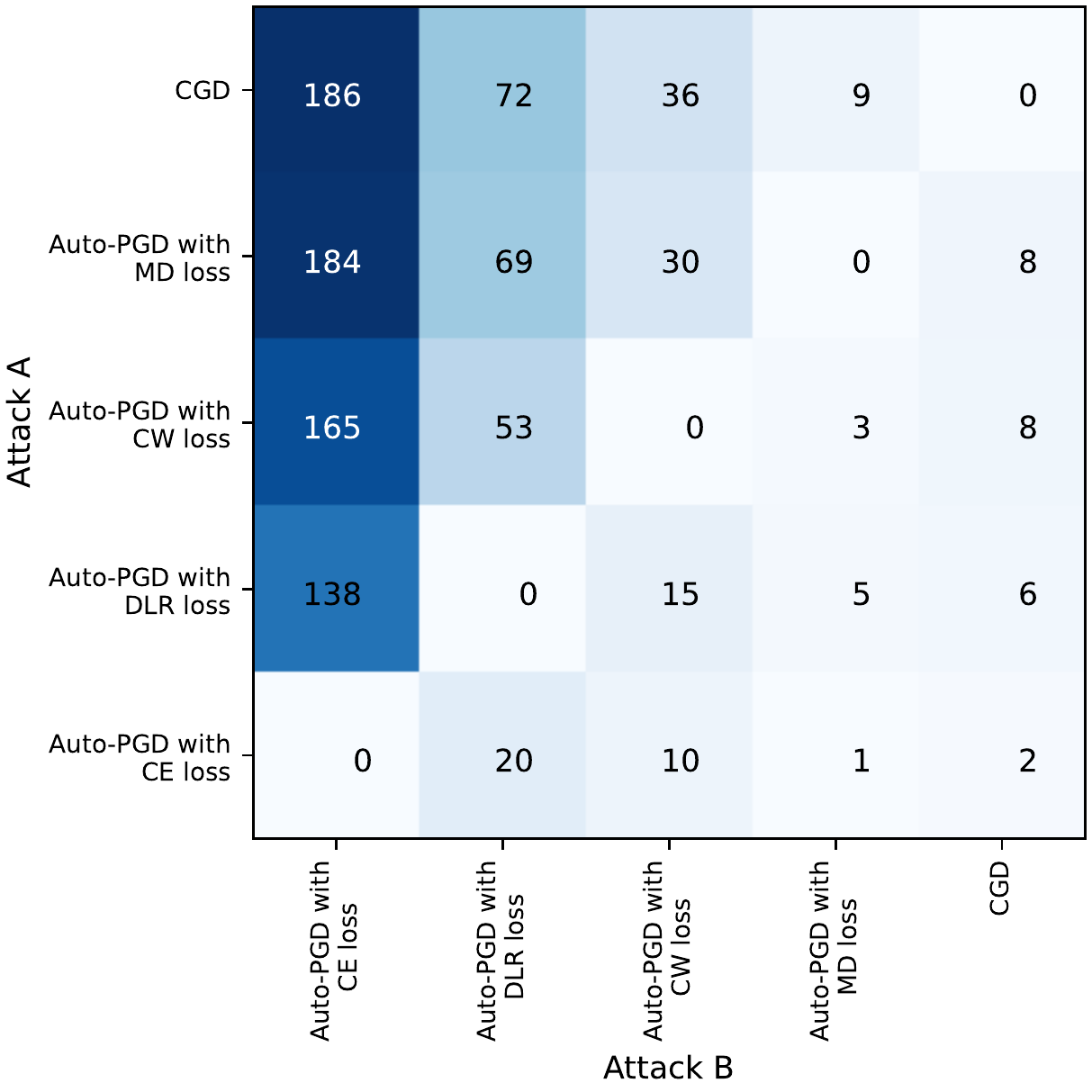}}
\caption{ This image shows the average number of  successful adversarial examples found by attack A 
but not by attack B in each set of 10,000 attempts on the testing set of CIFAR10 using 
$\epsilon=8/255$ 
against the \WZYBMG~\cite{iclr20:WZYBMG20} defense, rounded to whole numbers.}
\end{figure}

\begin{figure}[ht!]
\centerline{\includegraphics[width=0.95\columnwidth]{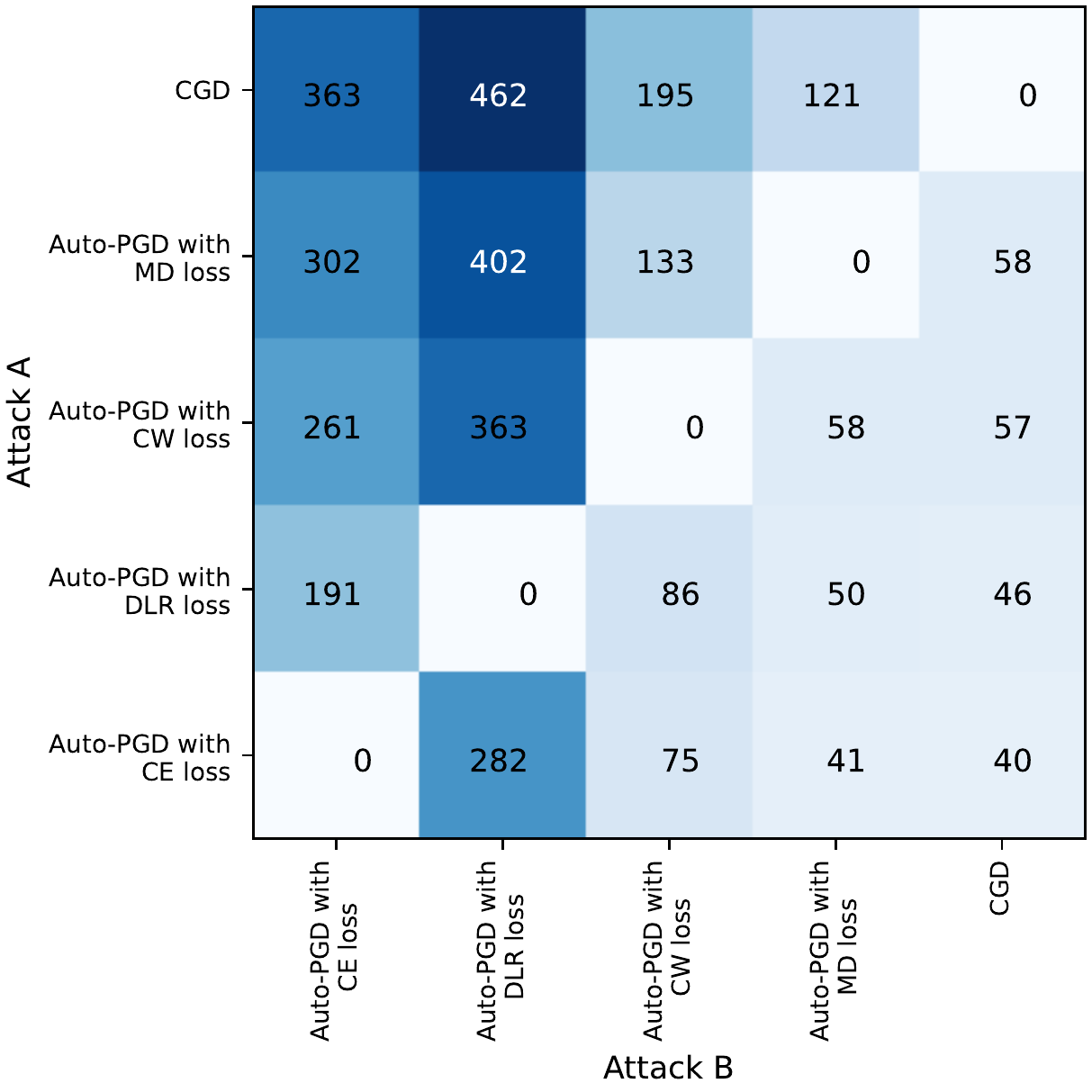}}
\caption{ This image shows the average number of  successful adversarial examples found by attack A 
but not by attack B each set of 10,000 attempts on the testing set of CIFAR10 using 
$\epsilon=16/255$ 
against the \WZYBMG~\cite{iclr20:WZYBMG20} defense, rounded to whole numbers.}
\end{figure}

\begin{figure}[ht!]
\centerline{\includegraphics[width=0.95\columnwidth]{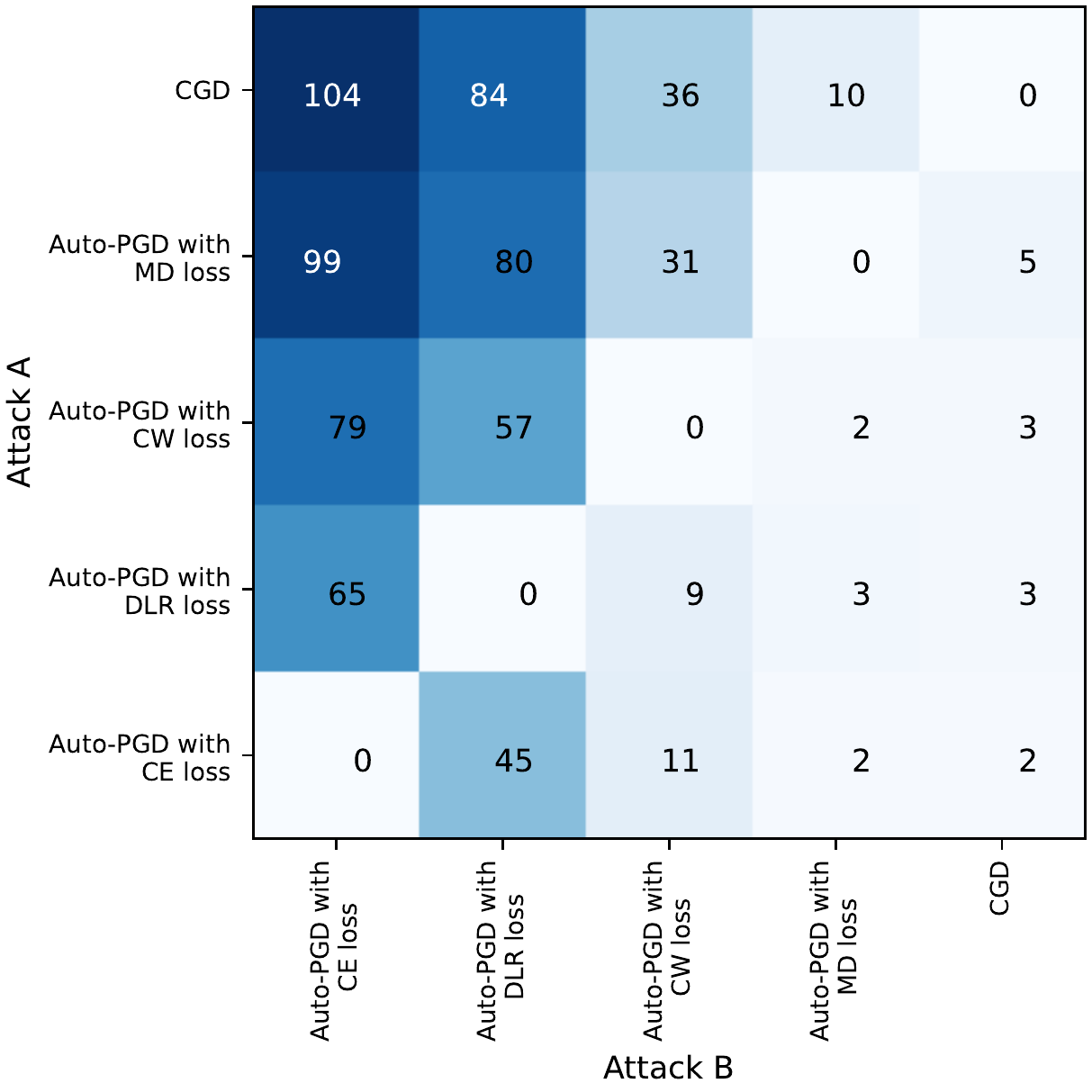}}
\caption{ This image shows the average number of  successful adversarial examples found by attack A 
but not by attack B in each set of 10,000 attempts on the testing set of CIFAR10 using 
$\epsilon=8/255$ 
against the \SWMJ~\cite{NeurIPS20:SWMJ20} defense, rounded to whole numbers.}
\end{figure}

\begin{figure}[ht!]
\centerline{\includegraphics[width=0.95\columnwidth]{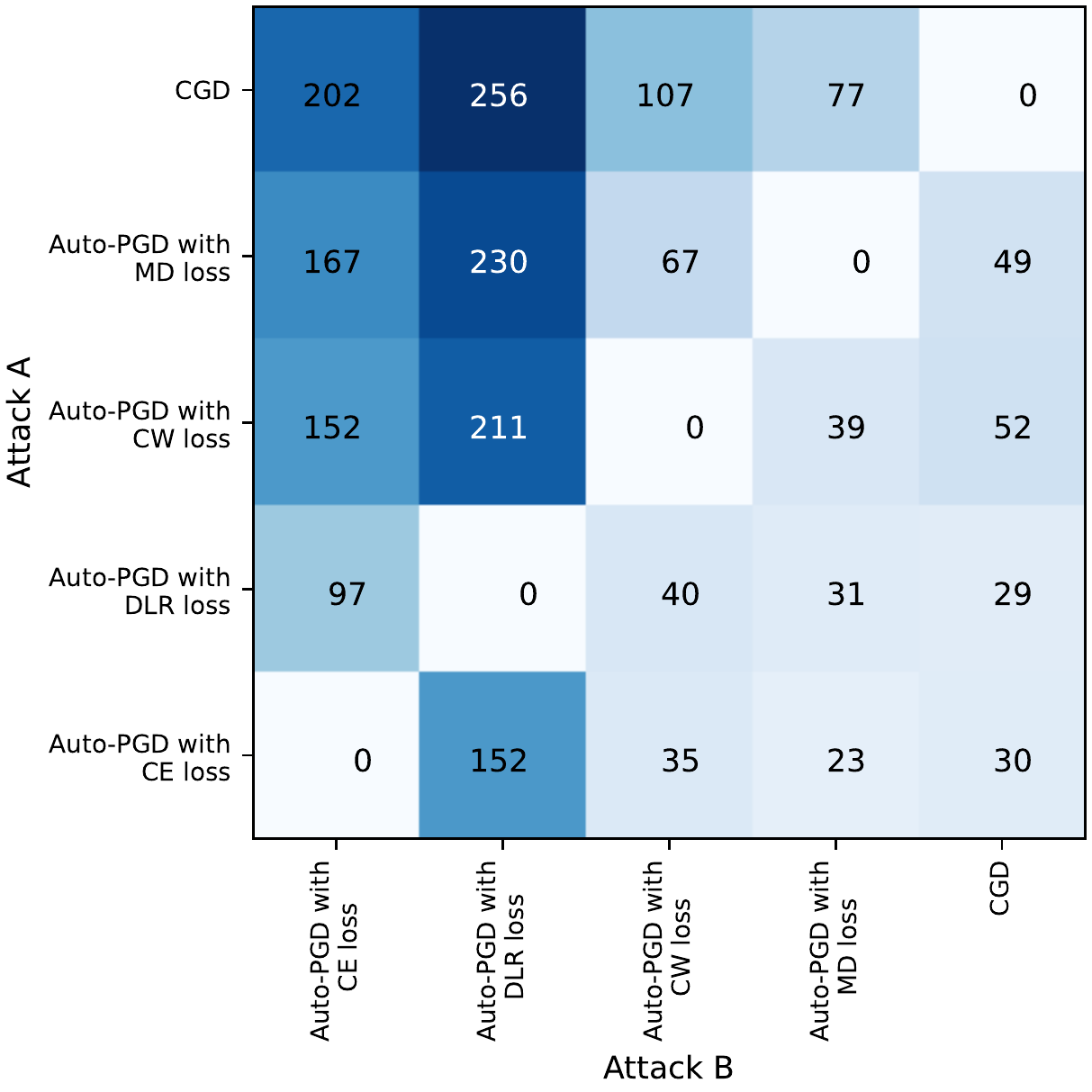}}
\caption{ This image shows the average number of  successful adversarial examples found by attack A 
but not by attack B in each set of 10,000 attempts on the testing set of CIFAR10 using 
$\epsilon=16/255$ 
against the \SWMJ~\cite{NeurIPS20:SWMJ20} defense, rounded to whole numbers.}
\end{figure}

\begin{figure}[ht!]
\centerline{\includegraphics[width=0.95\columnwidth]{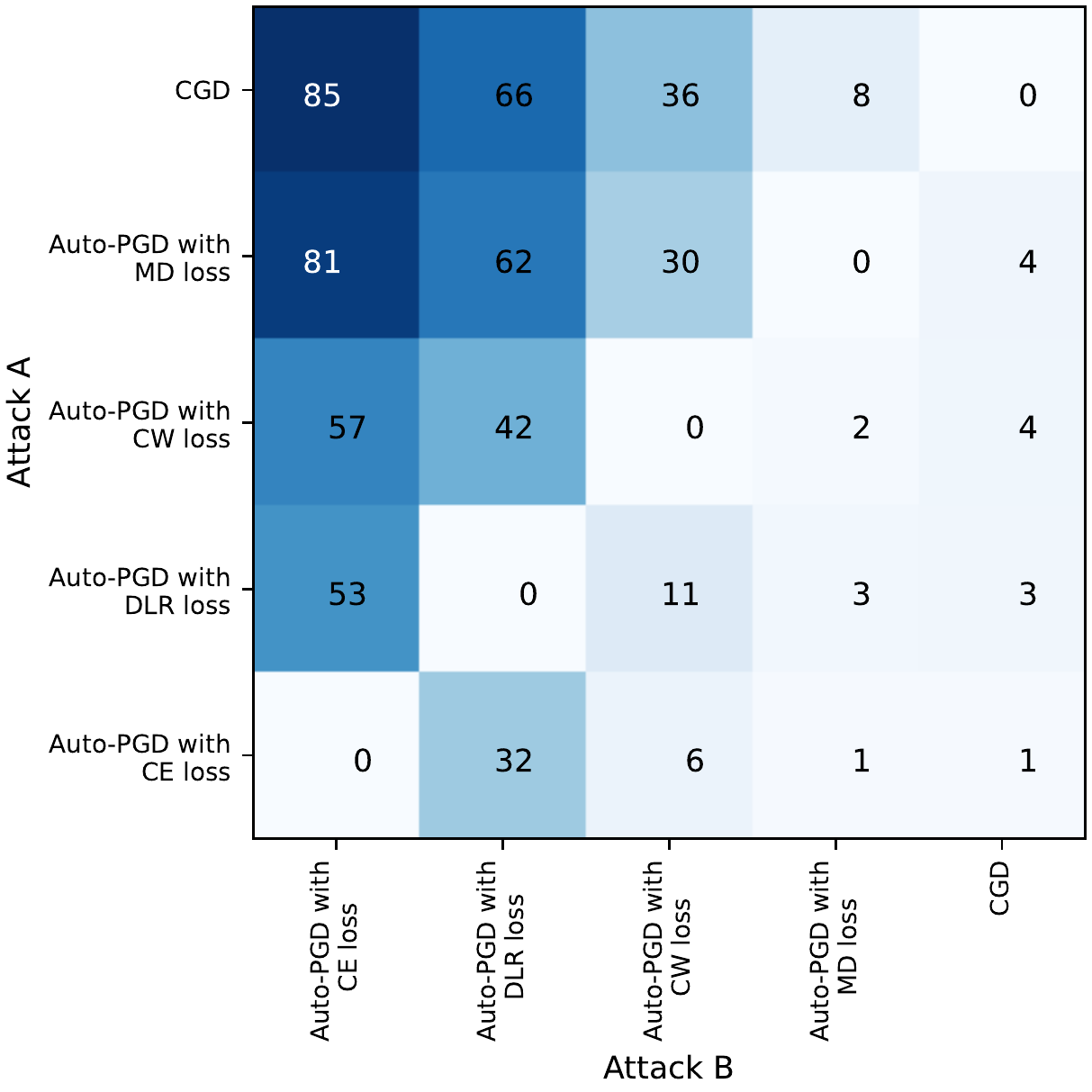}}
\caption{ This image shows the average number of  successful adversarial examples found by attack A 
but not by attack B in each set of 10,000 attempts on the testing set of CIFAR10 using 
$\epsilon=8/255$ 
against the \CRSLD~\cite{NeurIPS19:CRSLD19} defense, rounded to whole numbers.}
\end{figure}

\begin{figure}[ht!]
\centerline{\includegraphics[width=0.95\columnwidth]{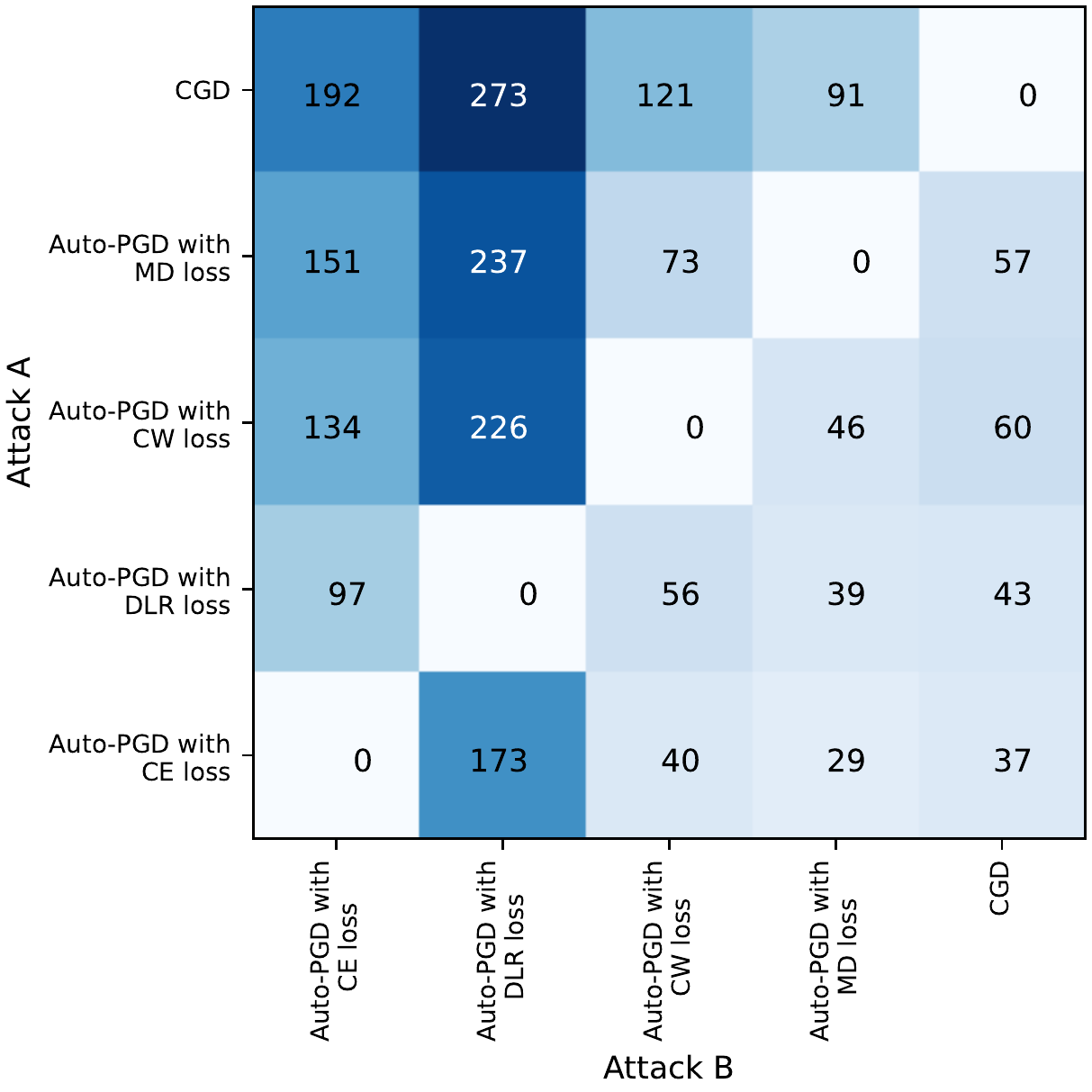}}
\caption{ This image shows the average number of  successful adversarial examples found by attack A 
but not by attack B in each set of 10,000 attempts on the testing set of CIFAR10 using 
$\epsilon=16/255$ 
against the \CRSLD~\cite{NeurIPS19:CRSLD19} defense, rounded to whole numbers.}
\end{figure}

\begin{figure}[ht!]
\centerline{\includegraphics[width=0.95\columnwidth]{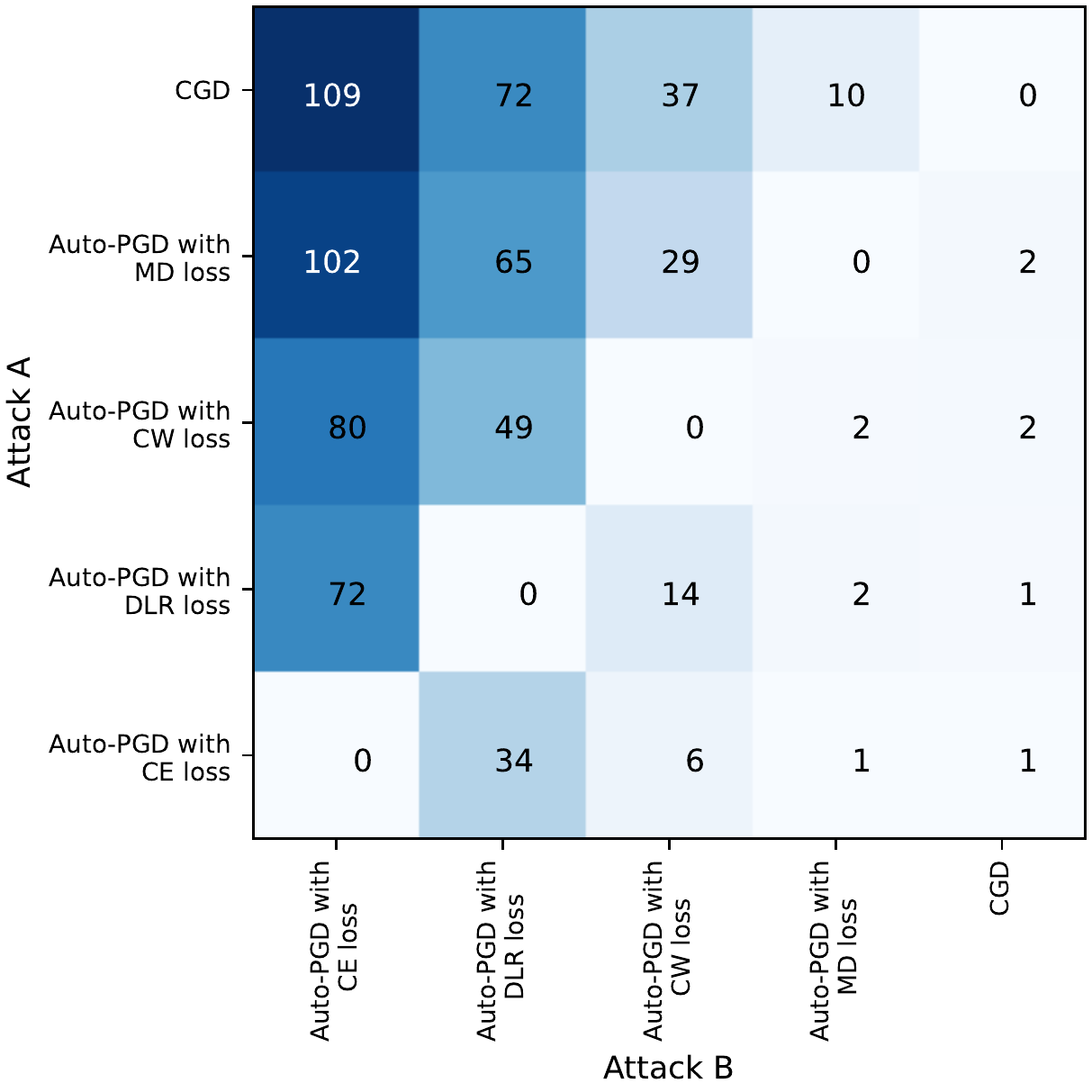}}
\caption{ This image shows the average number of  successful adversarial examples found by attack A 
but not by attack B in each set of 10,000 attempts on the testing set of CIFAR10 using 
$\epsilon=8/255$ 
against the \WXW~\cite{NeurIPS20:WXW20} defense, rounded to whole numbers.}
\end{figure}

\begin{figure}[ht!]
\centerline{\includegraphics[width=0.95\columnwidth]{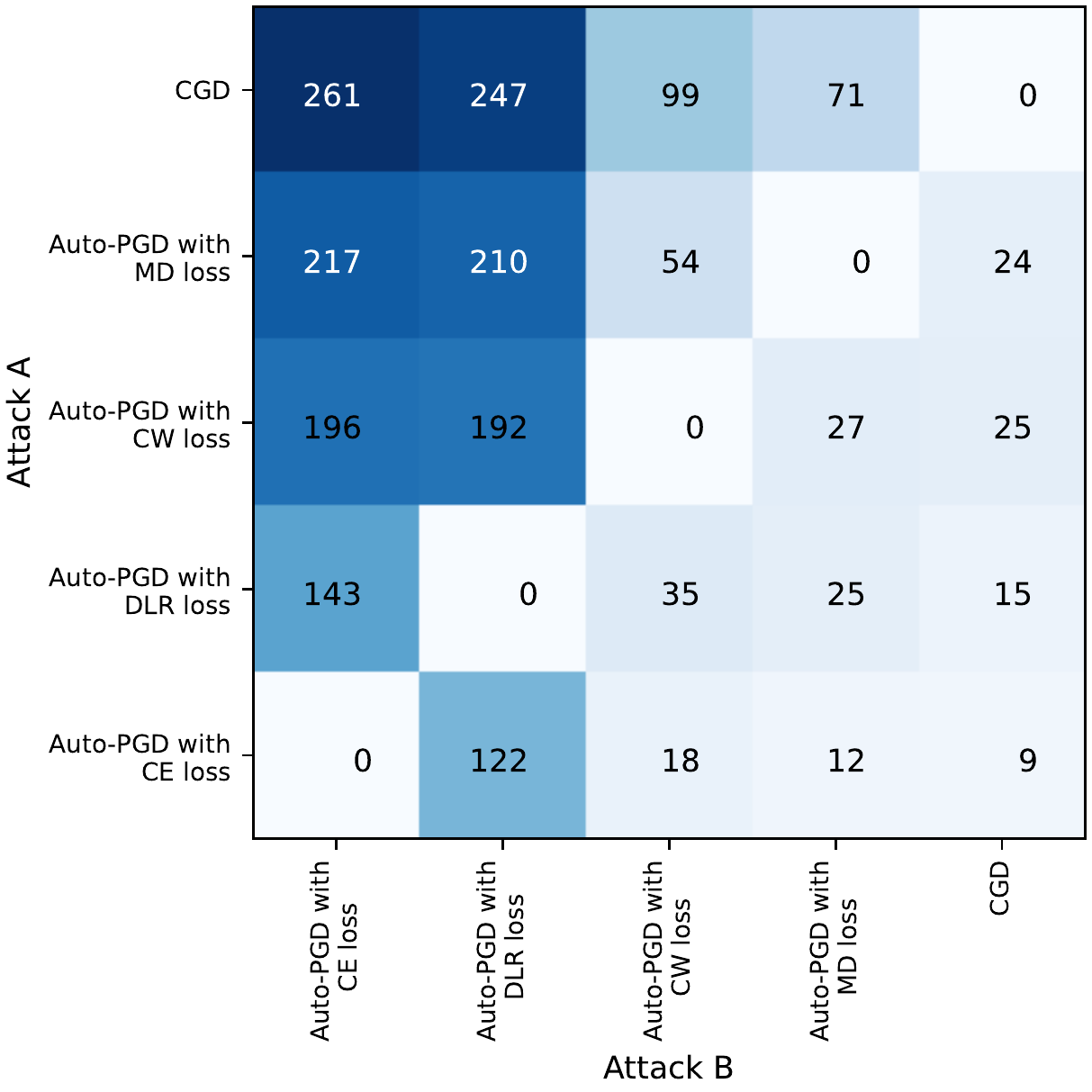}}
\caption{ This image shows the average number of  successful adversarial examples found by attack A 
but not by attack B in each set of 10,000 attempts on the testing set of CIFAR10 using 
$\epsilon=16/255$ 
against the \WXW~\cite{NeurIPS20:WXW20} defense, rounded to whole numbers.}
\end{figure}

\begin{figure}[ht!]
\centerline{\includegraphics[width=0.95\columnwidth]{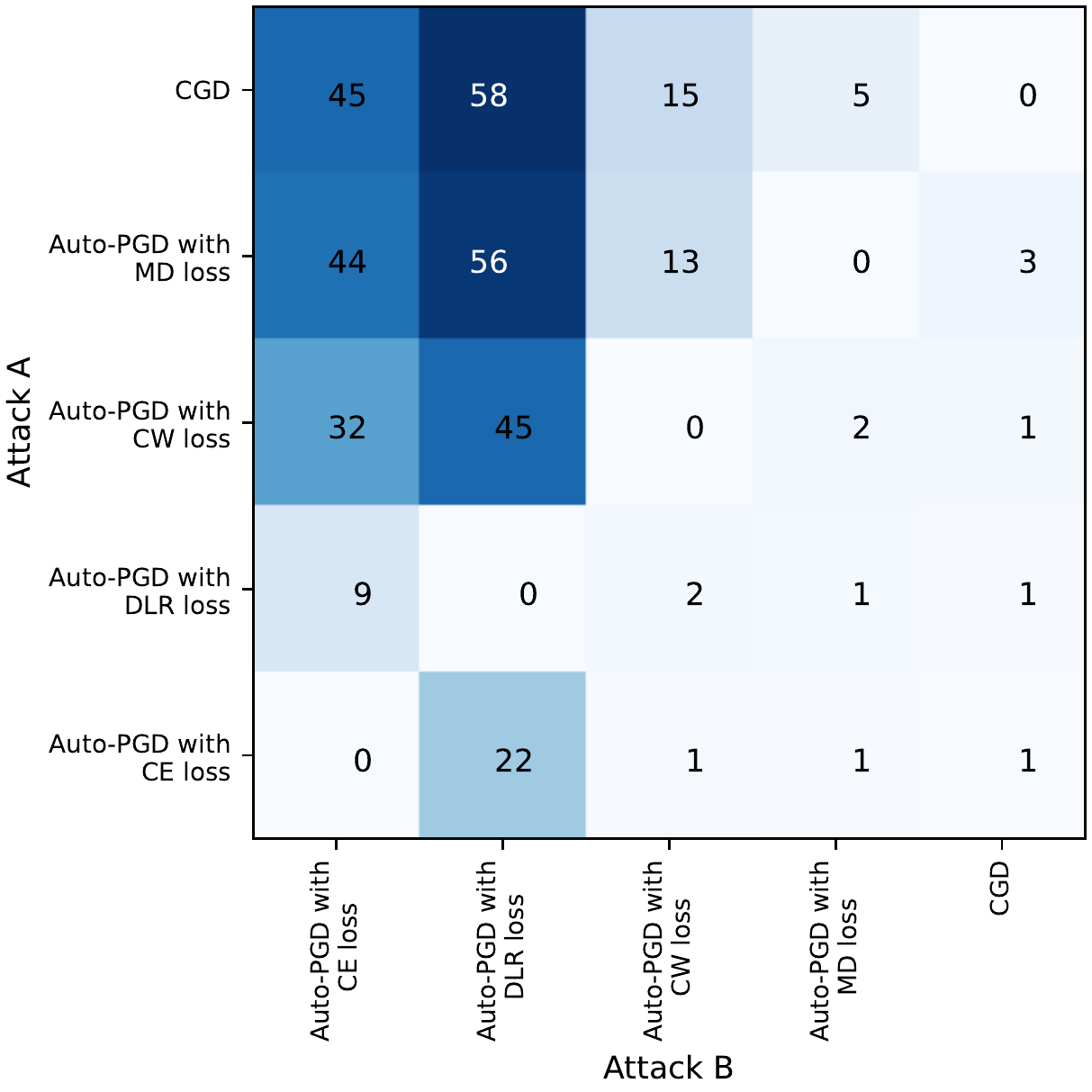}}
\caption{ This image shows the average number of  successful adversarial examples found by attack A 
but not by attack B in each set of 50,000 attempts on the testing set of \Imagenet using 
$\epsilon=4/255$ 
against the \SIEKM~\cite{NeurIPS20:SIEKM20} defense, rounded to whole numbers.}
\end{figure}

\begin{figure}[ht!]
\centerline{\includegraphics[width=0.95\columnwidth]{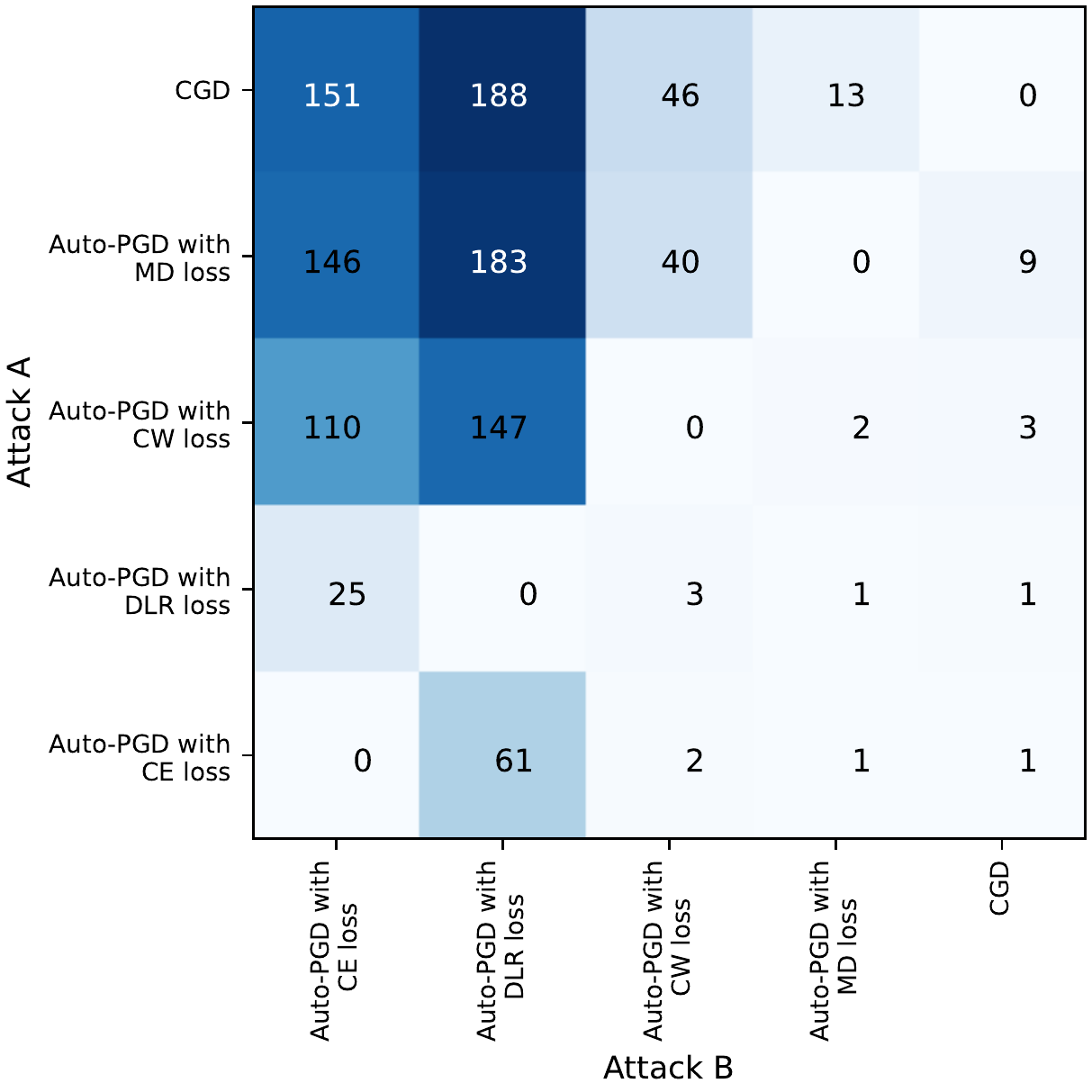}}
\caption{ This image shows the average number of  successful adversarial examples found by attack A 
but not by attack B in each set of 50,000 attempts on the testing set of \Imagenet using 
$\epsilon=8/255$ 
against the \SIEKM~\cite{NeurIPS20:SIEKM20} defense, rounded to whole numbers.}
\end{figure}

\begin{figure}[ht!]
\centerline{\includegraphics[width=0.95\columnwidth]{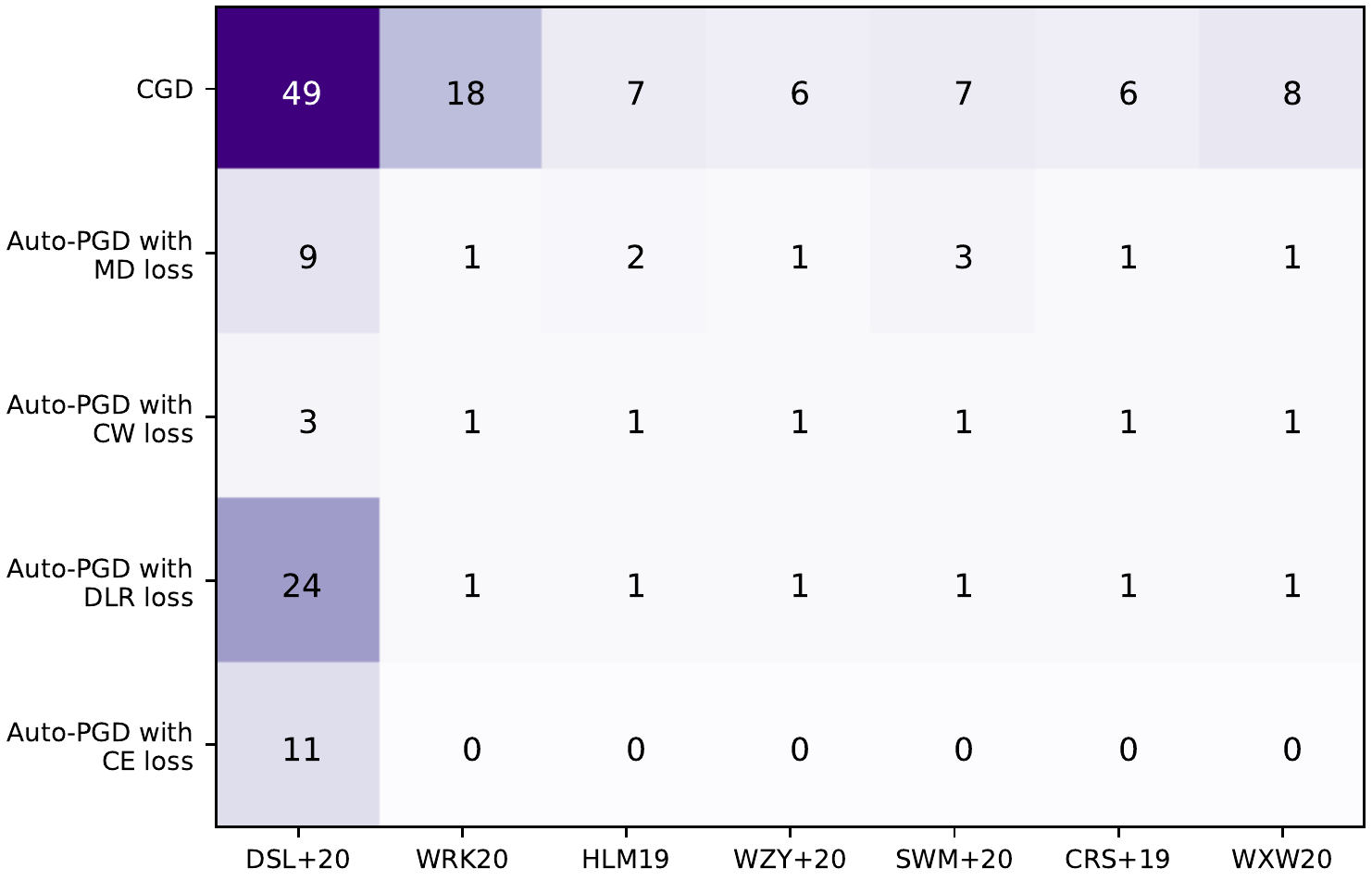}}
\caption{This is the average number of successful adversarial examples found by an attack but not by 
any of other attack methods, against each defense in each set of 10,000 attempts on the testing set 
of CIFAR10 using $\epsilon=8/255$, rounded to whole numbers. }
\label{CIFAR10-8-1VSothers}
\end{figure}

\begin{figure}[ht!]
\centerline{\includegraphics[width=0.95\columnwidth]{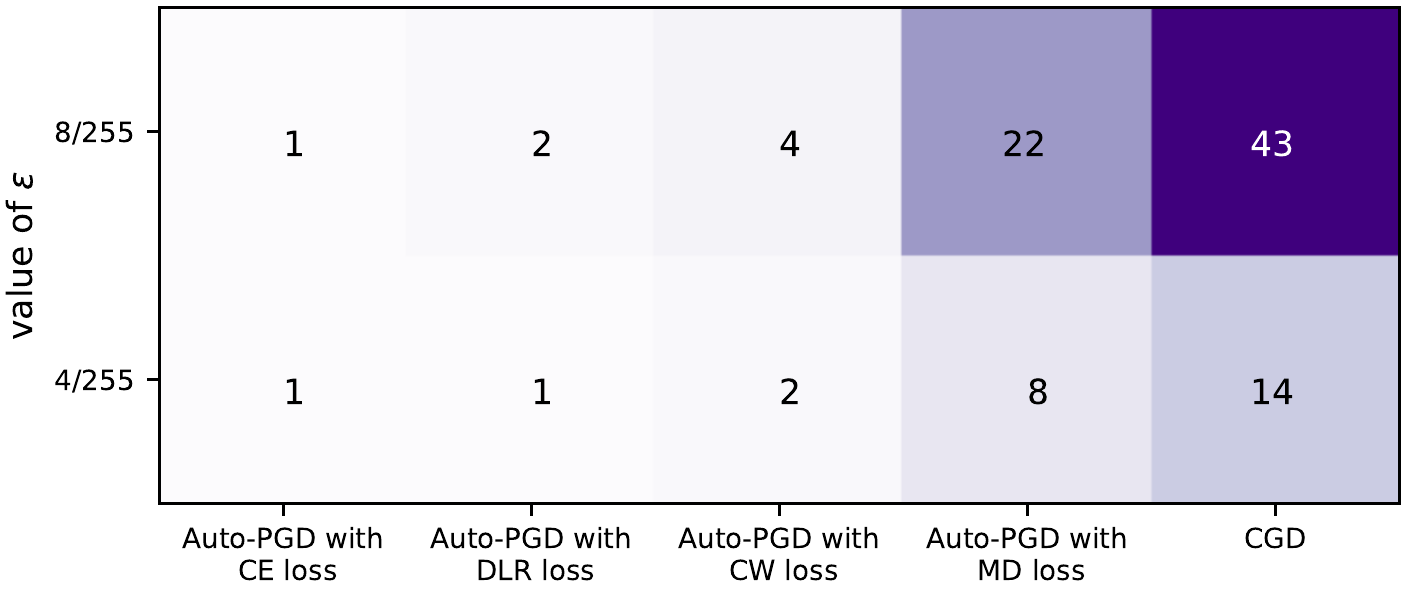}}
\caption{This is the average number of successful adversarial examples found by an attack but not by 
any other attack methods, against the \SIEKM~\cite{NeurIPS20:SIEKM20} defense in each set of 
50,000 attempts on the testing set of \Imagenet , no decimals are 
kept. }
\label{ImageNet1VSothers}
\end{figure}

\begin{figure}[ht!]
\centerline{\includegraphics[width=0.95\columnwidth]{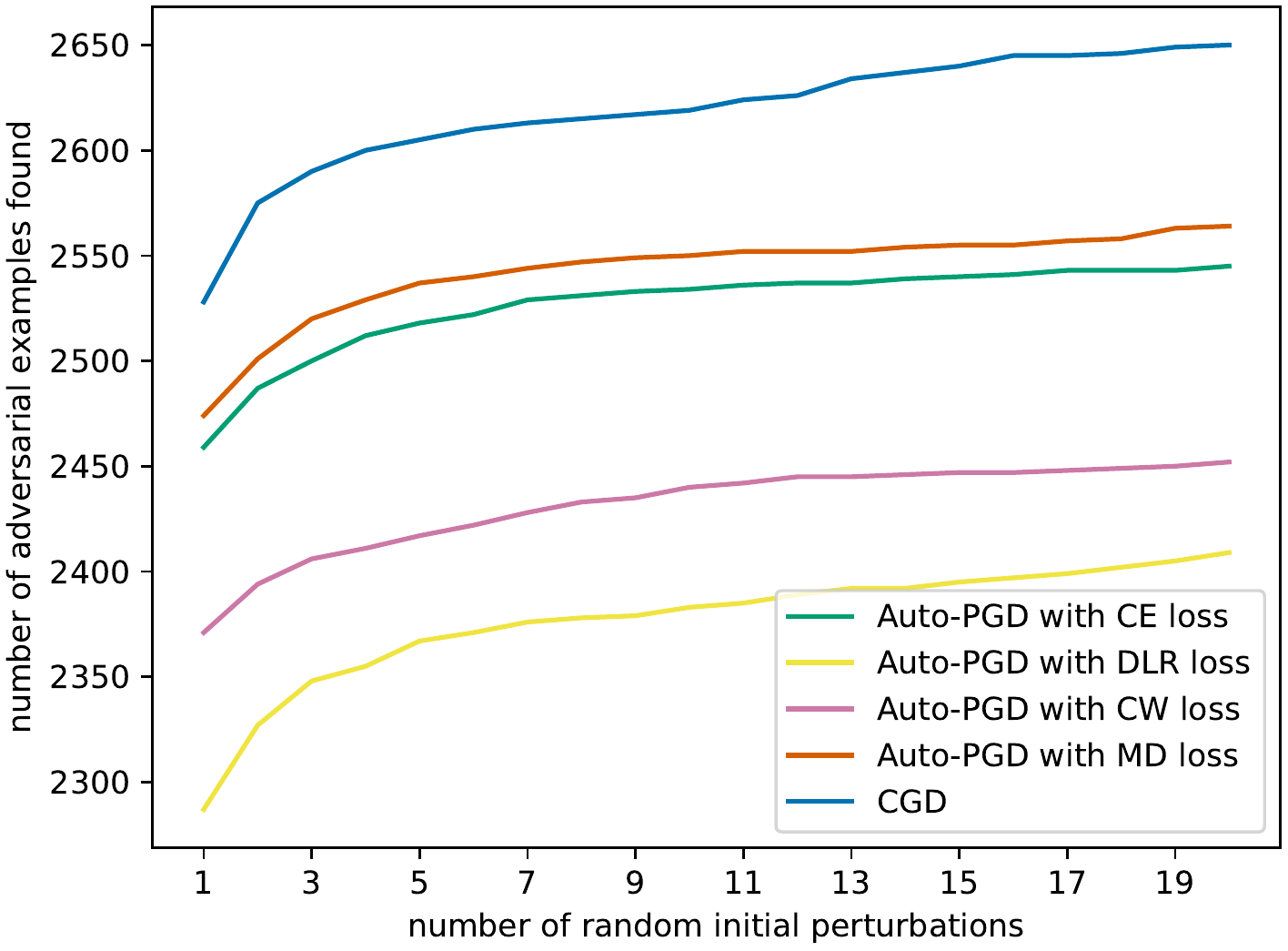}}
\caption{The image shows the number of adversarial examples each of the attacks found when they are 
allowed to use the specified number of random initial perturbations,
 on all 10,000 images from the testing set of CIFAR10, with $\epsilon=8/255$, against the 
\DSLH~\cite{iclr20:DSLH20} defense. }
\label{CIFAR10-DSLH-seeds8}
\end{figure}

At the same time, when the same attack method is given
different random initial perturbations, the attack always found a slightly different set of
adversarial examples, within the 20 random initial perturbations
we tried.  We observe the same
phenomena across attack methods, values of $\epsilon$, and defenses on CIFAR10,
as shown in \figrefs{CIFAR10-DSLH-seeds8}{CIFAR10-WXW-seeds8}.

\begin{figure}[ht!]
\centerline{\includegraphics[width=0.95\columnwidth]{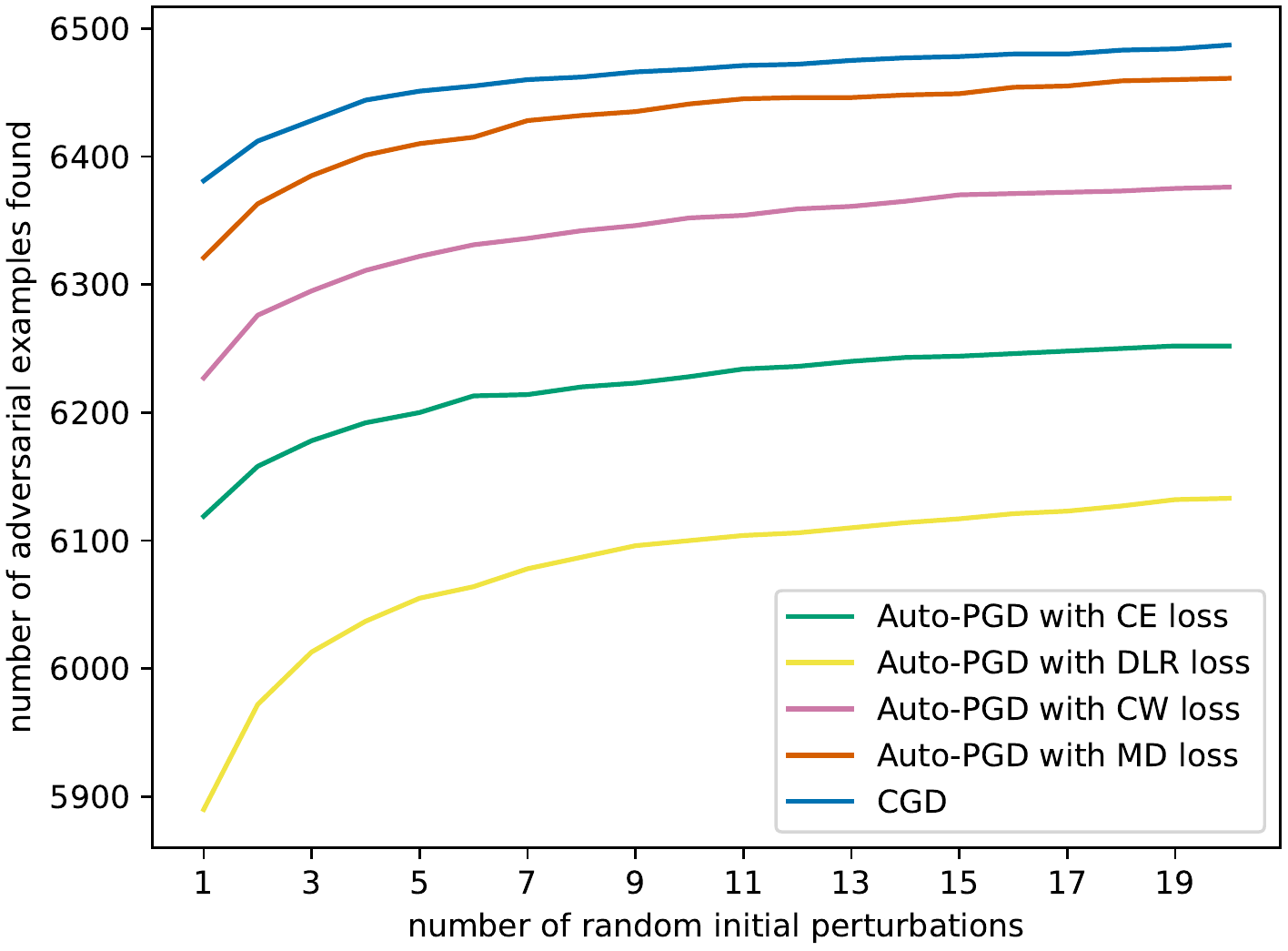}}
\caption{The image shows the number of adversarial examples each of the attacks found when they are 
allowed to use the specified number of random initial perturbations,
 on all 10,000 images from the testing set of CIFAR10, with $\epsilon=16/255$, against the 
\WRK~\cite{iclr20:WRK20} defense.}
\label{CIFAR10-WRK-seeds16}
\end{figure}

\begin{figure}[ht!]
\centerline{\includegraphics[width=0.95\columnwidth]{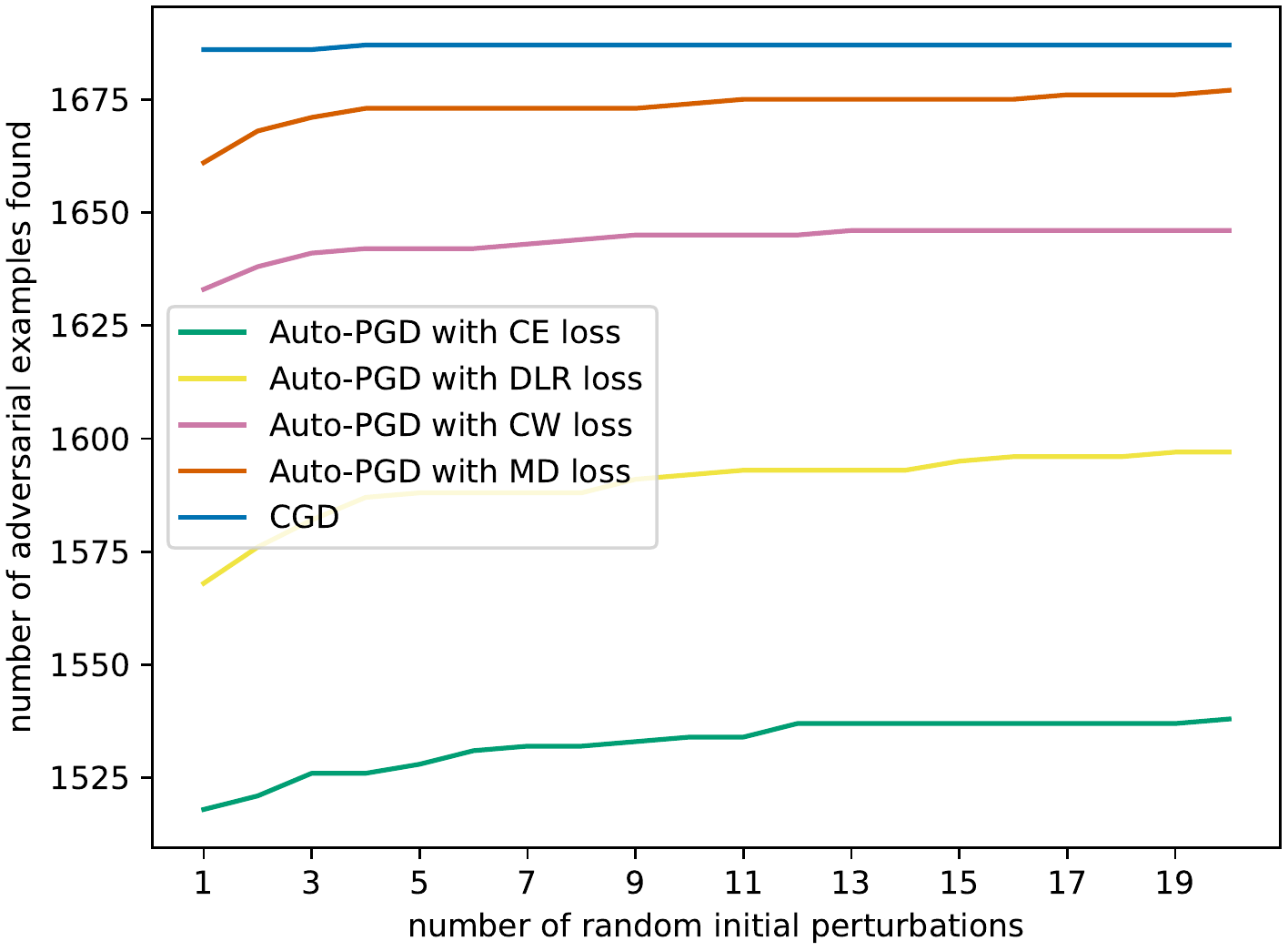}}
\caption{The image shows the number of adversarial examples each of the attacks found when they are 
allowed to use the specified number of random initial perturbations,
 on all 10,000 images from the testing set of CIFAR10, with $\epsilon=8/255$, against the 
\WRK~\cite{iclr20:WRK20} defense. }
\label{CIFAR10-WRK-seeds8}
\end{figure}

\begin{figure}[ht!]
\centerline{\includegraphics[width=0.95\columnwidth]{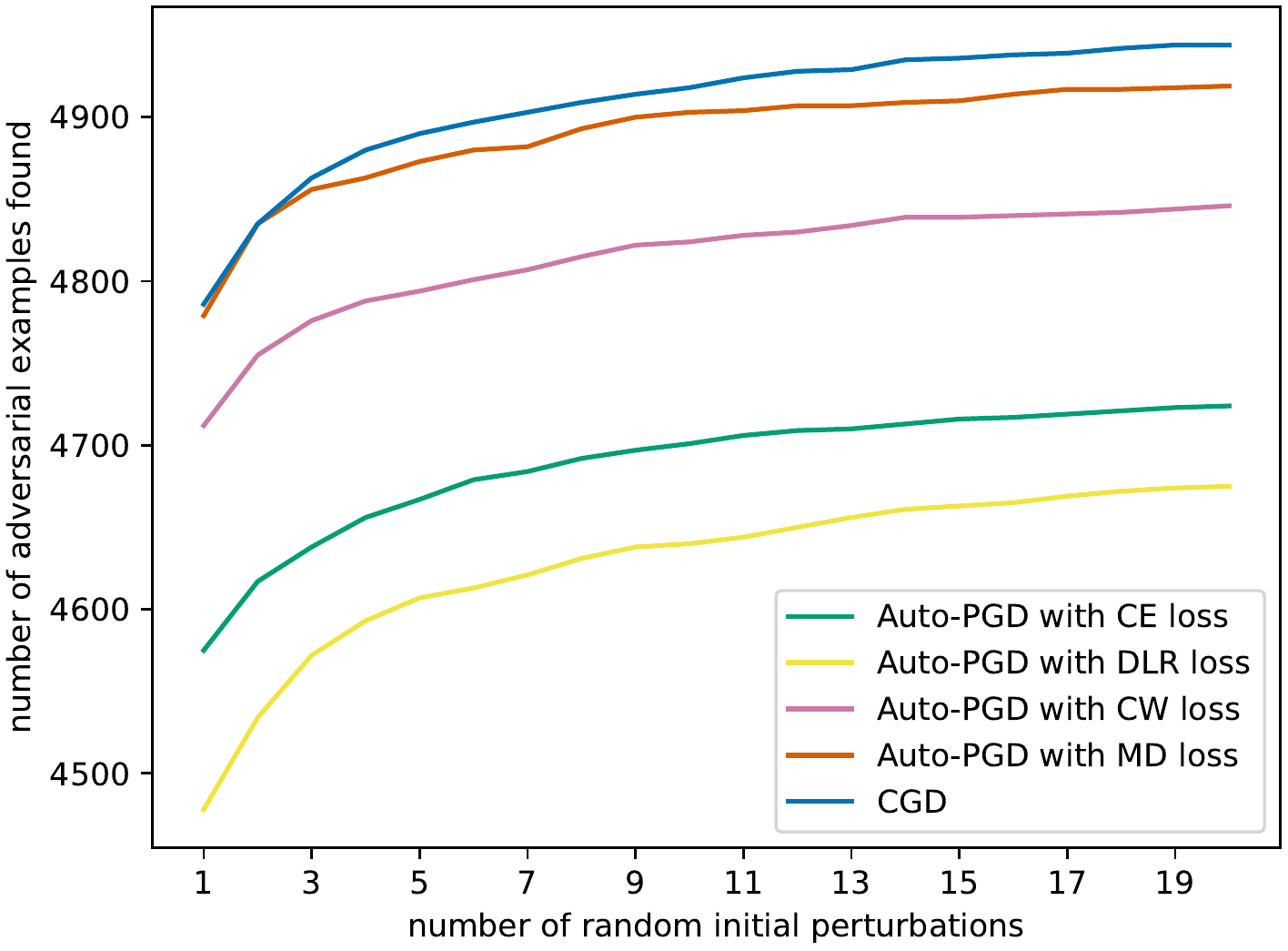}}
\caption{The image shows the number of adversarial examples each of the attacks found when they are 
allowed to use the specified number of random initial perturbations,
 on all 10,000 images from the testing set of CIFAR10, with $\epsilon=16/255$, against the 
\HLM~\cite{icml19:HLM19} defense.}
\label{CIFAR10-HLM-seeds16}
\end{figure}

\begin{figure}[ht!]
\centerline{\includegraphics[width=0.95\columnwidth]{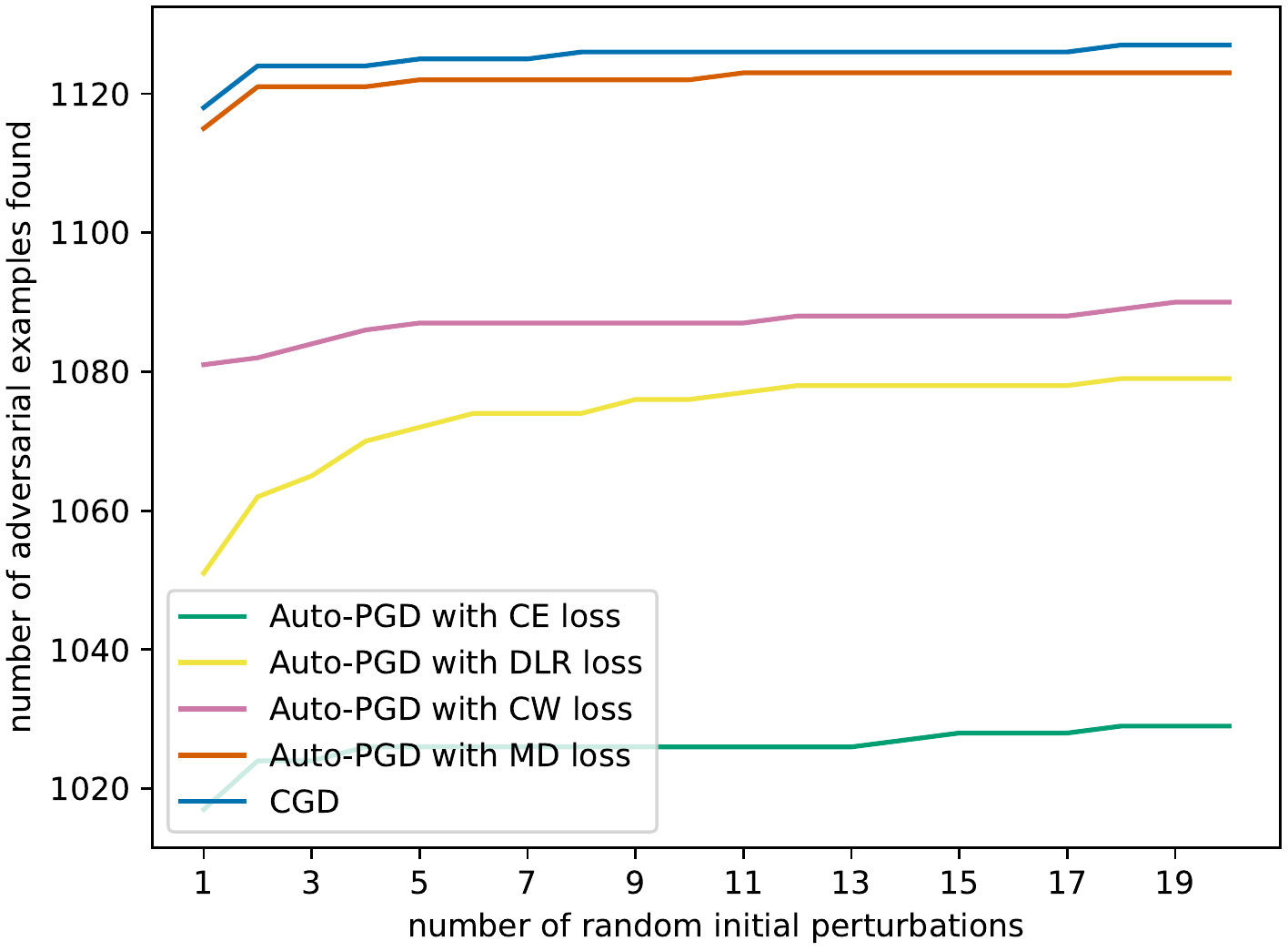}}
\caption{The image shows the number of adversarial examples each of the attacks found when they are 
allowed to use the specified number of random initial perturbations,
 on all 10,000 images from the testing set of CIFAR10, with $\epsilon=8/255$, against the 
\HLM~\cite{icml19:HLM19} defense.}
\label{CIFAR10-HLM-seeds8}
\end{figure}

\begin{figure}[ht!]
\centerline{\includegraphics[width=0.95\columnwidth]{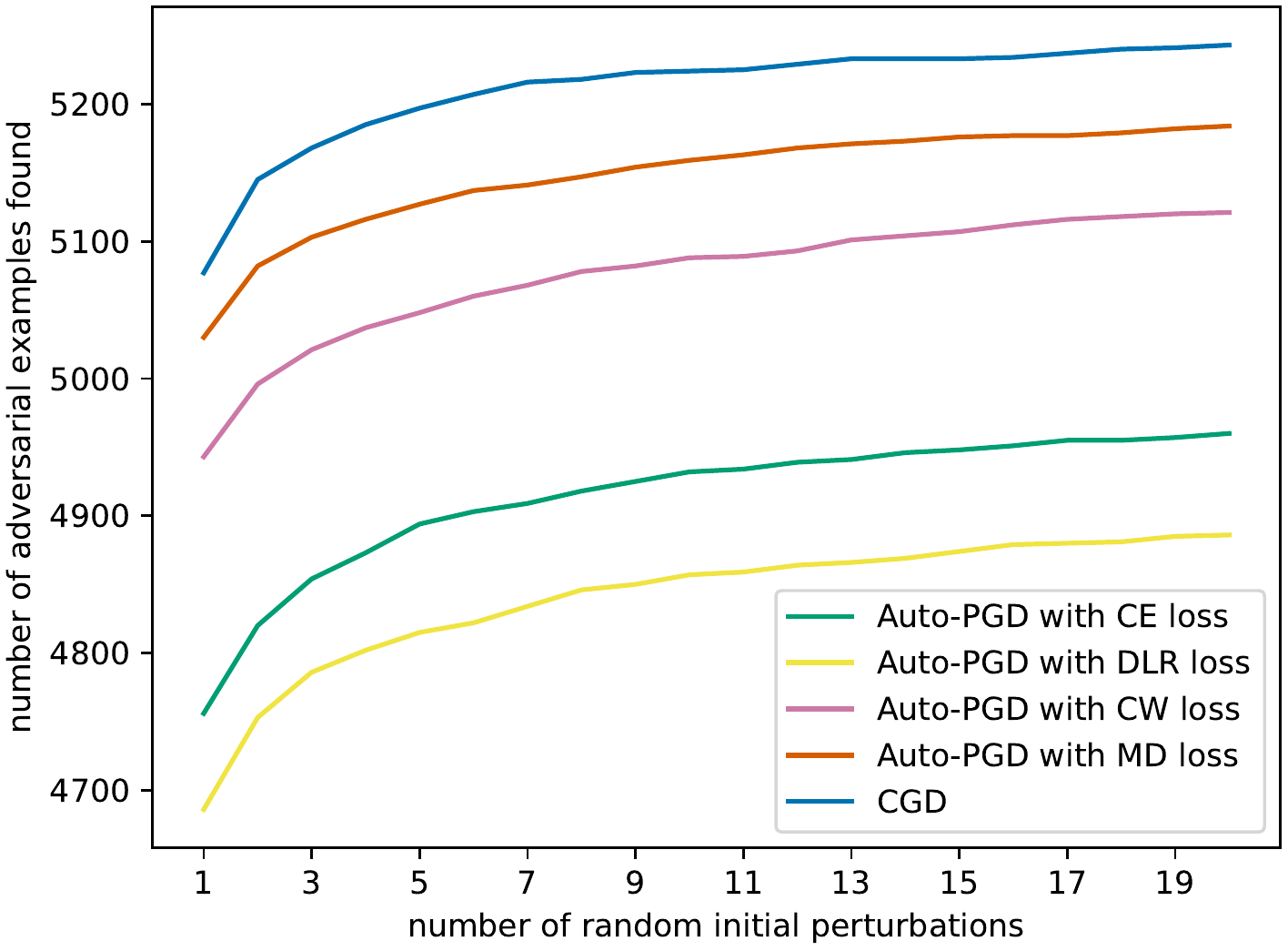}}
\caption{The image shows the number of adversarial examples each of the attacks found when they are 
allowed to use the specified number of random initial perturbations,
 on all 10,000 images from the testing set of CIFAR10, with $\epsilon=16/255$, against the 
\WZYBMG~\cite{iclr20:WZYBMG20} defense. }
\label{CIFAR10-WZYBMG-seeds16}
\end{figure}

\begin{figure}[ht!]
\centerline{\includegraphics[width=0.95\columnwidth]{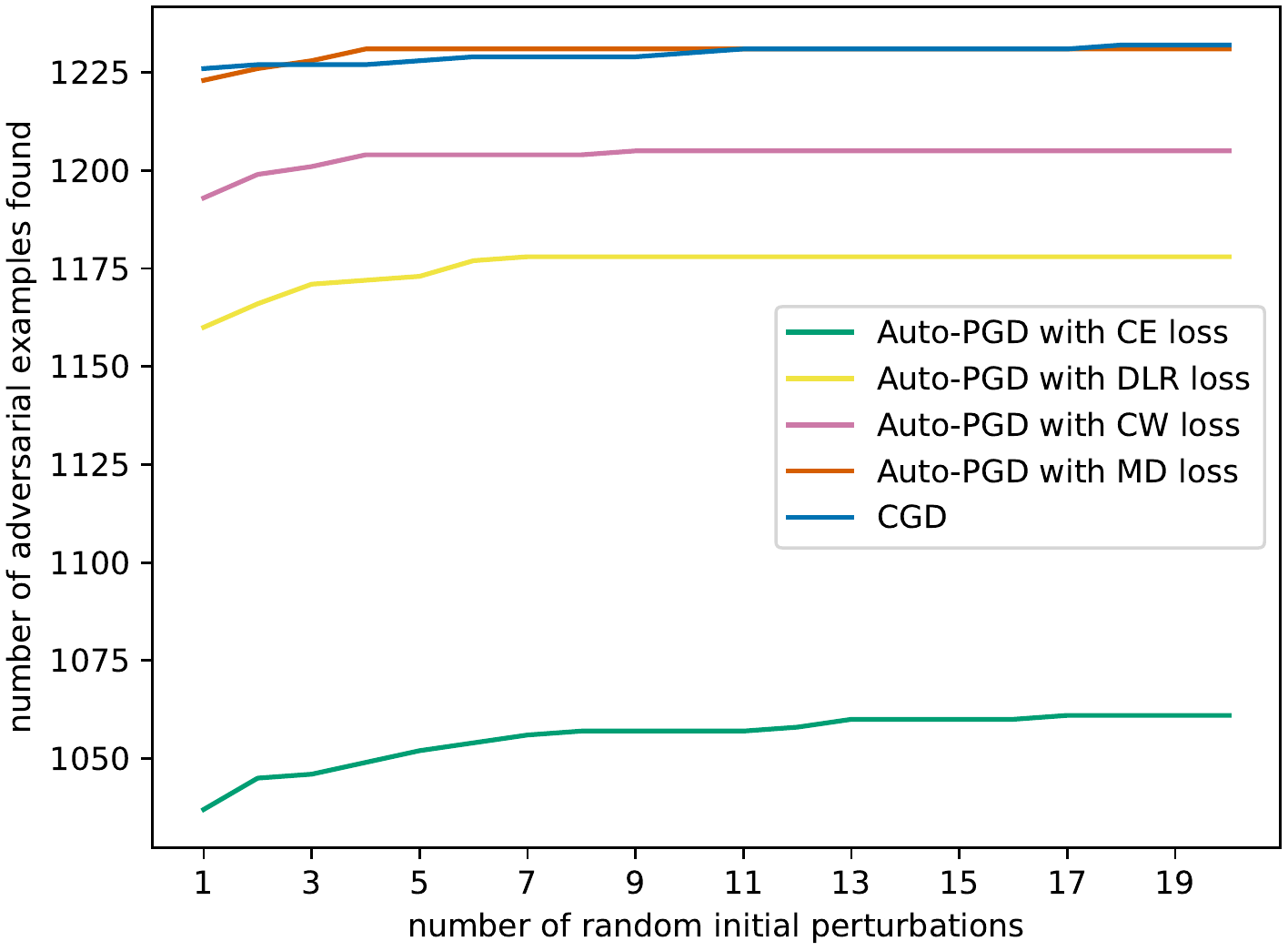}}
\caption{The image shows the number of adversarial examples each of the attacks found when they are 
allowed to use the specified number of random initial perturbations,
 on all 10,000 images from the testing set of CIFAR10, with $\epsilon=8/255$, against the 
\WZYBMG~\cite{iclr20:WZYBMG20} defense.}
\label{CIFAR10-WZYBMG-seeds8}
\end{figure}

\begin{figure}[ht!]
\centerline{\includegraphics[width=0.95\columnwidth]{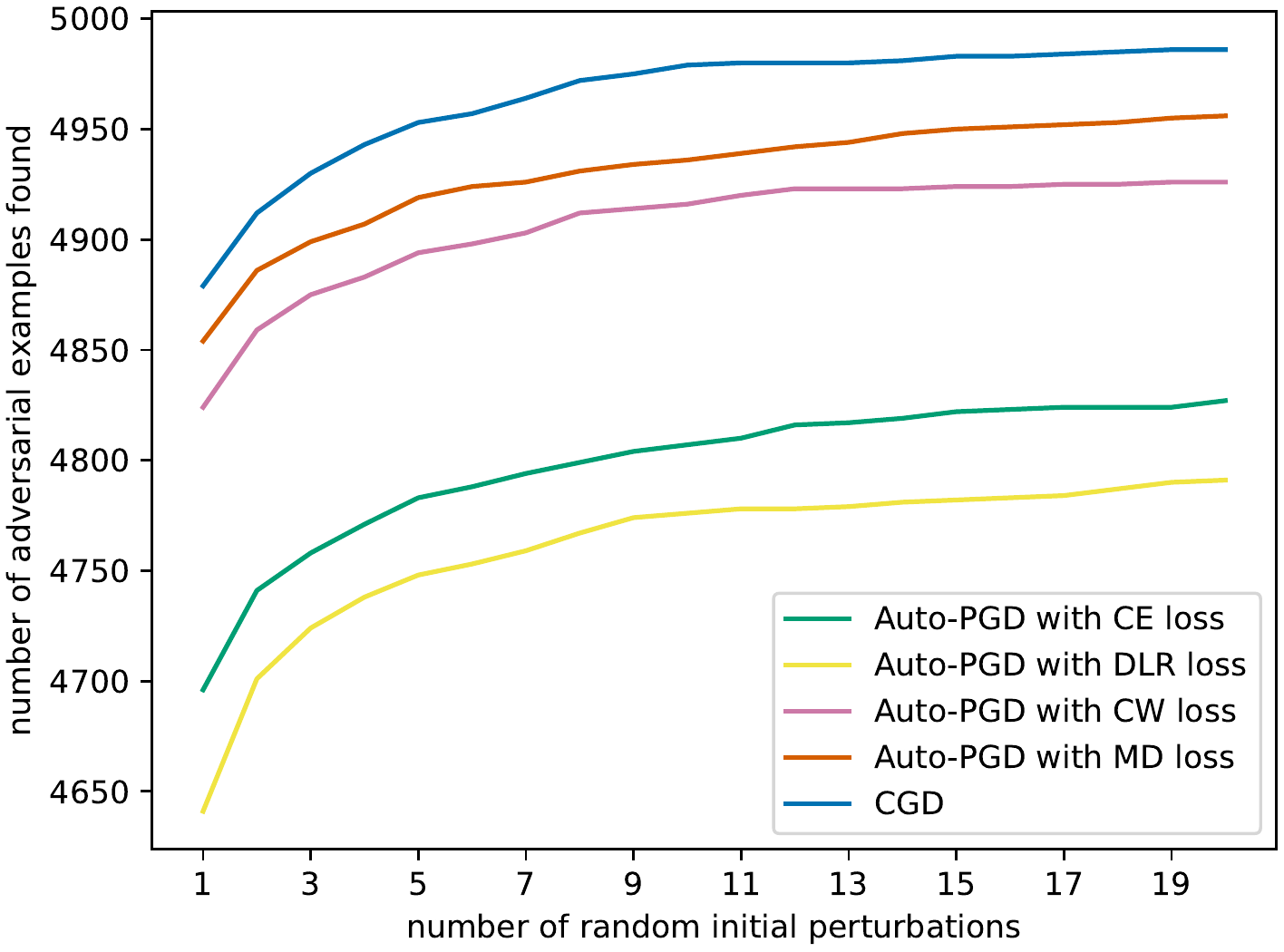}}
\caption{The image shows the number of adversarial examples each of the attacks found when they are 
allowed to use the specified number of random initial perturbations,
 on all 10,000 images from the testing set of CIFAR10, with $\epsilon=16/255$, against the 
\SWMJ~\cite{NeurIPS20:SWMJ20} defense. }
\label{CIFAR10-SWMJ-seeds16}
\end{figure}

\begin{figure}[ht!]
\centerline{\includegraphics[width=0.95\columnwidth]{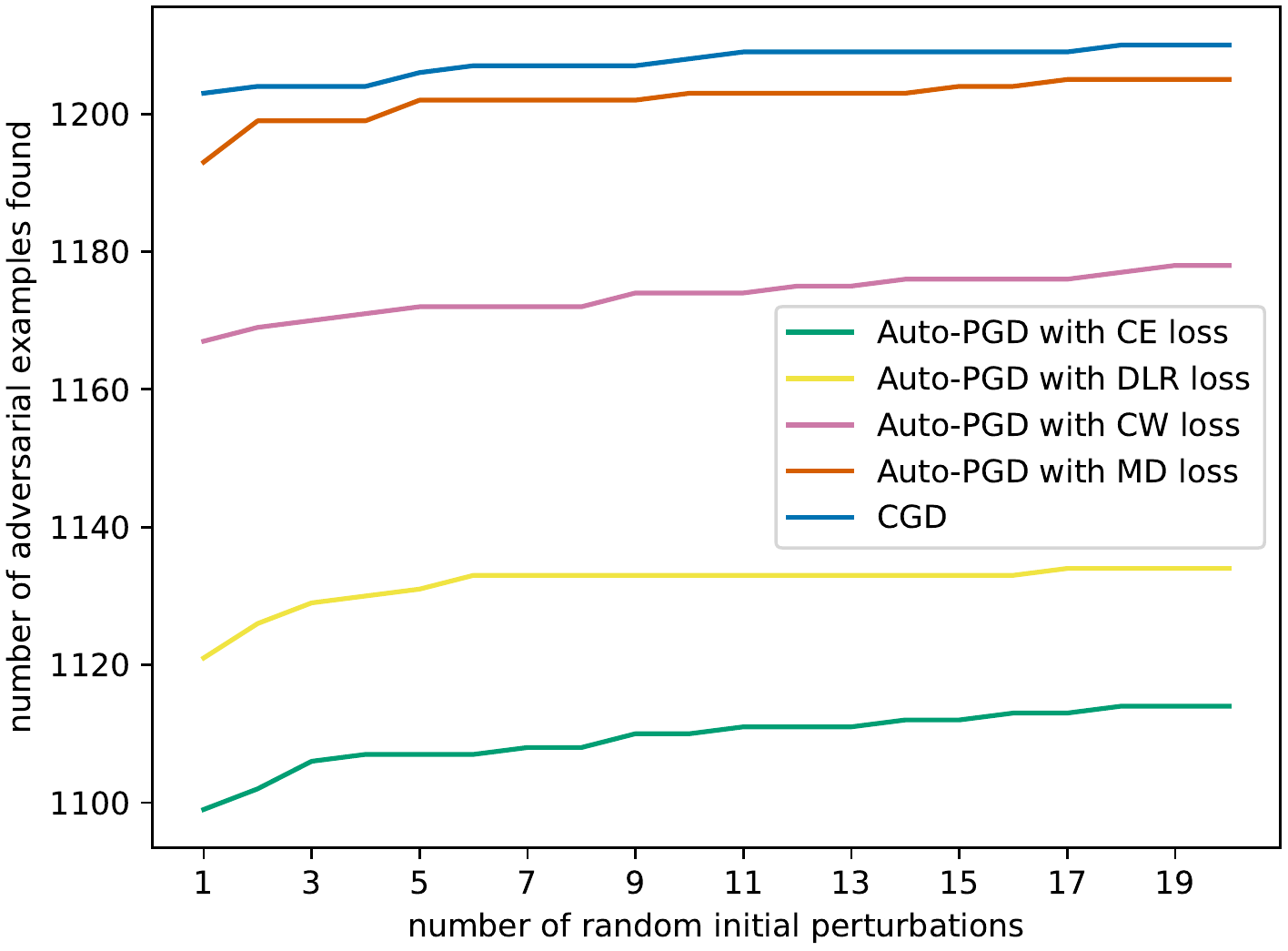}}
\caption{The image shows the number of adversarial examples each of the attacks found when they are 
allowed to use the specified number of random initial perturbations,
 on all 10,000 images from the testing set of CIFAR10, with $\epsilon=8/255$, against the 
\SWMJ~\cite{NeurIPS20:SWMJ20} defense. }
\label{CIFAR10-SWMJ-seeds8}
\end{figure}

\begin{figure}[ht!]
\centerline{\includegraphics[width=0.95\columnwidth]{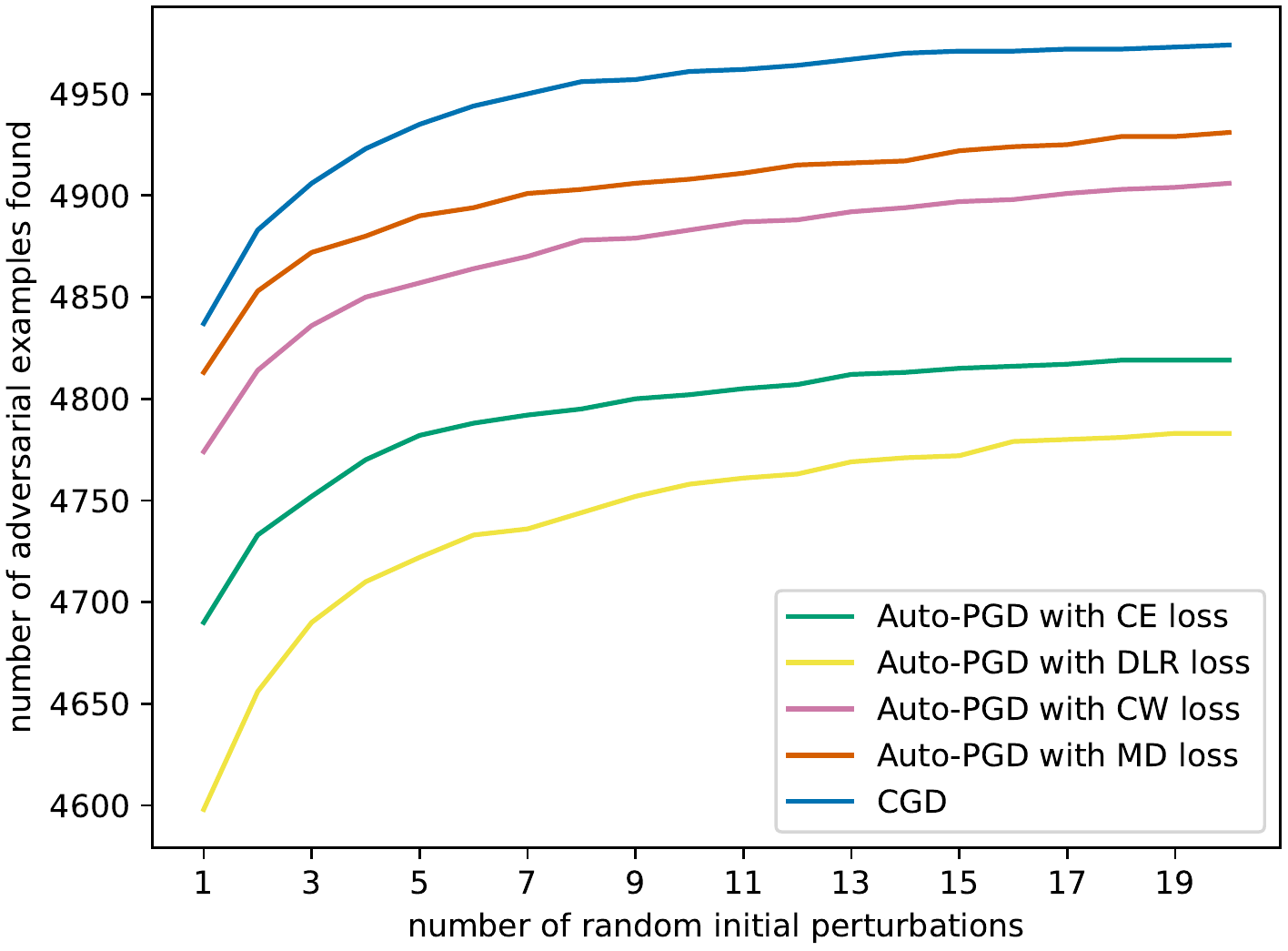}}
\caption{The image shows the number of adversarial examples each of the attacks found when they are 
allowed to use the specified number of random initial perturbations,
 on all 10,000 images from the testing set of CIFAR10, with $\epsilon=16/255$, against the 
\CRSLD~\cite{NeurIPS19:CRSLD19} defense.}
\label{CIFAR10-CRSLD-seeds16}
\end{figure}

\begin{figure}[ht!]
\centerline{\includegraphics[width=0.95\columnwidth]{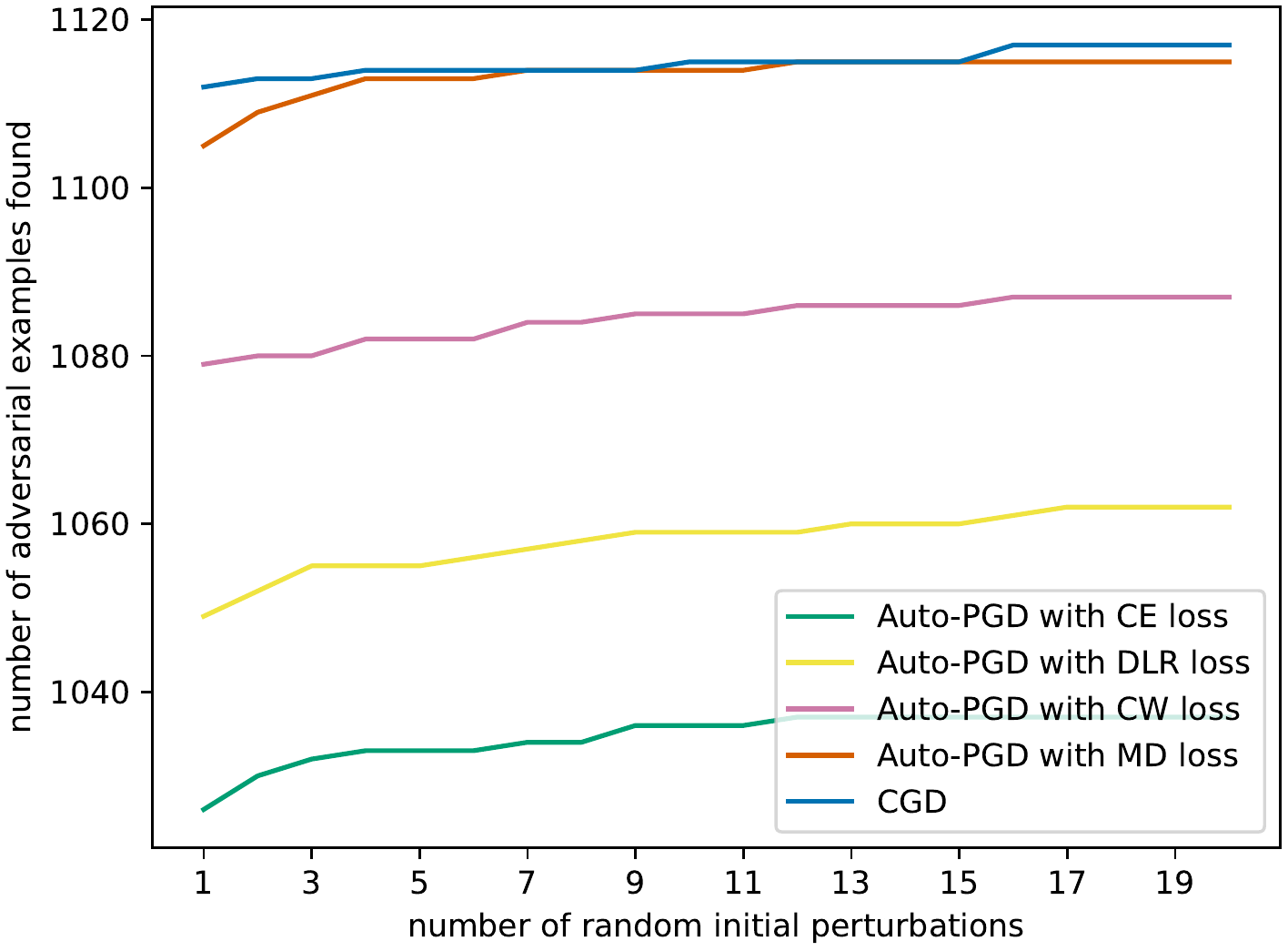}}
\caption{The image shows the number of adversarial examples each of the attacks found when they are 
allowed to use the specified number of random initial perturbations,
 on all 10,000 images from the testing set of CIFAR10, with $\epsilon=8/255$, against the 
\CRSLD~\cite{NeurIPS19:CRSLD19} defense.}
\label{CIFAR10-CRSLD-seeds8}
\end{figure}

\begin{figure}[ht!]
\centerline{\includegraphics[width=0.95\columnwidth]{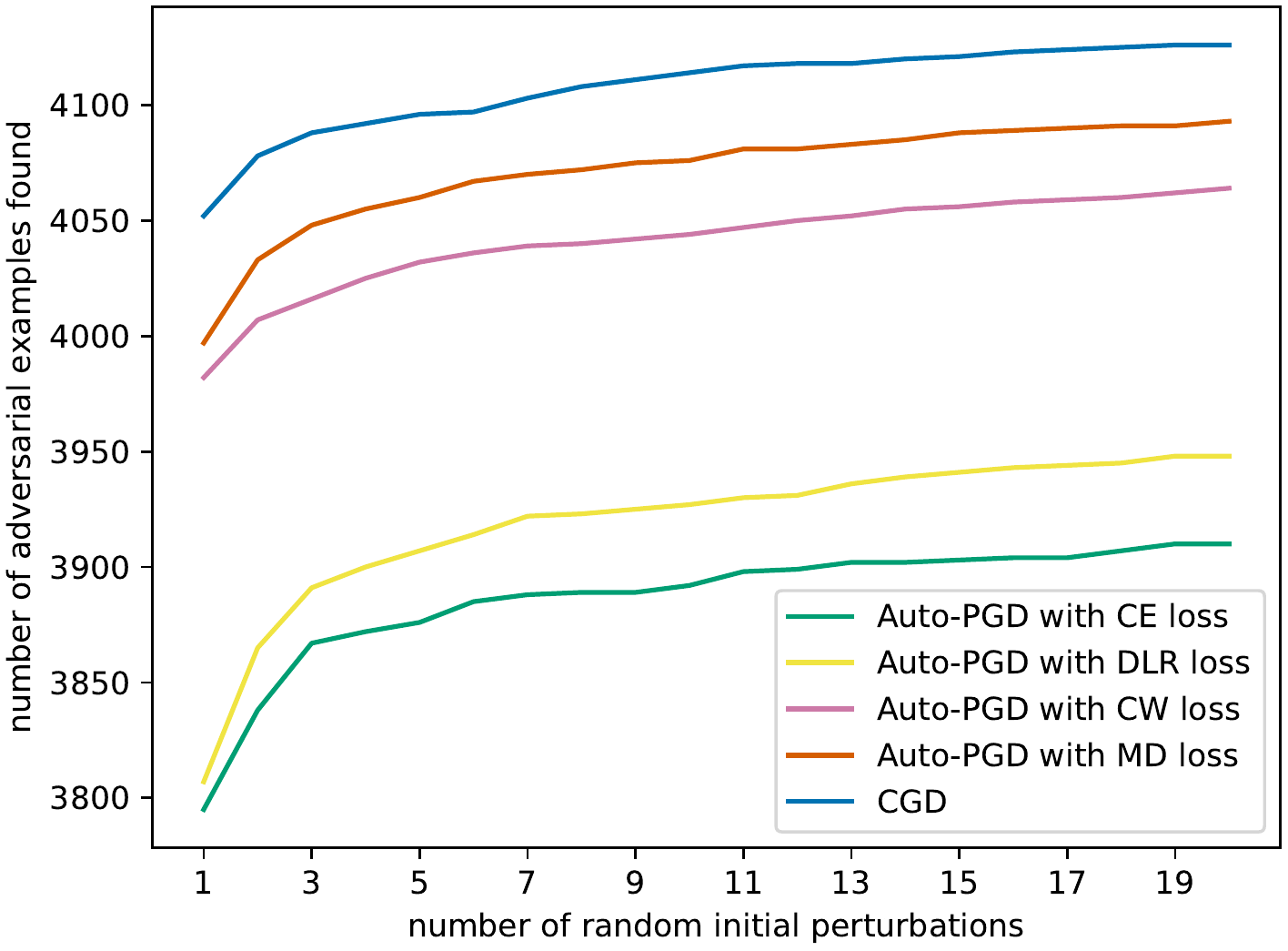}}
\caption{The image shows the number of adversarial examples each of the attacks found when they are 
allowed to use the specified number of random initial perturbations,
 on all 10,000 images from the testing set of CIFAR10, with $\epsilon=16/255$, against the 
\WXW~\cite{NeurIPS20:WXW20} defense. }
\label{CIFAR10-WXW-seeds16}
\end{figure}

\begin{figure}[ht!]
\centerline{\includegraphics[width=0.95\columnwidth]{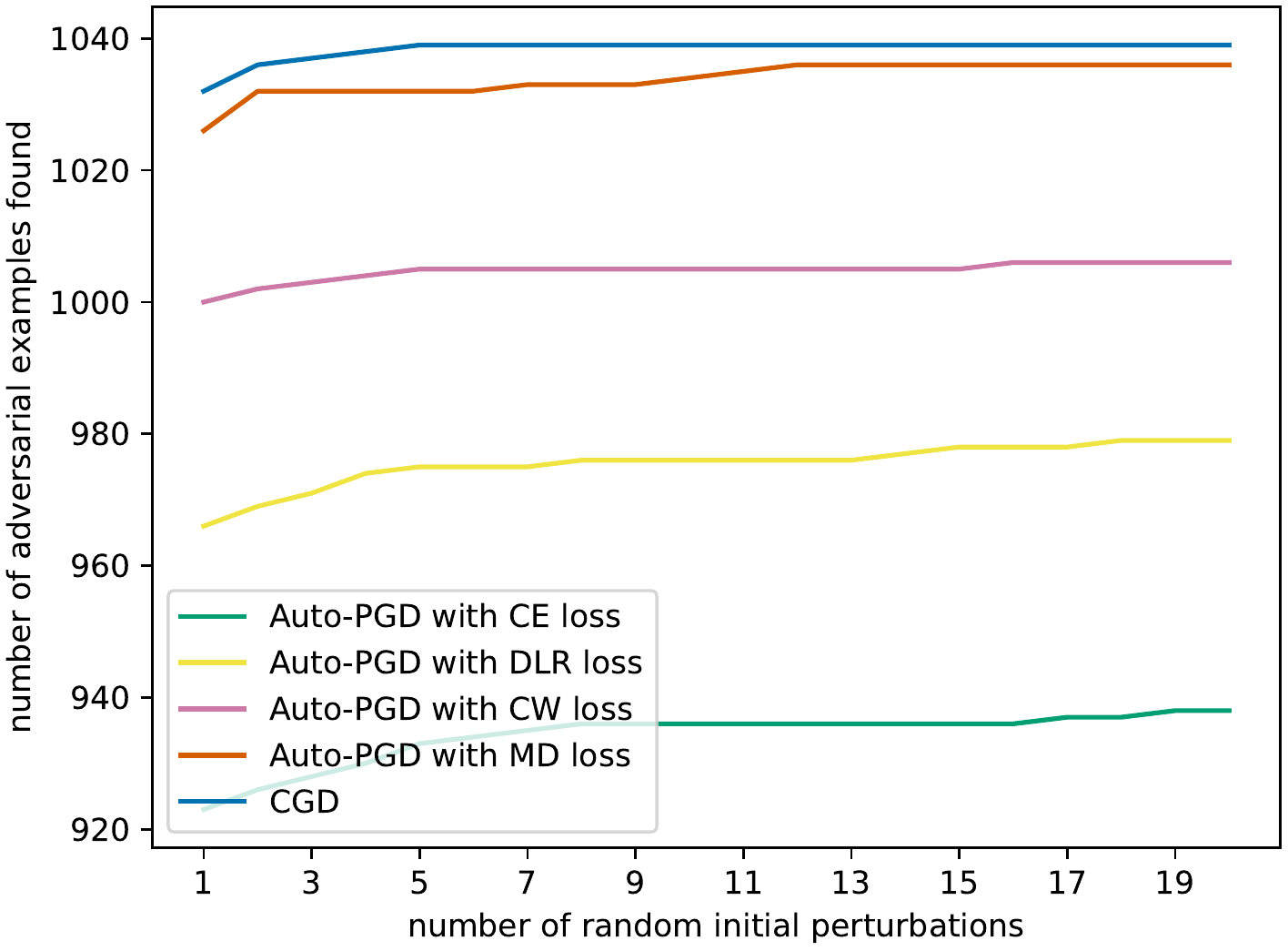}}
\caption{The image shows the number of adversarial examples each of the attacks found when they are 
allowed to use the specified number of random initial perturbations,
 on all 10,000 images from the testing set of CIFAR10, with $\epsilon=8/255$, against the 
\WXW~\cite{NeurIPS20:WXW20} defense. }
\label{CIFAR10-WXW-seeds8}
\end{figure}

\section{Gradient Based Quantization}
\label{app:smart:rounding}
\begin{figure}[ht!]
\centerline{\includegraphics[width=0.95\columnwidth]{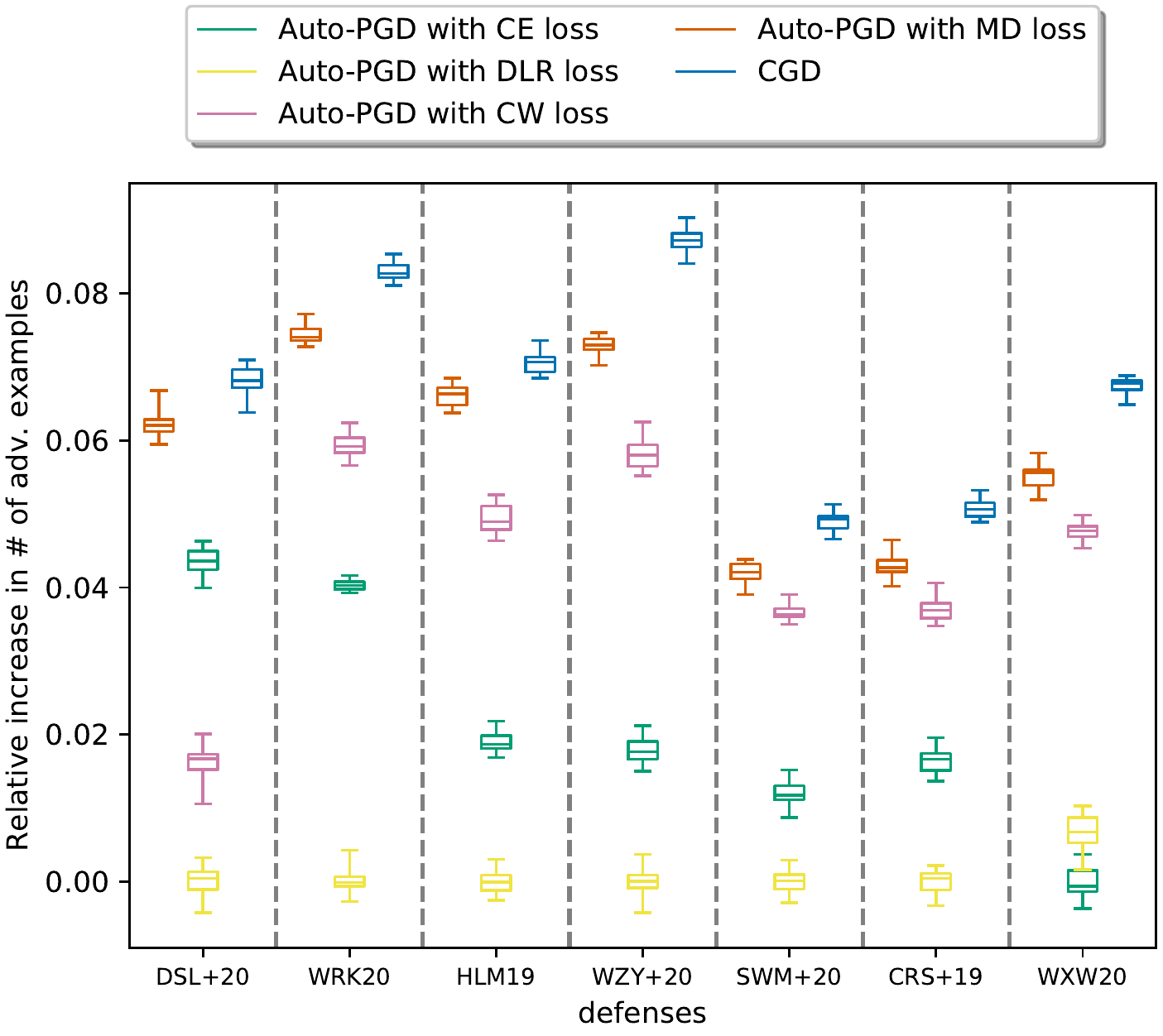}}
\caption{ This figure shows the relative improvement in the number of adversarial examples found by
  attacks on different defenses compared to the worst-performing
  attack, when all attacks are allowed to perform one more forward propagation to verify if 
  a gradient based quantized version of the current perturbation would lead to success.
  Experiments were performed using 10,000 images from the testing set of CIFAR10. We ran 
attacks using $\epsilon=16/255$, 20 different random initial perturbations with seeds 0--19, and a fixed 
random target offset with seed 
0. The result is normalized by the mean of the worst performing method against each model.}
\label{boxperformanceCIFAR16sr}
\end{figure}
Bonnet et al.\ proposed gradient based quantization, an approach to round the current perturbation along the sign of the gradients \cite{ihmmsec20:linfrounding}. We ran experiments with gradient-based quantization, using the formula
\[x_{\mathit{test}}=\mathit{round}(x'_i*255+sign(\frac{\partial L(x'_i,t)}{\partial x'_i})*.499999)/255\] in addition to the quantization described in \secref{sec:setup:Evaluation}, and declared an attack successful if either of the quantizations produced a perturbation that caused the model to predict the target class in any iteration. 
We ran all attacks at $\epsilon=16/255$ against defenses on CIFAR10, with the same setup as described in \secref{sec:setup}. 
The results are shown in \figref{boxperformanceCIFAR16sr}. With an additional 8.9--13.1\% time cost (measured using the same approach as in \secref{sec:result:time}), the attacks succeed only by an additional 0.00--0.04\%; the relative relationship between success rates of different attacks remains the same.

\end{document}